%% file: for_arxiv2.tex
\documentclass[11pt]{article} 
\usepackage{times}
\usepackage{url}            
\usepackage{booktabs}       
\usepackage{amsfonts}       
\usepackage{nicefrac}       
\usepackage{microtype}      
\usepackage{mkolar_definitions}
\usepackage{xspace}
\usepackage{multirow}
\usepackage{amsmath}
\usepackage{algorithm}
\usepackage{algorithmic}
\usepackage{color}
\usepackage{color, colortbl}
\usepackage{enumitem}
\usepackage{comment}
\usepackage{bm}
\usepackage{caption}
\usepackage{subcaption}

 \usepackage[left=2cm,top=2cm,right=2cm]{geometry}

\usepackage[scaled=.9]{helvet}

\usepackage{url}
\usepackage{authblk}
\usepackage{csquotes}
\usepackage{natbib}

\usepackage[dvipsnames]{xcolor}
\usepackage[colorlinks=true, linkcolor=blue, citecolor=blue]{hyperref}

\newcount\Comments  
\definecolor{darkgreen}{rgb}{0,0.5,0}
\definecolor{darkred}{rgb}{0.7,0,0}
\definecolor{teal}{rgb}{0.3,0.8,0.8}
\definecolor{orange}{rgb}{1.0,0.5,0.0}
\definecolor{purple}{rgb}{0.8,0.0,0.8}
\newcommand{\kibitz}[2]{\ifnum\Comments=1{\textcolor{#1}{\textsf{\footnotesize #2}}}\fi}

\makeatletter
\newcommand{\printfnsymbol}[1]{%
  \textsuperscript{\@fnsymbol{#1}}%
}
\makeatother

\title{A Minimax Learning Approach to Off-Policy Evaluation in Confounded Partially Observable Markov Decision Processes}

\author[1]{Chengchun Shi\thanks{Equal contribution.}}
\author[2]{Masatoshi Uehara\printfnsymbol{1}}
\author[3]{Jiawei Huang }
\author[3]{ Nan Jiang }
\affil[ ]{  \href{c.shi7@lse.ac.uk}{c.shi7@lse.ac.uk},\quad
\href{mu223@cornell.edu}{mu223@cornell.edu},\quad
\href{jiaweih@illinois.edu}{jiaweih@illinois.edu},\quad
\href{nanjiang@illinois.edu}{nanjiang@illinois.edu}}
\affil[1]{Department of Statistics, London School of Economics and Political Science}
\affil[2]{Department of Computer Science, Cornell University}
\affil[3]{Department of Computer Science, University of Illinois Urbana-Champaign}
\date{}

\definecolor{Gray}{gray}{0.9}

\input{notation.tex}

\renewcommand{\arraystretch}{2.}

\input{math_commands.tex}

\usepackage{mkolar_definitions}
\let\proglang=\textsf
\begin{document}

\maketitle

\begin{abstract}
We consider off-policy evaluation (OPE) in Partially Observable Markov Decision Processes (POMDPs), where the evaluation policy depends only on observable variables and the behavior policy depends on unobservable latent variables. Existing works either assume no unmeasured confounders, or focus on settings where both the observation and the state spaces are tabular. In this work, we first propose novel identification methods for OPE in POMDPs with latent confounders, by introducing bridge functions that link the target policy's value and the observed data distribution. We next propose minimax estimation methods for learning these bridge functions, and construct three estimators based on these estimated bridge functions, corresponding to a value function-based estimator, a marginalized importance sampling estimator, and a doubly-robust estimator. Our proposal permits general function approximation and is thus applicable to settings with continuous or large observation/state spaces. The nonasymptotic and asymptotic properties of the proposed estimators are investigated in detail. A \proglang{Python} implementation of our proposal is available at \url{https://github.com/jiaweihhuang/Confounded-POMDP-Exp}.
\end{abstract}

\section{Introduction}

Reinforcement learning (RL) has been successfully applied 
in online settings (e.g., video games) where interaction with environments is easy and the data can be adaptively generated. 
However, online interaction is often costly and dangerous for a number of high-stake domains ranging from health science to social science and economics. 
Offline (batch) reinforcement learning is concerned with 
policy learning and evaluation with limited and pre-collected data in a sample efficient manner \cite{levine2020offline}. The focus of this paper is the off-policy evaluation (OPE) problem, which refers to the task of estimating the value of an evaluation policy offline using data collected under a different behavior policy. It is critical in sequential decision-making problems from healthcare, robotics, and education 
where new policies need to be evaluated offline before online validation. 

There is a growing literature on OPE \citep[see e.g.,][]{precup2000eligibility,jiang2016doubly,thomas2016data,Liu2018,XieTengyang2019OOEf,yin2020asymptotically,liao2020batch,yang2020off,PananjadyAshwin2021,KuzborskijIlja2020COEa,ZhangMichaelR2021ADMf,shi2021deeply}. A common assumption made in the aforementioned works is that of no unmeasured confounders. Specifically, they assume the time-varying state variables are fully observed and no unmeasured variables exist that confound the
observed actions. However, this assumption is not testable
from the data. It is often violated in observational datasets generated from healthcare applications. 

To allow unmeasured confounders to exist, we model the observed data by a Partially Observable Markov Decision Process (POMDP). Under this framework, the behavior policy 
is allowed to depend on some unobserved state variables that confound the action-reward association. 
The goal of OPE in POMDPs is to estimate the value of an evaluation policy, 
which is a function of observed variables, using the data generated by such a behavior policy. 
This is a highly challenging problem. Directly applying 
the importance sampling methods \cite{precup2000eligibility,Liu2018} or the value function-based method \cite{munos2008finite} would yield a biased estimator, as we do not have access to the unobserved state variables. 
\citet{tennenholtz2020off, nair2021spectral} have made important progresses on this problem  by outlining a consistent OPE estimator in tabular settings. 
However, they are not applicable to settings with continuous observation/state spaces. 

In this paper, we study OPE in non-tabular POMDPs. Our contributions are summarized as follows. First, we provide a novel 
identification method for OPE with latent confounders. We only require the 
existence of two bridge functions, corresponding to a 
weight bridge function and a value bridge function. These bridge functions are defined as solutions to some integral equations. 
They can be interpreted as projections of 
the marginalized importance sampling weight and value functions defined on the latent state space onto the observation space. They are not always uniquely defined, but we do not require the uniqueness assumption to 
achieve consistent estimation. 

Second, we propose minimax learning methods to estimate the two bridge functions with function approximation. The proposed method allows us to model these bridge functions via certain highly flexible function approximators (e.g., neural networks) and is thus applicable to settings with continuous or large observation/state spaces. Based on the estimated bridge functions, we further propose three estimators for the target policy's value, corresponding to a value function-based estimator, a marginalized importance sampling (IS) estimator and a doubly-robust (DR) estimator.

Finally, we systematically study the nonasymptotic and asymptotic properties of the proposed estimators. 
Specifically, we first show the finite-sample rate of convergence under the realizability and Bellman closedness assumptions on value bridge function. 
Similar assumptions are imposed on the Q- or density ratio function in fully observable MDP settings \cite{munos2008finite,uehara2021finite}. Second, when the realizability assumption holds for both bridge functions, we 
establish the finite-sample rate of convergence of the proposed estimators without assuming Bellman closedness. 
Finally, when the bridge functions are uniquely defined,
we prove that the DR estimator is asymptotically 
normal and achieves the Cram\'er-Rao lower bound. We expect our 
findings to 
contribute to the theoretical understanding of offline RL \citep[see e.g.,][]{munos2008finite,chen2019information,KallusNathan2019EBtC}. 

\subsection{Related Works}\label{sec:related}
We discuss some related works on OPE with unmeasured confounders and spectral learning in this section. To save space, some additional related works on minimax estimation and negative controls are discussed in Appendix~\ref{sec:related_add}. 

\vspace{-0.3cm}
\paragraph{OPE with Unmeasured Confounders} 
To handle unmeasured confounders, existing OPE methods can be roughly divided into the following three categories. The first type of methods proposes to develop partial identification bounds for the policy value based on sensitivity analysis \cite{Kallus2020_confounding,NamkoongHongseok2020OPEF,Zhang2021}. These methods rely on certain assumptions that might be difficult to validate in practice. For instance, \citet{Kallus2020_confounding}  imposes a memoryless unmeasured confounding assumption. 


The second type of methods derive the value estimates by making use of some auxiliary variables in the observed data \citep{zhang2016markov,futoma2020popcorn,wang2020provably,bennett2021off,liao2021instrumental,ying2021proximal,shi2022off}. These methods are not directly applicable to the POMDP setting we consider. For instance, \citet{zhang2016markov,wang2020provably,liao2021instrumental,shi2022off} propose to model the observed data via a confounded MDP. 
In these settings, the time-varying observations satisfy the Markov property. 
However, the Markov assumption is violated in POMDPs.

The third type of methods adopt the POMDP model to formulate the confounded OPE problem and develop value estimators in tabular settings \citep{tennenholtz2020off,nair2021spectral}. However, as we have commented, these methods are ineffective in non-tabular settings with continuous or large observation/state spaces.

\vspace{-0.3cm}
\paragraph{Spectral learning in POMDPs} 

Our proposal is closely related to a line of works on developing spectral learning methods in the POMDP literature \citep{song2010hilbert, boots2011closing,  hsu2012spectral, anandkumar2014tensor,hefny2015supervised, kulesza2015spectral,jin2020sample}. These methods are originally designed for learning system dynamics in the absence of unmeasured confounders. 
They would yield biased value estimators in the presence of unmeasured confounders.



\section{Problem Formulation}
We consider a 
discounted Partially Observable Markov Decision Process (POMDP) 
defined as $$\langle \Scal,\Acal,\Ocal, \{r_t\}_{t\ge 0},\{P_t\}_{t\ge 0},\{Z_t\}_{t\ge 0},\gamma \rangle,$$ where $\Scal$ denotes the state space, $\Acal$ denotes the finite action space, $\Ocal$ denotes the observation space, $P_t:\Scal \times \Acal \to \Delta(\Scal)$ denotes 
the state transition kernel at time $t$, $Z_t:\Scal\to \Delta(\Ocal)$ denotes the observation function that defines the conditional distribution of the observation given the state at time $t$, $r_t:\Scal\times \Acal \to \mathbb{R}$ 
denotes a bounded reward function that depends on the state-action pair 
at time $t$, and $\gamma\in [0,1)$ is a discount factor that balances the immediate and future rewards. 
$P_t,r_t$ and $Z_t$ are unknown to us and need to be inferred from the observed data. 

\begin{figure}[!t]
	\centering
	\includegraphics[width=\linewidth]{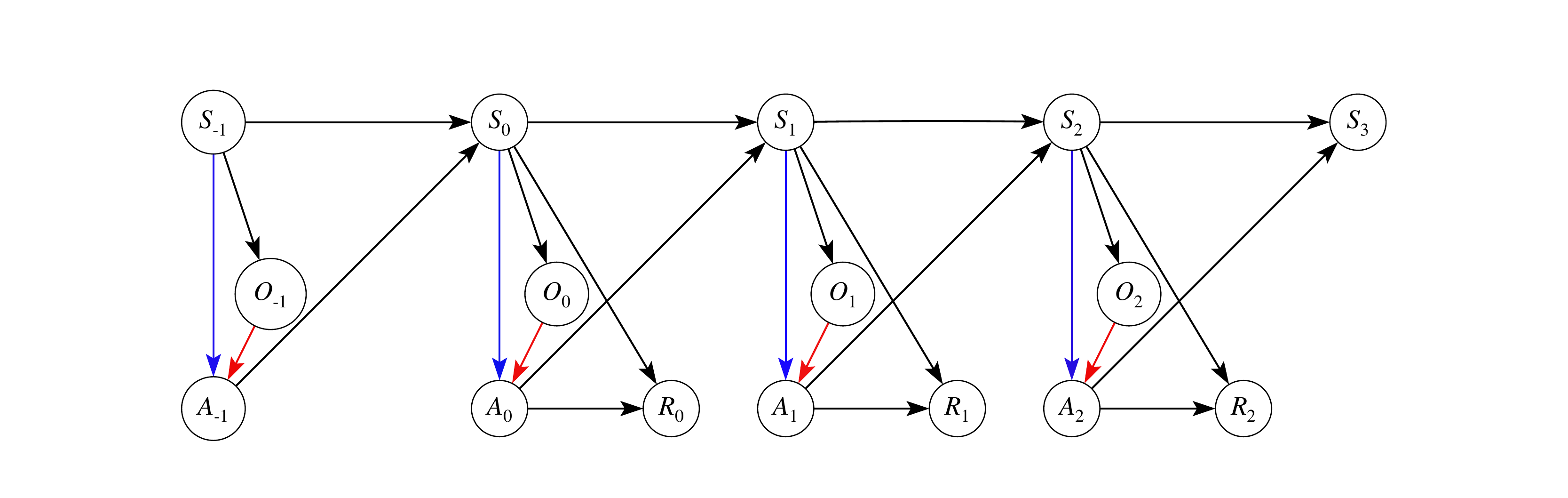}
	\caption{The data generating process in POMDPs. The {\color{red} evaluation policy} in {\color{red}red} depends on the observed variables. The {\color{blue}behavior policy} depends on the hidden state variables. }
	\label{fig:POMDP}
\end{figure}

The data generating process in POMDPs is depicted in Figure \ref{fig:POMDP}. 
Suppose the environment is in a given state
$s\in \Scal$. The agent selects an action $a\in\Acal$. Then the system transits into a new state $s'$ and gives an immediate reward $r(s,a)$ to the agent.
While we cannot directly observe the latent state $s$, we have access to an observation $o\sim Z(\cdot|s)$. 
This model is adopted from \citet{tennenholtz2020off} where the observation does not depend on the previous action. Nonetheless, one may define a
new state/observation vector by concatenating the current state/observation and the past action. The resulting formulation is reduced to the general POMDP setup. A policy is a set of time-dependent decision rules $\pi=\{\pi_t\}$ where each $\pi_t:\Scal\times \Ocal \to \Delta(\Acal)$ maps the state-observation pair to a probability distribution over the action space. For 
a given policy $\pi$, 
the data 
generating process can be summarized as 
$S_0\sim \nu_0,O_0\sim Z_0(\cdot|S_0),A_0\sim \pi_0(\cdot|S_0,O_0),R_0=r_0(S_0,A_0),S_1\sim P_0(\cdot|S_0,A_0),\cdots$ where $\nu_0$ denotes the initial state distribution. 
The observed data up to $t$ is 
given by $\tau_t=(O_0,A_0,R_0,\cdots,O_t,A_t,R_t)$. We denote its distribution by $P_{\pi}$. The policy value $J(\pi)$ is defined as the expected cumulative reward $\E_{\pi}[\sum_{t=0}^{H}\gamma^t R_t]$ for some $0<H\le +\infty$, where the expectation is taken w.r.t.~the distribution of trajectories induced by  policy $\pi$. 

For a given evaluation policy $\pi^e=\{\pi^e_t\}_t$, 
the goal of OPE is to estimate $J(\pi^e)$ from an observational dataset 
generated by a behavior policy $\pi^b=\{\pi^b_t\}_t$. 
Similar to \citet{tennenholtz2020off}, we focus on the setting where $\bpol$ depends only on the latent state and $\epol$ depends only on the observation. See also \pref{fig:POMDP}. 
Under this model assumption, the state variables serve as a confounder between the action and the reward at each time. Since the state is not fully observed, the no unmeasured confounder assumption is violated. 

\begin{table}[!t]
\renewcommand{\arraystretch}{1}
\small
\begin{center}
\begin{tabular}{c|c|c|c}
   $\varepsilon$ & True Value & Ours & Naive \\ \hline
   0.25 & 4.94 & 5.01 & 7.0 \\
   0.5 & 4.99 & 5.008 & 5.004\\
   0.75 & 5.03 & 5.07 & 3.008\\
\end{tabular}
\end{center}
\caption{The proposed OPE estimator and the naive IS or value function-based estimator which replaces the state with the observation when applied to the toy data example with 1e4 trajectories and 100 times points per trajectory. The naive IS and value function-based estimator are computed via minimax weight and Q-function learning \citep{uehara2020minimax}, respectively. The two estimators are the same under the tabular setting and are denoted by ``Naive". \label{tab:toy}
}
\end{table}

Finally, we illustrate the challenge of OPE with latent confounders. If we were to observe the latent state, following the standard OPE methods, we could identify the policy value using the marginalized IS or the value function-based estimator. However, since we cannot observe the state, these methods are not applicable. Naively replacing the state with the observation would yield biased estimators. 

To elaborate, we design a toy example with binary observation, state,
and action spaces. Specifically, the target policy is a uniform random policy and 
the behavior policy is given by $\pi_t^b(1|1)=\pi_t^b(0|0)=1-\varepsilon$,  
for some constant $0<\varepsilon<1$ and any $t\ge 0$. 
When $\varepsilon=0.5$, the action is independent of the latent state and no unmeasured confounders exist. The naive IS or value function-based estimator which replaces the state with the observation is expected to be consistent in that case. We 
report the true value (computed via the Monte-Carlo method), our proposed value estimator and the naive estimator when applied to a large dataset with 1e4 trajectories and 100 time points per trajectory in Table \ref{tab:toy}. It can be seen that the naive method works only when $\varepsilon=0.5$, as expected. In contrast, the proposed estimator is consistent in all cases.




\section{Partially Observable Contextual Bandits }\label{sec:bandit}

\begin{figure}[!t]
	\centering
	\begin{subfigure}{.45\linewidth}
		\centering
		\includegraphics[width=\linewidth]{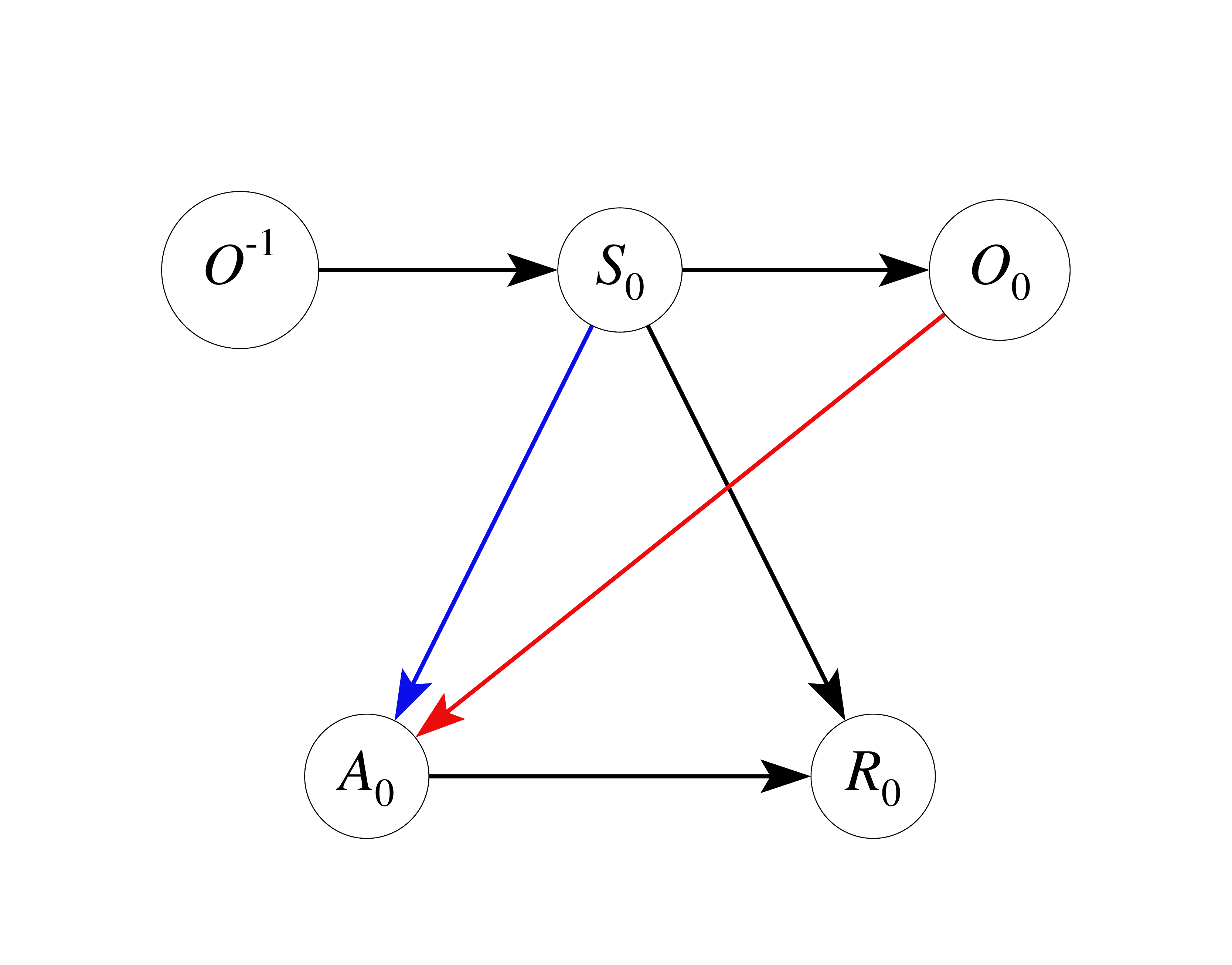}
		\caption{Partially Observable Contextual Bandits}
		\label{fig:POMDP_bandit}
	\end{subfigure}%
	\hspace{0.1em}
	\begin{subfigure}{.53\linewidth}
		\centering
		\includegraphics[width=\linewidth]{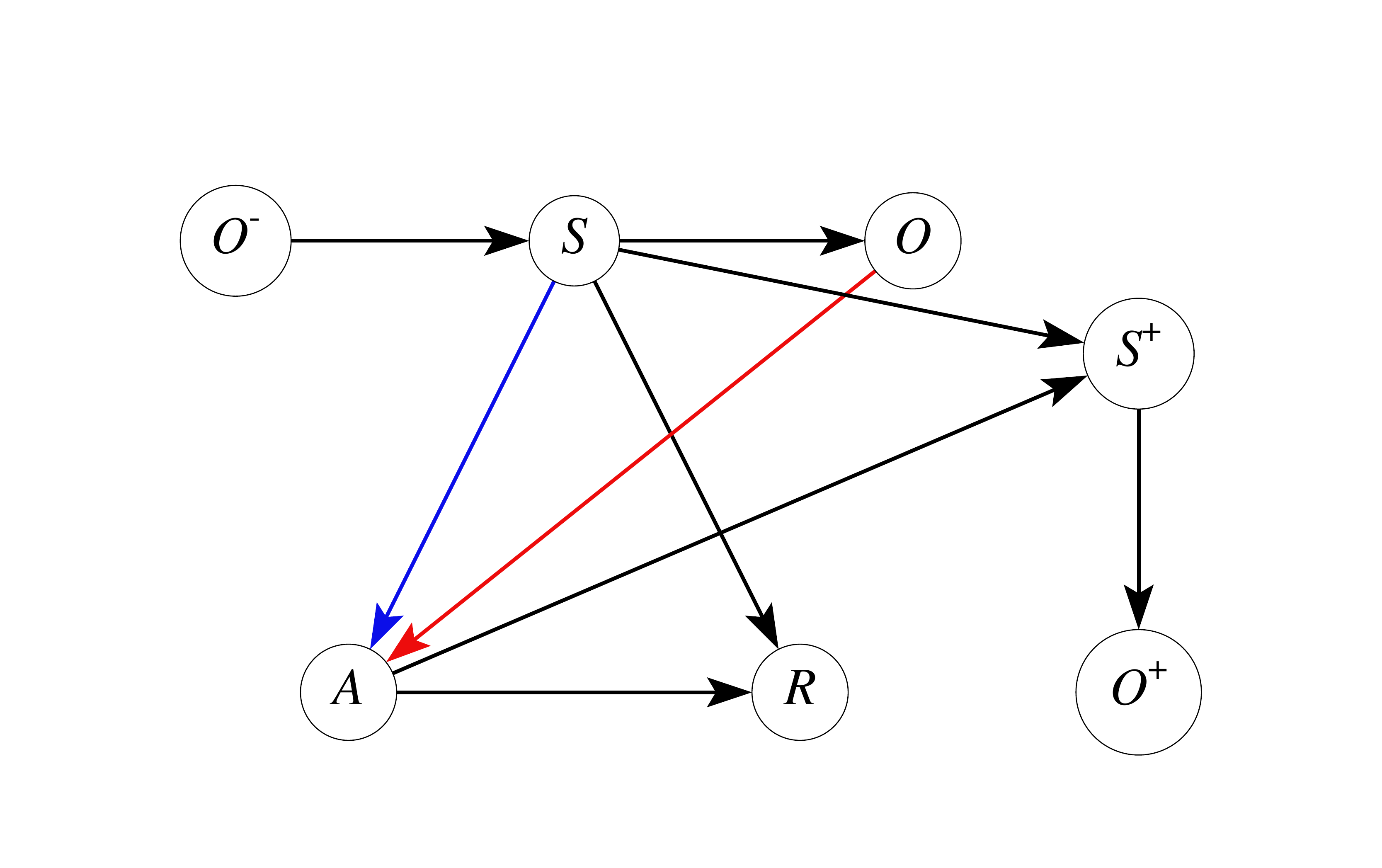}
		\caption{Time-homogeneous POMDPs.} 
	\label{fig:POMDP_homo}
\end{subfigure}
\caption{Red lines depict the dependency of evaluation policies on the observed variables. Blue lines depict the dependence of behavior policies on the state variables. }
\end{figure}

As a warm-up, we start with a partially observable contextual bandit setting (a special case of POMDP with horizon that equals 1) to 
present our main idea. Suppose we have $N$ data tuples that are i.i.d.~$\{O_{-1},A_0,O_0,R_0\}$ tuples where $O_{-1}$ denotes the additional observation that is conditionally independent of $(A_0,O_0,R_0)$ given $S_0$. 
The corresponding Bayesian network is depicted in \pref{fig:POMDP_bandit}. 

The IS and value-function based estimators are given by
\begin{eqnarray*} \textstyle
\EE\bracks{\eta_0(S_0,A_0)R_0}, \EE\left[v_0(S_0)\right]
=\sum_{a} \EE q_0(S_0,a)\pi^e_0(a\mid S_0),
\end{eqnarray*}
respectively, 
where $q_0(S_0,A_0)=\E[R_0\mid S_0,A_0]$,  $\eta_0=\pi^e_0/\pi^b_0$ denotes the IS ratio,  
$\sum_{a}$ is an abbreviation of $\sum_{a \in \Acal}$, and the expectation $\EE$ without any subscript is taken w.r.t. to the distribution of the offline data. As we have commented, these methods are not applicable since $S_0$ is unobserved.


To handle unmeasured confounders, we first assume the existence of certain bridge functions that link the target policy's value and the observed data distribution. 
\begin{assum}[Existence of bridge functions]\label{assum:key}
	There exist functions $b'_V:\Acal\times \Ocal \to \RR,b'_W:\Acal\times \Ocal\to \RR$ that satisfy 
	\begin{eqnarray}\label{eq:bridge}
		\EE[b'_V(A_0,O_0)\mid S_0,A_0] 	= \EE[R_0 \pi_0^e(A_0 \mid O_0)\mid A_0,S_0],\\\label{eq:bridge2}
		\E[b'_W(A_0,O_{-1})\mid S_0,A_0] =1/\pi_0^b(A_0\mid S_0).
	\end{eqnarray}
	We refer to $b'_V(\cdot)$ as a value bridge function and $b'_W(\cdot)$ as a weight bridge function. 
\end{assum}

By definition, weight bridge and value bridge functions can be interpreted as projections of the importance sampling weight (i.e., inverse propensity score) and value functions defined on the latent state space onto the observation space. It is worthwhile to mention that we do \emph{not} require these bridge functions to be \emph{uniquely} defined.
When  $\Scal$ and $\Ocal$ are discrete, the existence of solutions to the integral equations in \pref{eq:bridge} and \pref{eq:bridge2} 
is equivalent to certain matrix rank assumptions; see the assumptions in \citet{nair2021spectral,boots2011closing,hsu2012spectral}. 
These assumptions require observed variables to contain sufficient information about unobserved states. 
Specifically, 
define a matrix $\Omega_a = \Prr_{\bpol}(\bO_0| A_0=a,\bO_{-1})$  whose $(i,j)$-th element is given by $\Prr_{\bpol}(\bO_0=o_i| A_0=a,\bO_{-1}=o_j)$ where $o_i$ denotes the $i$th element in the observation space. 
Then 
they require $\rank(\Omega_a)= |\Scal|$ for any $a\in \Acal$. See Appendix \ref{app:moreass2} for details. When $\Scal$ and $\Ocal$ are continuous, it follows from Picard's theorem that Assumption \pref{assum:key} 
is implied by several conditions \citep[Theorem 2.41]{carrasco2007linear}.
Here, we remark the value bridge function is target policy dependent and it might be more appropriate to write $b_V'$ as $b_{V'}^{\epol}$. Nonetheless, to ease the notation, we will remove the superscript $\epol$ throughout the paper.  

Next, we show that Assumption \pref{assum:key} is a sufficient condition for policy value identification. 
The following theorem shows that if we know $b'_V$ and $b'_W$ in advance, we can consistently estimate the target value $J(\pi^e)$ from the offline data. 

\begin{lemma}[Pseudo identification formula]\label{lem:first_step}
	Suppose Assumption \ref{assum:key} holds. Then, for any $b'_V,b'_W $ that solve \pref{eq:bridge}, 
	\begin{eqnarray}\label{eq:key}
		\begin{split}
		J(\pi^e)=\EE\bracks{\sum_a b'_V(a,O_0)}\quad\hbox{and}\\ J(\pi^e)=\EE[b'_W(A_0,O_{-1})R_0\pi^e_0(A_0\mid O_0)].
		\end{split} 
	\end{eqnarray}
\end{lemma}
These formulas outline an IS estimator and a value function-based estimator for $J(\pi^e)$. It remains to identify the bridge functions $b_V'$ and $b_W'$. However, even if Assumption \ref{assum:key} holds, it is still very challenging to learn $b_V'$ and $b_W'$ from the observed data as their definitions involve the unobserved state $S_0$. As such, we refer to \pref{lem:first_step} as the ``pseudo" identification formula. In the following, we introduce some versions of bridge functions that are identifiable from the observed data. 

\begin{definition}[Learnable bridge functions]\label{def:observed_bridge_bandit}
	The learnable value bridge function $b_V:\Acal\times \Ocal \to \RR$ and learnable weight bridge function $b_W:\Acal\times \Ocal \to \RR$ are solutions to 
	\begin{align}\label{eq:observed}
		\begin{split}
			\hspace{-0.2cm}\E[R_0\epol_0(A_0|O_0)|A_0,O_{-1}] =\E[b_V(A_0,O_0)|A_0,O_{-1}]\\\hbox{and}\quad
			1/P_{\bpol}(A_0|O_0) =  \E[b_W(A_0,O_{-1})|A_0,O_0 ].
		\end{split} 
	\end{align}
\end{definition}

Throughout this paper, we use $b'$ to denote a bridge function and $b$ to denote a \textit{learnable} bridge function. 
The following lemma shows that any bridge function is a learnable bridge function. 
\begin{lemma}\label{lem:observed}
	Bridge functions defined as solutions to \pref{eq:bridge} and  \pref{eq:bridge2} are learnable bridge functions. 
\end{lemma}

We next present an equivalent definition for $b_W$.
This characterization is useful when extending our results to the POMDP setting. 
\begin{lemma}\label{lem:some_characterization}
	$\E[b_W(A_0,O_{-1})\mid A_0,O_0]=1/P_{\bpol}(A_0\mid O_0)$ holds if and only if $$\E[\sum_a f(O_0,a)-b_W(A_0,O_{-1})f(O_0,A_0)]=0,\quad \forall f.$$
\end{lemma}

As we have commented, the class of bridge functions are difficult to estimate from the observed data. On the other hand, the class of learnable bridge functions is identifiable. However, they are \emph{not} necessarily bridge functions. Thus, we cannot invoke \pref{lem:first_step} for value identification. Nonetheless, perhaps surprisingly, the following theorem shows that we can plug-in any learnable bridge function $b_V$ and $b_W$ for $b_V'$ and $b_W'$ in \pref{eq:key} under Assumption \ref{assum:key}.

\begin{theorem}[Key identification formula]\label{thm:final_step}
		Assume the existence of value bridge functions and learnable weight bridge functions. Then, letting $b_W(\cdot)$ be a learnable weight bridge function, we have 
		\begin{align*}
			J(\epol)=\E[b_W(A_0,O_{-1})R_0] . 
		\end{align*}
		Suppose the existence of weight bridge functions and learnable value bridge functions. Then, letting $b_V(\cdot)$ be a learnable value bridge function, we have
		\begin{align*}
			J(\epol)=\E[\sum_a b_V(a,O_0) ] . 
		\end{align*}
\end{theorem}

We make a few remarks. First, the assumptions in the first and second statements do not imply each other. In both cases, Assumption \ref{assum:key}, the existence of bridge functions, plays a critical role in ensuring the validity of the above equations. Specifically, when Assumption \ref{assum:key} holds, the assumptions in both the first and second statements are automatically satisfied, since bridge functions are learnable bridge functions.
If we naively replace Assumption \ref{assum:key} with the existence of learnable bridge functions that satisfy \eqref{lem:observed}, then \pref{thm:final_step} is no longer valid. 

Second, Theorem \ref{thm:final_step} implies that we only need bridge functions to exist, but do not need to identify them. As long as they exist, learnable bridge functions can be used for estimation, which are not necessarily bridge functions. 

Third, similar identification methods have been developed in the causal inference literature for evaluating the average treatment effect with double negative controls; see Appendix \ref{sec:related_add}. However, their settings differ from ours. 
For instance, those 
works require policies to be constant functions of negative controls. In contrast, we allow our evaluation policy to depend on the observation which serves as negative controls in their setting.


Finally, \pref{thm:final_step} requires the existence of both the value and the (learnable) weight bridge functions. 
Under certain completeness assumptions, $J(\epol)$ can be uniquely identified by assuming the existence of value bridge functions only, and the discussion is deferred to the end of Section \ref{subsec:identify_homo}. 

\section{OPE in Time-homogeneous POMDPs}\label{sec:OPEtimehomo} 

In this section, we focus on time-homogeneous POMDPs where $P_t,Z_t,r_t,\epol_t,\bpol_t$ are stationary over time. We denote them by $P,Z,r,\epol,\bpol$, respectively. OPE in time-inhomogeneous POMDPs is investigated in Appendices \ref{sec:time_inho} and \ref{sec:history}. Our goal is to estimate the target policy's value in the infinite horizon setting ($H = \infty$). 
Consider a data tuple $(O^{-},S,O,A,R,S^+,O^{+})$ following 
\begin{align*}
	(O^-,S) \sim P_{\bpol}(\cdot), A\sim \bpol(\cdot|S),S^+ \sim P(\cdot|S,A),\\
	R=r(S,A),O^{+}\sim Z(\cdot|S^+), 
\end{align*}
where $P_{\bpol}$ denotes certain distribution over $\Ocal\times \Scal$. The observed data tuple is given by $(O^-,O,A,R,O^+)$. For example, for a data trajectory $\{(O_t,A_t,R_t)\}_{0\le t\le n+1}$ under $\pi_b$, the observations can be summarized as $\{(O_{t-1},O_t,A_t,R_t,O_{t+1})\}_{1\le t\le n}$. The above configuration is satisfied with 
$P_{\bpol}$ equal to the occupancy distribution under $\pi_b$ over $\Ocal\times \Scal$. This conversion from trajectories to tuples is commonly used in the offline RL literature \citep{chen2019information}.   
We assume the initial observation distribution $\nu_{\Ocal}(o)\coloneqq \int Z(o\mid s)\nu(s)\rd(s)$ is known to us. 
If unknown, it can be estimated by the empirical data distribution.
With a slight abuse of notation, we denote the distribution of the 
data as $P_{\bpol}(\cdot)$, e.g., $P_{\bpol}(a\mid s)=\bpol(a\mid s)$. In addition, $P_{\bpol}(s)$ denote the probability mass function (or the probability density function) when the state is discrete (or continuous).


\subsection{Identification}\label{subsec:identify_homo}

First, we extend our definitions of the bridge functions to the POMDP setting and require their existence in the assumption below. Let $w(s)=\sum_{t\ge 0} \gamma^t P_{\epol,t}(s)/P_{\bpol}(s)$ denote the marginal density ratio where $P_{\epol,t}(\cdot)$ denotes the probability density function of $S_t$ under the evaluation policy.  
In addition, let $q^{\epol}(a,s)= \E_{\epol} [\sum_{t\ge 0}\gamma^t R_t|A_0=a,S_0=s]$ denote the Q-function. These functions play an important role in constructing marginalized IS and value function-based estimators in fully observable MDPs. The following bridge functions play similar roles in POMDPs. 

\begin{assum}[Existence of bridge functions]\label{assum:existence}
	There exist value bridge functions $b'_V:\Acal\times \Ocal\to \RR$ and weight bridge functions $b'_W:\Acal\times \Ocal\to \RR $ that satisfy 
	\begin{align*}
		\E[b'_{V}(A_0,O_0)\mid A_0,S_0]& =\E_{\epol}[\sum_t \gamma^t R_t\epol(A_0 \mid O_0)|A_0,S_0], \\ 
		\E[b'_{W}(A,O^{-}) \mid A,S]  &= \frac{ w(S)}{\bpol(A|S)}.
	\end{align*}
\end{assum}
\vspace{-0.2cm}
We make several remarks. First, the existence of $b'_W$ implicitly requires $\bpol(a|s)>0$ and $w(s)<\infty$ for any $a$ and $s$. The latter condition is weaker than requiring $P_{\bpol}(s)>0$ for any $s$. In other words, we do not require the
full coverage assumption. 
Second, when the states are fully observed and $\Scal=\Ocal$, it follows immediately that 
$b'_V(a,s)=q^{\epol}(a,s)\epol(a\mid s)$ and $b'_W(a,s)=w(s)/\bpol(a\mid s)$. Similar to the bandit setting, these bridge functions can thus be interpreted as projections of the value functions and the marginalized IS weights onto the the observation space. Third, we do not require the bridge functions to be uniquely defined. 


Next, we define the learnable bridge functions. They are consistent with those in \pref{def:observed_bridge_bandit} when $\gamma=0$.

\begin{definition}[Learnable bridge functions]
	Learnable value bridge functions $b_V$ are defined as solutions to 
	\begin{eqnarray}\label{eqn:observed_value}
		\begin{split}
		\E[b_{V}(A,O)\mid A,O^{-}] =\E[R \epol(A\mid O)\mid A,O^{-}]\\+  \gamma\E\bracks{\sum_{a'}b_{V}(a',O^{+})\epol(A \mid O) \mid  A,O^{-}}. 
		\end{split}
	\end{eqnarray}
	Learnable weight bridge functions $b_W$ are defined as solutions to 
	\vspace{-0.2cm}
	\begin{align}\label{eq:observed_reward}
		\E[L_{\mathbb{W}}(b_W,f)  ]=0,\quad \forall f:\Ocal\times \Acal \to \RR
	\end{align}
	where 
	$L_{\mathbb{W}}(g,f)$ equals 
	\begin{align*}
		\sum_{a'}\gamma  g(A,O^{-})\epol(A|O)f(O^{+},a')-g(A,O^{-}) f(O,A) \\+(1-\gamma)\E_{\tilde O\sim \nu_{\Ocal}}[\sum_{a'}f(\tilde O,a')].
	\end{align*}
\end{definition}


Finally, we show our key identification formula under Assumption \ref{assum:existence}. It extends \pref{lem:observed} and \pref{thm:final_step} to the POMDP setting.

\begin{theorem}[Key identification theorem]\label{thm:key_rl} 
	~
	\begin{enumerate}[leftmargin=*]\vspace{-0.2cm}
		\item Any bridge function is a learnable bridge function. 
		\item Suppose the existence of the value bridge function and learnable weight bridge function $b_W(\cdot)$. Then,  $$ J(\epol)=\frac{1}{1-\gamma} \E[b_{W}(A, O^{-})R\epol(A\mid O)].$$ 
		\item Suppose the existence of the weight bridge function and learnable value bridge function $b_V(\cdot)$. Then, $$J(\epol)=\E_{\tilde O\sim \nu_{\Ocal}}[\sum_a b_V(a,\tilde O)] .$$
	\end{enumerate}
\end{theorem}
Similar to the bandit setting, any bridge function is a learnable bridge function, but the reverse is not true. However, the target policy's value can be consistently estimated based on any learnable bridge function. In addition, similar to the bandit setting, the existence is equivalent to the rank assumption $\rank(\mathrm{Pr}_{\pi^b}(\mathbf{O}_0 \mid a, \mathbf{O}^-))= |\Scal|$ in tabular case. 

Theorem \ref{thm:key_rl} forms the basis of our estimation procedure. It outlines a marginalized IS estimator and a value function-based estimator for policy evaluation.  We remark that conditions in bullet points 2 and 3 do not imply each other. In addition, when Assumption \pref{assum:existence} holds, 
conditions in 2 and 3 are automatically satisfied. 

To identify policy values, we have so far imposed the following assumptions: (a) learnable value bridge functions exist and weight bridge functions exist; (b) learnable weight functions exist and value bridge functions exist. In this section, we show that the policy value can also be identified with 
the existence of the learnable value bridge functions only. However, in that case, we would need to impose some 
completeness conditions. We summarize the results in the following theorem.

\begin{theorem}[Identification with completeness]\label{thm:without0}
Suppose a learnable value bridge function $b_V$ exists, and a completeness assumption holds:
    \begin{align*}
        \E[g(S,A) \mid A,O^-]=0 \implies g(S,A)=0. 
    \end{align*}
    Then,  we have  $J(\epol) = \EE_{\tilde o\sim \nu_{\Ocal}}[\sum_a b_V(a,\tilde o)].$
\end{theorem}

Compared to \pref{thm:key_rl}, the existence of weight bridge functions is replaced with the completeness assumption.



\subsection{Estimation}\label{subsec:estimation}

We first propose three value estimators given certain consistently estimated learnable bridge functions $\hat b_V$ and $\hat b_W$. 
We next introduce minimax estimation methods for $b_V$ and $b_W$. 

Given some $\hat b_V$ and $\hat b_W$, we can construct the following IS and value function-based estimators accordingly to 
\pref{thm:key_rl}. Specifically, define $\hat J_{\IS}$ and $\hat J_{\VM}$ to be 
\begin{align*}
	\frac{1}{1-\gamma}\E_{\Dcal}[\hat b_{W}(A, O^{-})R\pi^e(A\mid O)]\,\,\hbox{and}\,\,  \E_{\tilde O\sim \nu_{\Ocal}}[\sum_a \hat b_V(a,\tilde O)],
\end{align*}

respectively, where $\E_{\Dcal}$ denotes the empirical average over all observed data tuples. 

In addition, we can combine the aforementioned two estimators for policy evaluation. This yields the following doubly-robust estimator, $\hat J_{\DR}=\E_{\Dcal}[J(\hat b_W,\hat b_V)]$ where
\begin{eqnarray}\label{eqn:Jfg}
\begin{split}	
	J(f,g) \coloneqq \E_{\tilde O\sim \nu_{\Ocal}}[\sum_a g(a,\tilde O)]+ \frac{f(A,O^{-})}{1-\gamma} \\ \times \Big[\{R+\gamma \sum_{a'}g(a',O^{+})\}\epol(A\mid O)-g(A,O)\Big], 
\end{split}	
\end{eqnarray}
for $f:\Acal\times \Ocal \to \RR$ and $g:\Acal\times \Ocal \to \RR$. 
It has the desired doubly-robust property, as shown in Section \ref{sec:theoryDRmethod}. 

In fully observable MDPs, $\hat J_{\IS}$ reduces to the marginalized IS estimator \cite{Liu2018}, and $\hat J_{\DR}$ reduces to the doubly-robust estimator \citep{KallusNathan2019EBtC}. 
We remark that our proposal is not a trivial extension of these existing methods to the POMDP setting. The major challenge lies in developing identification formulas in \pref{thm:key_rl} to correctly identify the target policy's value. These results are not needed in settings without unmeasured confounders.

Next, we present our proposal to estimate $b_V$ and $b_W$. To estimate $b_V$, a key observation is that, it satisfies the following integral equation, $\E[L_{\mathbb{V}}(b_V,f)]=0 $ for any $f$ where
$L_{\mathbb{V}}(g,f)$ is given by
\begin{align*}
	[\{R +  \gamma \sum_{a'}g(a',O^{+})\}\epol(A| O)- g(A,O)]f(A,O^-).
\end{align*}
This motivates us to develop the following minimax learning methods. Specifically, we begin with two function classes $\Vcal,\Vcal^{\dagger}\subset \{\Acal\times \Ocal \to \RR \}$ and  a regularization parameter $\lambda\in \mathbb{R}^+$. We next define the following minimax estimator, 
\begin{align}\label{eqn:hatbV}
	\hat b_V=\argmin_{g\in \Vcal }\max_{f\in \Vcal^{\dagger}}\E_{\Dcal}[L_{\mathbb{V}}(g,f)]-\lambda \E_{\Dcal}[f^2]. 
\end{align}

Here, we use the function class $\Vcal$ to model the oracle value bridge function, and the function class $\Vcal^{\dagger}$ to measure the discrepancy between a given $g\in \Vcal$ and 
the oracle learnable value bridge function. In practice, we can use linear basis functions, neural networks, random forests, reproducing kernel Hilbert spaces (RKHSs), etc., to parameterize these functions. 
In the fully observable MDP setting, the above optimization reduces to Modified Bellman Residual Minimization (MBRM) when $\lambda=0.5$ \cite{antos2008learning} and Minimax Q-learning (MQL) when $\lambda=0$ \cite{uehara2020minimax}. It is worthwhile to mention that fitted Q-iteration \citep[FQI,][]{ernst2005tree}, a popular policy learning method in MDPs, cannot be straightforwardly extended to the POMDP setting. This is because the regression estimator in each iteration of FQI will be biased under the POMDP setting.


Similarly, 
according to \pref{eq:observed_reward}, 
we consider the following estimator for the weight bridge function,
\begin{align}\label{eqn:hatbW}
	\hat b_W=\argmin_{g\in \Wcal }\max_{f\in \Wcal^{\dagger}}\E_{\Dcal}[L_{\mathbb{W}}(g,f)]-\lambda \E_{\Dcal}[f^2],
\end{align}
for some function classes $\Wcal,\Wcal^{\dagger} \subset \{\Acal\times \Ocal \to \RR \}$ 
and some $\lambda\in \mathbb{R}^+$.

In practice, we recommend to use linear models or RKHSs to parametrize $V^{\dagger}$ and $\Wcal^{\dagger}$. This allows us to get the closed-form expression for the inner maximization. 
To the contrary, $\Vcal$ and $\Wcal$ can be parametrized by any function classes such as neural networks. When $\Vcal$ and $\Wcal$ are linear models or RKHSs as well, we can obtain the complete closed-form solution for $\hat b_W$. We discuss this further in Appendix \ref{app:morePOMDP}.



\subsection{Theoretical Results for Minimax Estimators} \label{subsec:theory}

We first investigate the nonasymptotic properties of the value function-based estimator $\hat J_{\VM}$. 
We next study the asymptotic property of the DR estimator $\hat J_{\DR}$. Results of the importance sampling estimator can be similarly derived. We discuss this in Appendix \ref{app:morePOMDP}. 
Our results extend some recently established OPE theories in MDPs \citep{munos2008finite,chen2019information,KallusNathan2019EBtC,uehara2021finite} to the POMDP setting. We assume the observed dataset $\mathcal{D}$ consist of $n$ i.i.d.~copies of $(O^-, O, A, R, O^+)$. This i.i.d.~assumption is commonly employed in the RL literature to simplify the proof \citep[see e.g.,][]{dai2020coindice}.


\subsubsection{Value function-based  methods}\label{subsec:theory_value}

We begin with the value function-based estimator. 
First, to measure the discrepancy between the estimated learnable value bridge function $\hat b_V$ and $b_V$, 
we introduce the Bellman residual error for POMDPs as follows: 
\begin{definition}
	The Bellman residual operator $\Tcal$
	maps a given function $f: \Acal\times \Ocal\to \RR$ to another function $\Tcal f$
	such that
	\begin{align*}
		\Tcal f(a,o)=\E[R\epol(A\mid O)+\gamma  \sum_{a'}f(a',O^+)\epol(A\mid O)\\-f(A,O)\mid A=a,O^-=o],\quad \forall a,o.
	\end{align*}
	The 
	bellman residual error is defined as $\E[\{\Tcal f(A,O^{-})\}^2]$.  
\end{definition}

By definition, the Bellman residual error is zero for any learnable bridge function. As such, it 
quantifies how a given function $f$ deviates from the oracle value bridge functions. In MDPs, it reduces to the standard Bellman residual error.

We next establish the rate of convergence of $\hat b_V$ in the following theorem. 
\begin{theorem}[Convergence rate of $\hat b_V$]\label{thm:con_results}
	Set $\lambda>0$ in \pref{eqn:hatbV}. 
	Suppose (a) there exists certain learnable bridge function that satisfies \pref{eqn:observed_value}; 
	(b) $\Vcal$ and $\Vcal^{\dagger}$ are finite hypothesis classes; (c) $\Vcal$ contains at least one learnable bridge function; (d) $\Tcal \Vcal\subset \Vcal^{\dagger}$; (e) There exist some constants $C_{\Vcal}$ and $C_{\Vcal^{\dagger}}$ such that $\forall v\in \Vcal,\|v\|_{\infty}\leq C_{\Vcal}$ and $\forall v\in \Vcal^{\dagger},\|v\|_{\infty}\leq C_{\Vcal^{\dagger}}$ Then, there exists some universal constant $c>0$ such that for any $\delta>0$, with probability $1-\delta$, $\E[\{\Tcal \hat b_V(A,O^{-})\}^2]^{1/2}$ is upper bounded by \vspace{-0.2cm}
	\begin{align*}
		c \max(1,C_{\Vcal},C_{\Vcal^{\dagger}})\sqrt{\log(|\Vcal||\Vcal^{\dagger}|c/\delta)/n}.
	\end{align*} 
\end{theorem}

It is important to note we only assume the existence of the learnable value bridge function in (a). 
We do not impose any assumptions on weight bridge functions. 
Assumption (b) can be further 
relaxed by assuming that $\Vcal$ and $\Vcal^{\dagger}$ are general hypothesis classes. In that case, the convergence rate will be characterized by the critical radii of function classes constructed by $\Vcal$ and $\Vcal^{\dagger}$. 
See e.g., \citet{uehara2021finite} for details.
Assumption (c) is the realizability assumption. Since learnable bridge functions are not unique, we only require $\Vcal$ to contain one of them. 
Assumption (d) is the (Bellman) closedness assumption. It requires that the discriminator class $\Vcal^\dagger$ is sufficiently rich and the operator $\Bcal$ is sufficiently smooth.  
These assumptions are valid in several examples, including tabular and linear models. To save space, we relegate the related discussions to Appendix \ref{app:morePOMDP}. 

Next, we derive the convergence guarantee of the policy value estimator.  

\begin{theorem}[Convergence rate of $\hat J_{\mathrm{VM}}$]\label{thm:final_policy}
	Suppose the weight bridge functions $b'_W(\cdot)$ and learnable value bridge functions exist, and Assumption (a)-(e) in \pref{thm:con_results} hold. 
	Then, there exists some universal constant $c>0$ such that for any $\delta>0$, with probability $1-\delta$, $| J(\epol)-\hat J_{\mathrm{VM}}|$ is upper bounded by
	\begin{align*}\textstyle
		\frac{c}{1-\gamma}\{\max(1,C_{\Vcal},C_{\Vcal^{\dagger}})\E[b'^2_W(A,O^{-})]\}^{\frac{1}{2}}\sqrt{\frac{\log(|\Vcal||\Vcal^{\dagger}|c/\delta)}{n}}.
	\end{align*}
\end{theorem}
Theorem \ref{thm:final_policy} requires a stronger condition than Theorem \ref{thm:con_results}, as we assume the existence of weight bridge function. 
Its proof relies on the following key equation,
\begin{align*}
	|J(\epol)-\hat J_{\mathrm{VM}}|\leq \E[b'^2_W(A,O^{-})]^{1/2}\E[\{\Tcal \hat b_V(A,O^{-})\}^2]^{1/2},
\end{align*}
where the upper bound for the second term on the right-hand-side is given in Theorem \ref{thm:con_results}. 

Notice that Theorem \ref{thm:con_results} relies on the closedness assumption, e.g., Assumption (d). 
We remark that this condition is \emph{not} necessary to derive the convergence rate of the final policy value estimator. In Theorem \ref{thm:without} (see Appendix \ref{app:morePOMDP}), we show that when $b_V \in \Vcal,b_W \in \Vcal^{\dagger}$, 
similar results can be established \emph{without} any closedness conditions.




\subsubsection{DR methods}\label{sec:theoryDRmethod}
We focus on the DR estimator in this section. We first establish its doubly-robustness property  in Theorem \ref{thm:dr_infinite}. It implies that as long as either $\hat b_W$ or $\hat b_V$ is consistent, the final estimator $\hat J_{\DR}$ is consistent.
\begin{theorem}[Doubly-robustness property]\label{thm:dr_infinite}
Suppose Assumption \ref{assum:existence} holds. 
Then $J(f,g)$ defined in \pref{eqn:Jfg} equals $J(\epol)$ as long as either $f=b_W$ or $g=b_V$. 
\end{theorem}



We next show that $\hat J_{\mathrm{DR}}$ is efficient in the sense that it is asymptotically normal with asymptotic variance equal to the Cram\'er-Rao Lower Bound. 


First, we provide the Cram\'er-Rao Lower Bound. To simplify the technical proof, we focus on the tabular setting. Nonetheless, we conjecture that the same result still holds in the non-tabular setting as well. We leave further investigation of this conjecture to future work.
\begin{theorem}\label{thm:cramer}
Suppose $\Scal,\Acal,\Ocal$ are finite discrete spaces. Assume $\rank(\Pr_{\bpol}(\bO \mid A=a, \bO^{-}))=|\Scal|=|\Ocal|$. The Cram\'er-Rao Lower Bound is given by
\begin{align*}
		V_{\mathrm{EIF}} &=\E[ J(b_W, b_V)^2]. 
\end{align*}
\end{theorem}

We impose a rank assumption in Theorem \ref{thm:cramer}. This condition implies that the bridge functions are uniquely defined, and so are the learnable bridge functions. We remark that without such a uniqueness assumption, the Cram\'er-Rao Lower Bound is not well-defined. 





Next, we analyze the property of DR estimator. To avoid imposing certain Donsker conditions \citep{VaartA.W.vander1998As}, 
we focus on a sample-split version of $\hat J_{\mathrm{DR}}$ as in \citet{ZhengWenjing2011CTME}. It splits all data trajectories in $\Dcal$ into two independent subsets $\Dcal_1,\Dcal_2$, computes the estimated bridge function $\hat b^{(1)}_V,\hat b^{(1)}_W$ ($\hat b^{(2)}_V,\hat b^{(2)}_W$) based on the data subset in $\Dcal_2$ ($\Dcal_1$), and does the estimation of the value based on the remaining dataset. Finally, we aggregate the
resulting two estimates to get full efficiency. This yields the following DR estimator, 
\begin{align*}
\hat J^{*}_{\mathrm{DR}}=0.5 \E_{\Dcal_1}[ J(\hat b^{(1)}_W,\hat b^{(1)}_V)]+0.5 \E_{\Dcal_2}[ J(\hat b^{(2)}_W,\hat b^{(2)}_V)]. 
\end{align*}
We summarize the results in the following theorem. Notice that they are valid in the non-tabular setting as well. 

\begin{theorem}\label{thm:efficiency}
Assume the existence and uniqueness of bridge and learnable bridge functions. Suppose $\hat b_W,b_W,\hat b_V,b_V$ are uniformly bounded by some constants, $\E[\{\hat b^{(j)}_W- b_W\}^2(A,O^{-})]=o_p(1),\E[\{\Tcal \hat b^{(j)}_V(A,O^{-})\}^2]=o_p(1), \\ \E[ \{\sum_a\hat b^{(j)}_V(a,O)-b_V(a,O) \}^2 ]^{1/2}=o_p(1),\E[\{\hat b^{(j)}_W- b_W\}^2(A,O^{-})]^{1/2}\E[\{\Tcal \hat b^{(j)}_V(A,O^{-})\}^2]^{1/2}=o_p(n^{-1/2})$ for $j=1,2$. Then, $\sqrt{n}(\hat J^{*}_{\DR}-J(\epol))$ weakly converges to a normal distribution with mean 0 and variance $V_{\mathrm{EIF}}$. 
\end{theorem} 
We again, make a few remarks. First, although the estimated bridge functions are required to satisfy certain nonparametric rates only, the resulting value estimator achieves a parametric rate of convergence (e.g. $n^{-1/2}$). This is essentially due to the doubly-robustness property, which ensures that the bias of the estimator can be represented as a product of the
difference between the two estimated bridge functions and their oracle values. 
Second, \pref{thm:efficiency} derives the asymptotic variance of $\hat J^{*}_{\DR}$, which can be consistently estimated from the data. 
This allows us to perform a handy Wald-type hypothesis testing.  
Finally, \pref{thm:efficiency}
requires a stronger assumption, i.e., the uniqueness of bridge functions, which implies that $|\Scal|=|\Ocal|$. Without this assumption, it remains unclear how to define the $L_2$ error of the estimated bridge function. However, we would like to remark that imposing the uniqueness condition is not a weakness of our analysis, but the CR lower bound statement we hope to prove does not make sense if the condition fails. 

\input{./time_inhomogeneous}
\section{Experiments}
In this section, we evaluate the empirical performance of our method using two synthetic datasets. In both datasets, the state spaces are continuous. Hence, existing POMDP evaluation methods such as \citet{tennenholtz2020off} and \citet{nair2021spectral} are not directly applicable. The discounted factor $\gamma$ is fixed to 0.95 in all experiments. 

\subsection{One-Dimensional Dynamic Process}\label{sec:1dcon}
\noindent \textbf{Environment} ~We first consider a simple dynamic process with a one-dimensional continuous state space and binary actions. Similar environments have been considered in the literature \citep[see e.g.,][]{shi2022dynamic}. 
The initial state distribution, reward function and state transition are given by
\begin{eqnarray*}
S_0\sim \mathcal{N}(0,0.5^2),\quad R_{t} = S_{t} + 2A_{t-1}-1,\\ S_{t+1} = 0.5 S_t + (2A_t - 1) + \mathcal{N}(0, 0.5^2),
\end{eqnarray*}
respectively. The observation is generated according to the additive noise model, $O_t=S_t+N(0,\sigma_O^2)$. Here, the variance parameter $\sigma_O$ characterizes the degree of partial observability and hence the degree of unmeasured confounding. In the extreme case where $\sigma_O=0$, the states become fully-observable and no unmeasured confounders exist. We set the behavior and target policies to be sigmoid functions of the state and the observation, respectively,
\begin{eqnarray*}
    \bpol(1|s)=\frac{1}{1+\exp(s+1)}, \epol_w(1|o)=\frac{1}{1+\exp(wo+1)},
\end{eqnarray*}
for $w\in \{-3,-2,1,2\}$. 

\begin{figure}[t]
    \centering
    \includegraphics[scale=0.25]{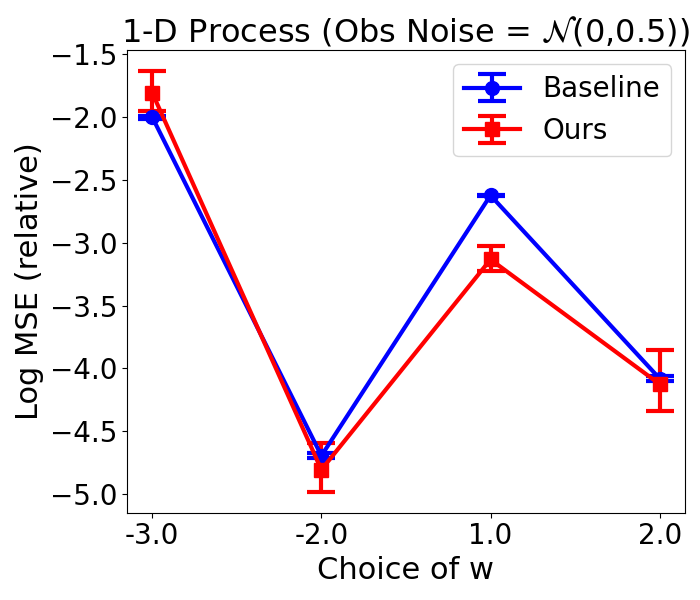}
    \includegraphics[scale=0.25]{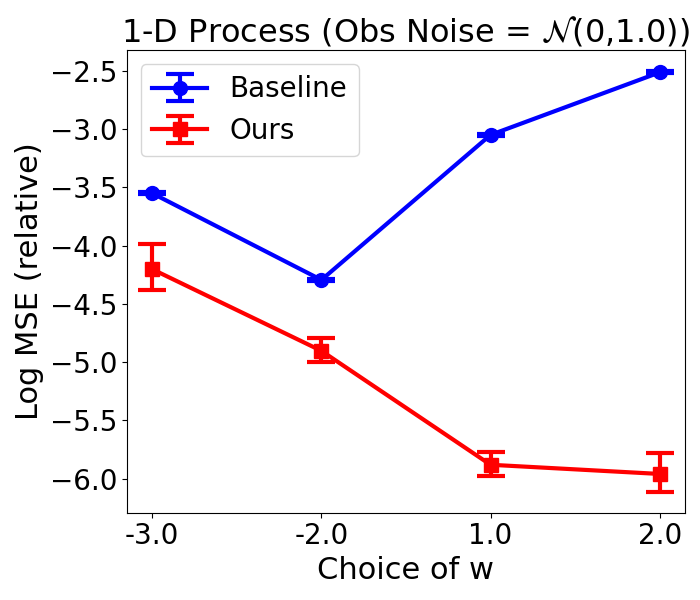}
    \includegraphics[scale=0.25]{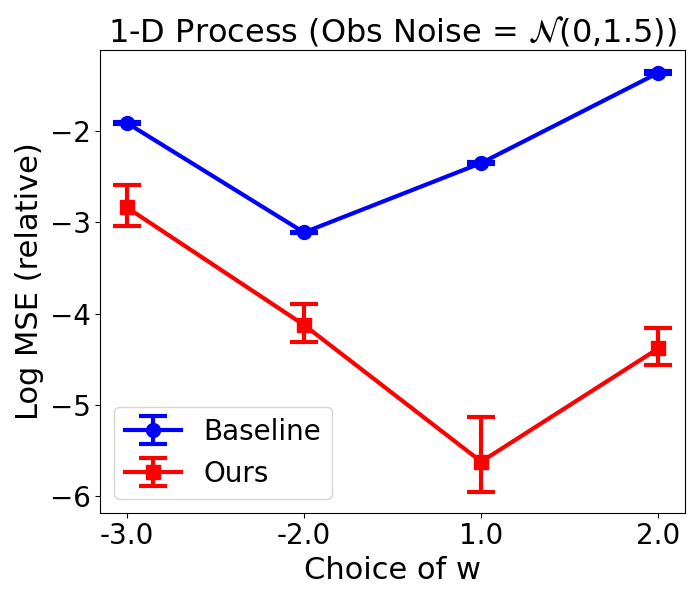}
    \includegraphics[width=11.67cm,height=4cm]{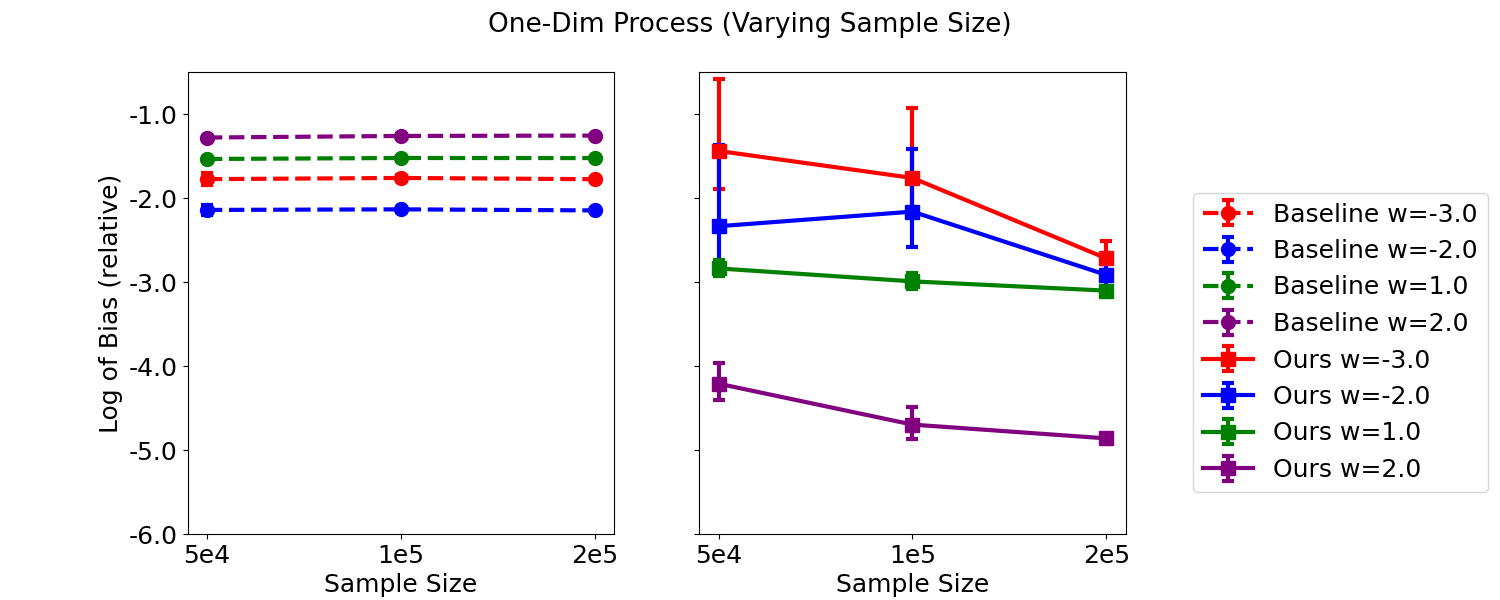}
    \caption{Logarithms of relative biases and mean squared errors (see Appendix \ref{app:moreexp} for the detailed definitions) of the proposed and the baseline methods, with difference choices of $w$ and sample sizes. Confidence intervals are calculated from 10 simulations. Upper panels: sample size equals 2e5. $\sigma_O$ equals $0.5$, $1.0$ and $1.5$, from left to right. Bottom panels: $\sigma_O$ is fixed to 1.}
    \label{fig:Toy_Varying_NoiseLevel}
    \vspace{-0.3cm}
\end{figure}

\begin{figure}[t]
    \centering
    \includegraphics[scale=0.3]{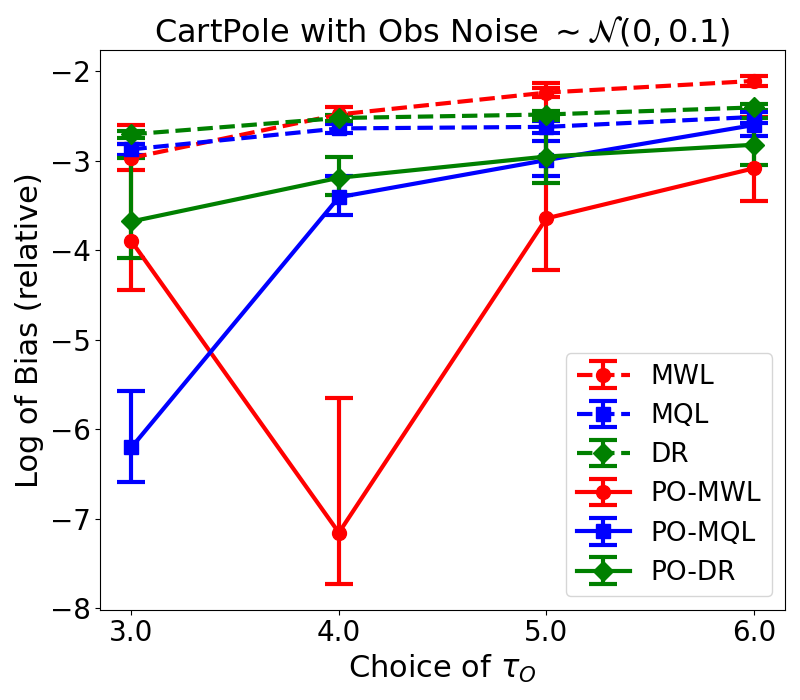}
    \includegraphics[scale=0.3]{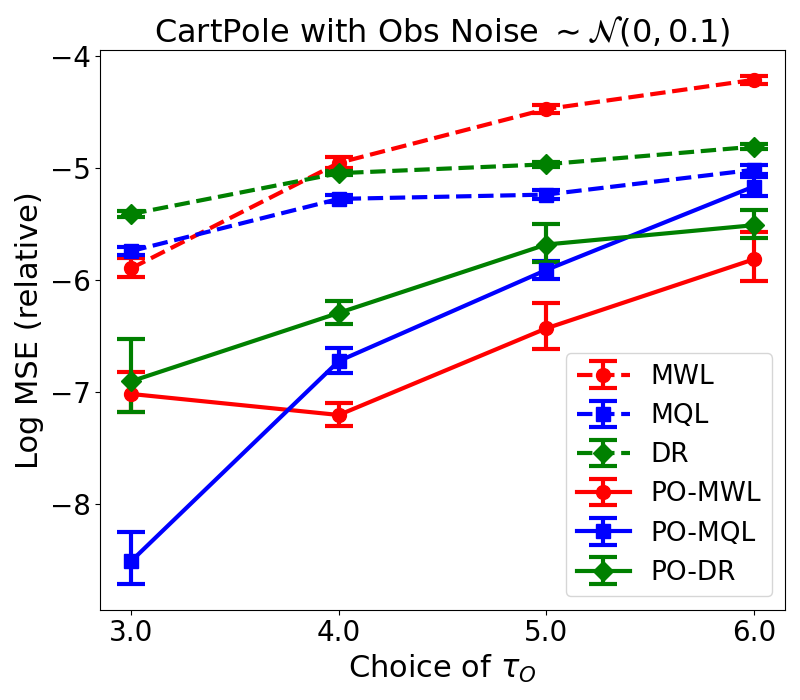}
    \includegraphics[scale=0.35]{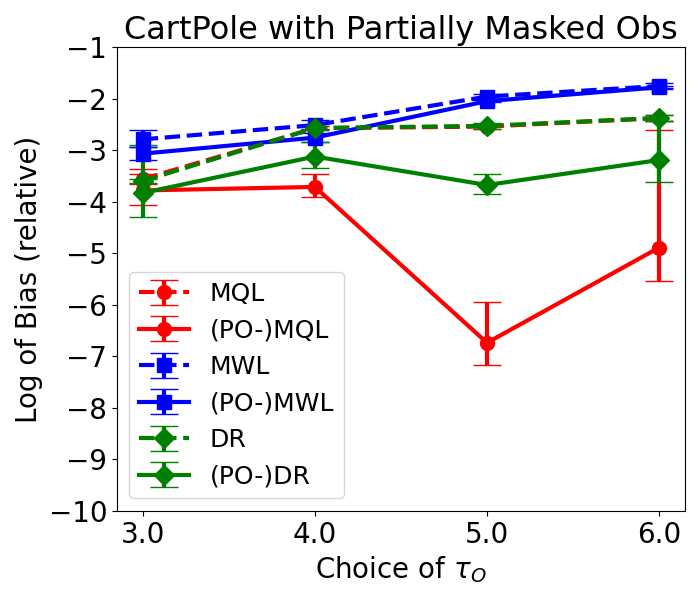}
    \includegraphics[scale=0.35]{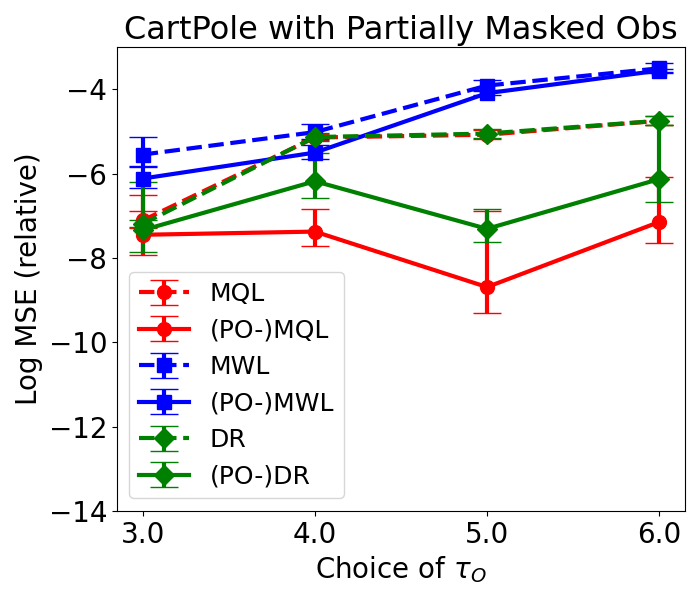}
    \caption{Logarithms of relative biases (left) and MSEs (right) of the proposed (solid lines) and the baseline (dashed lines) estimators and the associated confidence interval, based on 10 simulations, with different choices of the temperature parameter $\tau_O$. Observation in top panels is generated by injecting Gaussian noises whereas the location of the cart is removed from the observation in bottom panels.} \vspace{-0.3cm}
    \label{fig:CartPole_Results}
\end{figure}

\noindent \textbf{Implementation and Baseline Method} ~We use linear basis functions to approximate the bridge functions. In this case, the proposed three estimators ($\hat J_{\VM},\hat J_{\IS},\hat J_{\DR}$) coincide with each other \citep[see a similar phenomenon in][]{uehara2020minimax}, so we compute the value function-based estimator only. We compare our estimator to the standard linear value function-based estimator that assumes no unmeasured confounders, i.e., LSTDQ \citep{lagoudakis2003least}. 
See additional implementation details  in Appendix \ref{app:moreexp}. 


\noindent \textbf{Results} ~For each simulation, we first generate 2000 trajectories with 100 time points per trajectory. This yields a total of 2e5 observations. Reported in the upper panels of Figure \ref{fig:Toy_Varying_NoiseLevel} are the logarithms of relative mean squared errors of the proposed and baseline methods, with different choices of $\omega$ and $\sigma_O$. We make a few observations. First, when $\sigma_O\ge 1$, the proposed estimator is significantly better than the baseline estimator. 
Second, when $\sigma_O=0.5$, the two methods perform comparably. As we have commented, $\sigma_O$ measures the degree of unmeasured confounding. For moderately large $\sigma_O$, the baseline estimator cannot handle unmeasured confounders, yielding a biased estimator. In contrast, a smaller $\sigma_O$ produces a less confounded dataset. The two estimators thus achieve similar performance. 

We next fix $\sigma_O$ to 1.0, vary the number of trajectories generated in each simulation, and report the corresponding 
relative biases and mean squared errors of the proposed and baseline methods as well as the associated confidence intervals in the bottom panels of Figure \ref{fig:Toy_Varying_NoiseLevel} and Figure \ref{fig:Toy_Varying_SampleSizeMSE} (see Appendix \ref{app:moreexp}). It can be seen that the baseline estimator suffers from a large bias. Their bias and MSE are constant as functions of sample size. To the contrary, the bias and MSE of the proposed estimator generally decrease with the sample size, demonstrating its consistency. 


\subsection{CartPole}
We next consider the CartPole environment from the OpenAI Gym environment \citep{brockman2016openai}. 
The state variables are 4-dimensional and fully-observable. To create partially observable environments, we 
simulate observations by 
either adding independent Gaussian noises to each dimension of the states, i.e., $O^{(j)}=S^{(j)}(1+\mathcal{N}(0,0.1^2))$, $1\le j\le 4$, or removing the location
of the cart from the observation. To generate data, we apply DQN \citep{mnih2015human} to the data that include latent states instead of observations, and set the behavior policy to a softmax policy based on the estimated Q-function. The evaluation policy is set to another softmax policy based on DQN applied to the observational data (i.e., no latent states) with the temperature parameter given by $\tau_O$. For each simulation, we collect the dataset according to the behavior policy until the sample size reaches 2e5. We consider three baseline estimators, corresponding to the minimax Q-learning (MQL), minimax weight learning (MWL) and DR estimators \citep{uehara2020minimax}. These estimators cannot handle unmeasured confounders. The proposed value function-based, marginalized IS and DR estimators are denoted by PO-MQL, PO-MWL and PO-DR, respectively. We parametrize the bridge functions using a two-layer neural network, and set the function spaces $\Vcal^{\dagger}$ and $\Wcal^{\dagger}$ to RKHSs to facilitate the computation. Some additional details about the environment and implementation are given in Appendix \ref{app:moreexp}. Results are reported in Figures \ref{fig:CartPole_Results}. It can be seen that the proposed estimator achieves better performance in all cases.


\section{Conclusion}

We study OPE on POMDPs where behavior policies depend on unobserved state variables.  We propose a
novel identification method for the target policy’s value in the presence of unmeasured confounders. Our proposal
only relies on the existence of bridge functions. We further propose minimax learning methods for computing these estimated bridge functions that can be naturally coupled with function approximation to handle a continuous or large state/observation space. We also develop three types of policy value estimators based on the bridge function estimators 
and provide their nonasymptotic and asymptotic properties. 
In \pref{sec:related_add}, we discuss the different between our proposal and a highly-related concurrent work by \citet{bennett2021}.

\section*{Acknowledgements}
CS's research was partly supported by the EPSRC grant EP/W014971/1. MU was partly supported by MASASON Foundation. NJ's last involvement in the project was in December 2021, and acknowledges funding support from ARL Cooperative Agreement W911NF-17-2-0196, NSF IIS-2112471, NSF CAREER award, and Adobe Data Science Research Award. 
\bibliography{rl}



\newpage
\appendix
\onecolumn

\section{Additional Related Works}\label{sec:related_add}

We discuss some additional related works in this section. 

\paragraph{Minimax Estimation} 
Our proposal defines the value and weight bridge functions as solutions to some integral equations. Nonparametrically estimating these bridge functions 
is closely related to nonparametric instrumental variables (NPIV) estimation. In NPIV estimation, the standard regression estimator is biased and variants of minimax estimation methods have been developed to correct the bias \citep{DikkalaNishanth2020MEoC,muandet2019dual,pmlr-v70-hartford17a,bennett2020variational}.
In the RL literature, the minimax learning method has also been widely used for OPE without unmeasured confounders \cite{antos2008learning,chen2019information,nachum2019dualdice,feng2019kernel,uehara2020minimax}. 

\paragraph{Negative Controls} 
Our proposal is closely related to a line of research on developing causal inference methods to evaluate the average treatment effect (ATE) with double-negative control adjustment. Among those works, \citet{miao2018identifying} outlined a consistent estimator for ATE with a categorical confounder. More recent methods generalize their approach to the continuous confounder setting \cite{deaner2018proxy,cui2020semiparametric,singh2020kernel,ghassami2021minimax,xu2021deep,mastouri2021proximal}. All the aforementioned methods focus on a contextual bandit setting (i.e., one-shot decision making)
and are not directly applicable to POMDPs that involve sequential decision making.

\paragraph{Comparison with \citet{bennett2021}}

The difference between the work of \citet{bennett2021} and our proposal can be summarized as follows. 
Methodologically, the IS, VM, DR estimators developed in the two papers are \emph{different}. 
First, 
consider the IS estimator as an example. 
In fully-observable MDPs, 
our proposed estimator is reduced to marginal IS estimators \citep{Liu2018,XieTengyang2019OOEf} whereas the IS estimator developed by \citet{bennett2021} is reduced to the cumulative IS estimator \citep{precup2000eligibility}.
Second, \citet{bennett2021} do not introduce value bridge functions as in our paper. Instead, they introduce an outcome bridge function to approximate the reward function at each time step $0\leq t\leq H-1$. Because of the difference between these definitions, 
their estimating procedure for the outcome bridge functions involves the estimated weight bridge functions. 
In contrast, the estimation of value bridge functions in our paper is agnostic to the estimation of weight bridge function. 

Theoretically, we focus on the derivation of finite-sample error bounds. To the contrary, \citet{bennett2021} focus on establishing asymptotic properties. In addition, the efficiency bound they developed differs from ours (see Theorem \ref{thm:cramer}). Specifically, they focused on the non-tabular case whereas our bounded is limited to the tabular setting. Moreover, since we focus on evaluating Markovian and stationary policies, in fully-observable MDPs, our bounds 
are reduced to those in \citet{KallusNathan2019EBtC} whereas their bound is reduced to the one in \citet{jiang2016doubly}.

\section{Some Additional Details Regarding OPE in Partially Observable Contextual Bandits}\label{app:moreass2}

We specialize our results in Section \ref{sec:bandit} to tabular settings. We use the following notation. Let $X,Y$ be random variables taking values $\{x_1,\cdots,x_n\}$ and $\{y_1,\cdots,y_m\}$, respectively. Then we denote a $n\times m$ matrix with elements $\mathrm{Pr}(x_i\mid y_j)$ by $\mathrm{Pr}(\mathbf{X}\mid \mathbf{Y})$. Similarly, $\Prr(\mathbf{X})$ denotes a $n$-dimensional vector with elements $x_i$. We denote the Moore-penrose inverse of $\mathrm{Pr}(\mathbf{X}\mid \mathbf{Y})$ by $\mathrm{Pr}(\mathbf{X}\mid \mathbf{Y})^+$. 

We next introduce the following assumption.
\begin{assum}\label{asm:weak1}
	Suppose $\rank(\Prr_{\bpol}(\bO_0|\bS_0))= |\Scal|$ and $\rank(\Prr_{\bpol}(\bO_{-1}|\bS_0))= |\Scal|$. 
\end{assum}

As we have mentioned, Assumption \pref{asm:weak1} implies that $|\mathcal{O}|\ge |\mathcal{S}|$ and is weaker than the condition in \citet{tennenholtz2020off} that requires the state and observation spaces have the same cardinality. 
The following lemma states that \pref{asm:weak1} is equivalent to Assumption 2 in \citet{nair2021spectral}. 
\begin{lemma}\label{lem:equivalence}
Suppose overlap conditions: 
\begin{align*}
     \Prr_{\bpol}(O^- = o^- \mid A_0= a)>0,\quad \Prr_{\bpol}(O^- = o^- \mid A_0= a)>0 
\end{align*}
for any $(s,a,o) \in \Scal\times \Acal \times \Ocal$. Then, Assumption \pref{asm:weak1} holds if and only if 
	$\rank(\Prr_{\bpol}(\bO_0| A_0=a,\bO_{-1}))= |\Scal|$ for any $a\in \Acal$. 
\end{lemma}
Then following \citet{nair2021spectral}, the policy value can be explicitly identified as follows: 
\begin{theorem}[Identification formula in the tabular setting]\label{thm:discrete_identify}
	\begin{align*}
		J(\pi^e)=\sum_{a,r,o}r\pi^e_0(a|o)\Prr_{\bpol}(r,o| \bO_{-1}) \{\Prr_{\bpol}(\bO_0| A_0=a,\bO_{-1})\}^{+}\Prr_{\bpol}(\bO_0). 
	\end{align*}
\end{theorem}


    

\section{Some Additional Details Regarding OPE in Time-Homogeneous POMDPs}\label{app:morePOMDP}
As we have commented, we can use linear basis functions or RKHSs to estimate the bridge functions. Specifically, when $\Vcal^{\dagger}$ or $\Wcal^{\dagger}$ is set to be the unit ball in an RKHS, then there exists a close form expression for the inner maximization problem \citep[see e.g.,][]{Liu2018}. When linear basis functions are used, the proposed minimax optimization has explicit solutions, as we elaborate in the example below.
\begin{example}[Linear models]\label{example:linear_model}
	Let $\Vcal=\Vcal^{\dagger}=\{(a,o)\mapsto \theta^{\top}\phi(a,o):\theta \in \Theta\}$ and $\lambda=0$, where $\phi:\Acal\times \Ocal \to \RR^{d}$ is a feature vector and $\Theta\subset \RR^d$. When the parameter space $\Theta$ is sufficiently large, it is immediate to see that
	\begin{align*}
		\hat b_V = \prns{\E_{\Dcal}[\phi(A,O^{-})\phi^{\top}(A,O)] -\gamma  \E_{\Dcal}\bracks{\sum_{a'}\epol(A\mid O)\phi(A,O^{-})\phi^{\top}(a',O^+)} }^{+}\E_{\Dcal}[R \epol(A\mid O)\phi(A,O^-)]. 
	\end{align*}
	The resulting policy value estimator is given by
	\begin{align*}
		\E_{\Dcal}[\sum_a\phi(a,O)]^{\top}\prns{\E_{\Dcal}[\phi(A,O^{-})\phi^{\top}(A,O)] -\gamma  \E_{\Dcal}\bracks{\sum_{a'}\epol(A\mid O)\phi(A,O^{-})\phi^{\top}(a',O^+)} }^{+}\E_{\Dcal}[R \epol(A\mid O)\phi(A,O^-)]. 
	\end{align*}
	
	Consider the standard MDP setting where $S=O=O^-,S^+=O^+$. By setting  $\Vcal=\{(s,a) \mapsto \theta^{\top}\epol(a\mid s)\phi(a,s)\},\Vcal^{\dagger}=\{(s,a) \mapsto \theta^{\top}\phi(a,s)/\epol(a\mid s)\}$, the above estimator reduces to classical LSTDQ estimator \cite{lagoudakis2003least}: 
	\begin{align*}
		\E_{\Dcal}[\phi(\epol,S)]^{\top}\prns{\E_{\Dcal}[\phi(A,S)\phi^{\top}(A,S)] -\gamma  \E_{\Dcal}\bracks{\phi(A,S)\phi^{\top}(\epol,S^+)} }^{+}\E_{\Dcal}[R \phi(A,S)], 
	\end{align*}
	where $\phi(\epol,s)=\E_{a\sim \epol(s)}[\phi(a,s)]$. 
	
	Alternatively, we may set $\mathcal{V}:= \{(a,o) \rightarrow \pi^e(a|o)\theta^\top\phi(a,o):\theta \in \Theta\}$ instead. As discussed in Section \ref{sec:1dcon}, we find such a parametrization has smaller approximation error in the implementation. The corresponding estimator is given by
    \begin{align*}
    	\E_{O\sim\nu_O,a\sim \pi^e(\cdot|O)}[\phi(a,O)]^{\top}\big(\E_{\Dcal}[&\phi(A,O^{-})\phi^{\top}(A,O)\epol(A\mid O) \\
    	&-\gamma  {\phi(A,O^{-})\epol(A\mid O) \EE_{a'\sim\epol(\cdot|O^+)}[\phi^{\top}(a',O^+)]}]\big)^{+}\E_{\Dcal}[R \epol(A\mid O)\phi(A,O^-)].
    \end{align*}
\end{example}

We next introduce two examples to further elaborate Theorem \ref{thm:con_results}.

\begin{example}[Tabular models]
	Consider the tabular case, and $\Vcal$ and $\Vcal^{\dagger}$ are fully expressive classes, i.e., linear in the one-hot encoding vector over $(\Acal \times \Ocal)$.  Then, $\log(|\Vcal||\Vcal^{\dagger}|)$ is substituted by $|\Ocal||\Acal|$. The Bellman closedness $\Tcal \Vcal\subset \Vcal^{\dagger}$ is satisfied.  Our finite sample result circumvents the potentially complicated matrix concentration argument in POMDPs  \citep[see e.g.,][]{hsu2012spectral} by viewing the problem from a general perspective. 
\end{example}

\begin{example}[Linear models]
	Consider the case where $\Vcal$ and $\Vcal^{\dagger}$ are linear models. Then, $\log(|\Vcal||\Vcal|^{\dagger})$ can be substituted by the dimension of the feature vector $d$. The Bellman closedness assumption is satisfied when $P_{\bpol}(R,O,O^+ \mid A,O^-)$ is linear in $\phi(A,O^-)$. In fully-observable MDPs, this reduces to the linear MDP model \citep{jin2020provably}, i.e.,  $P_{\bpol}(R,S^+ \mid A,S)$ is linear in $\phi(A,S)$. 
\end{example}

The next theorem shows that the finite sample rate of convergence for the value function-based estimator can be established without any closedness assumption.

\begin{theorem}[Convergence rate without closedness]\label{thm:without}
	Set $\lambda$ in equations \ref{eqn:hatbV} and \ref{eqn:hatbW} to zero.
	Suppose (a) Assumption \ref{assum:existence} holds so that the bridge functions exist; (b) $\Vcal$ and $\Vcal^{\dagger}$ are symmetric finite hypothesis classes; (c) $\Vcal$ contains certain learnable reward bridge function; (d) $\Vcal^{\dagger}$ contains learnable weight bridge function; (e) $\forall v\in \Vcal,\|v\|_{\infty}\leq C_{\Vcal}$ and $\forall v\in \Vcal^{\dagger},\|v\|_{\infty}\leq C_{\Vcal^{\dagger}}$. Then, with probability $1-\delta$, 
	\begin{align*}
		|J(\epol)-\hat J_{\mathrm{VM}}| \leq   c \max(1,C_{\Vcal},C_{\Vcal^{\dagger}})^2(1-\gamma)^{-2}\sqrt{\log(|\Vcal||\Vcal^{\dagger}|c/\delta)/n},
	\end{align*}
	for some universal constant $c>0$. 
\end{theorem}

Next, we discuss the theoretical properties of the IS estimator. To establish its convergence rate, we can define the adjoint Bellman residual operator $\Tcal'$ as in \cite{uehara2021finite}, and derive the convergence rate of $\E[\{\Tcal'\hat b_W(A,O^{+})\}^2]^{1/2}$ under realizability ($b_W\in \Wcal$) and (adjoint) closedness ($\Tcal' \Wcal\subset \Wcal^{\dagger}$). Similar to the proof of \pref{thm:final_policy}, this (adjoint) Bellman residual error can be translated into the error of the estimated policy value. Without the closedness assumption, similar results can be derived when $b_W \in \Wcal$ and $b_V \in \Wcal^{\dagger}$. 



\section{Experiments}\label{app:moreexp}
\paragraph{Measure of Estimation Error}
Given $n$ datasets $D_1,D_2,...,D_n$, the estimators computed based on each dataset $\hat{V}_1,...,\hat{V}_n$, and the true value $V$, we define the relative bias to be
\begin{align*}
    |\frac{1}{n}\sum_{i=1}^n \frac{\hat{V}_i}{V}-1|.
\end{align*}
Define the relative mean squared error to be:
\begin{align*}
    |\frac{1}{n}\sum_{i=1}^n\Big(\frac{\hat{V}_i-V}{V}\Big)^2|.
\end{align*}
In our experiments, we use the above two definitions to measure the estimation error of different estimators.

\paragraph{Additional Details for the 1d Continuous Dynamic Process Example}
Notice that the definition of the bridge value function involves the evaluation policy. Instead of directly using linear models to parametrize $b_V$, we set the function space to $\mathcal{V}:= \{(a,o) \rightarrow \pi^e(a|o)\theta^\top\phi(a,o):\theta \in \Theta\}$. We find that such a parametrization has smaller approximation error. The closed-form expression of the resulting estimator is given in Section \ref{app:morePOMDP}. We use the Python function RBFsampler 
to generate random Fourier features. To mitigate the randomness arising from the features, for each dataset and each method, we use 5 different random seeds to generate 5 sets of RBF features and use the average value as the final estimator. The RBF kernel is set to $\exp(-5x^2)$, and the feature dimension is set to 100. 

\paragraph{Additional Figures}
We next report the relative MSE of the two estimators in the following figure.
\begin{figure}[H]
    \centering
    \includegraphics[scale=0.4]{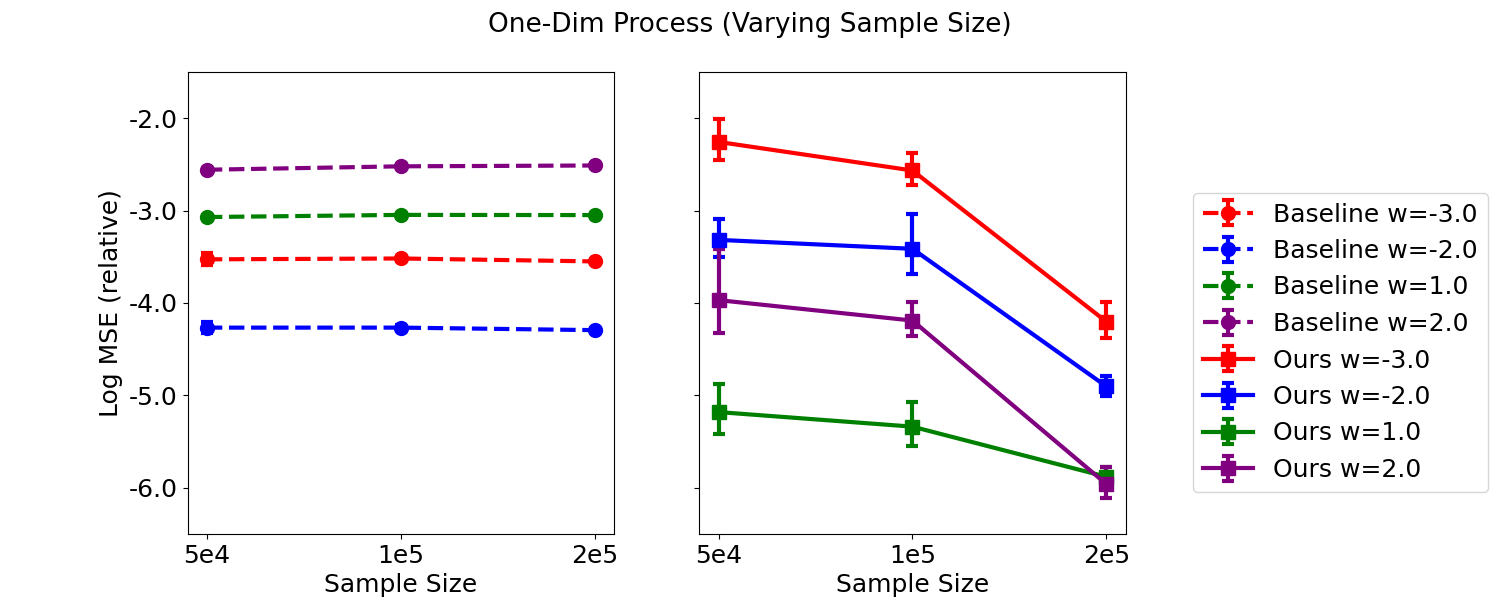}
    \caption{Experiments with varying sample size. Noise Level $\sigma_O=1.0$}
    \label{fig:Toy_Varying_SampleSizeMSE}
\end{figure}

\paragraph{CartPole Environment}
The state space is 4-dimensional, including position and velocity of cart, and angle and angle velocity of pole. The action space is $\{0,1\}$, corresponding to pushing the cart to the left or to the right. In addition, we use a modified reward function to better distinguish values among different policies. Specifically, the reward is defined as
\begin{align*}
    r = |2.0 - \frac{x}{x_{\text{clip}}}|\cdot |2.0 - \frac{\theta}{\theta_{\text{clip}}}| - 1.0
\end{align*}
where $x$ and $\theta$ are the position of Cart and angle of Pole, respectively, and $x_{\text{clip}}$ and $\theta_{\text{clip}}$ are the thresholds such that the episode will terminate (done = True) when either $|x| \geq x_{clip}$ or $|\theta| \geq \theta_{clip}$. Under this definition, the reward will be larger when the cart is closer to the center and the angle of the pole is closer to perpendicular. It is straightforward to show that $r$ is bounded between 0 and 3. Since we set $\gamma=0.5$, the value is bounded between 0 and 60.

\paragraph{CartPole Implementation}
We set the adversarial function spaces $\Vcal^\dagger$ and $\Wcal^\dagger$ to RKHSs to facilitate the computation. 
Both $b_v$ in PO-MQL and $b_w$ in PO-MWL are parameterized by a two-layer neural network with layer width = 256 and ReLU as activation function, and are optimized by a kernel loss function (see the derivation below). 

In the following, we use $K(x_1;x_2)$ to denote the RBF kernel:
\begin{align*}
    K(x_1;x_2) := \exp(-\frac{\|x_1-x_2\|_2}{2\beta^2})
\end{align*}
where $\beta$ denotes the bandwidth parameter. We choose $\beta = \text{med}/2$ during the training of PO-MWL ($b_w$) or MWL, and $\beta = \text{med}/5$ for PO-MQL ($b_v$) or MQL, where ``med'' is the median of the l2-distance over the samples in the dataset.
\paragraph{Derivation of the Loss Function for PO-MQL ($b_v$)}
Similar to the experiments in Toy environments, we reparameterize $g$ function with $\bar{g}\cdot \pi^e$, and therefore the predictor with $g$ should be adjusted by replacing $g$ with $\bar{g}\cdot \pi^e$:
\begin{align*}
    \EE_{O\sim \nu_O,A\sim \pi^e}[\bar{g}(O,A)\pi^e(A|O)].
\end{align*}
The loss function is given by:
\begin{align*}
    \max_f L_V^2(g,f)  =& \max_f \Big(\E[(R\pi^e(A|O)+\gamma \E_{a'\sim\pi^e}[g(a',O^+)]\pi^e(A|O)-g(A, O)\pi^e(A|O))f(A,O^-)]\Big)^2\\
    =&\max_f \Big(\Big\langle f, \E[R\pi^e(A|O)+\gamma \E_{a'\sim\pi^e}[g(a',O^+)]\pi^e(A|O)-g(A, O)\pi^e(A|O)]K(A,O^-;\cdot)\Big\rangle_{\mathcal{H}_K}\Big)^2\\
    =&\E[\Big(R\pi^e(A|O)+\gamma \E_{a'\sim\pi^e}[g(a',O^+)]\pi^e(A|O)-\pi^e(A|O)g(A, O)\Big)K(A,O^-;\barA,\barO^-)\\
    &\cdot\Big(\bar{R}\pi^e(\barA|\barO)+\gamma \E_{\bar{a}'\sim\pi^e}[g(\bar{a}',\barO^+)]\pi^e(\barA|\barO)-\pi^e(\barA|\barO)g(\barA,\barO)\Big)].\label{eq:kernel_loss_of_bv}
\end{align*}
\paragraph{Derivation of the Loss Function for PO-MWL ($b_w$)}
Similarly, we reparameterize $f$ function with $\bar{f}\cdot \pi^e$. Since we do not change the parameterization of $b_w$, we only need to adjust the loss function without changing the estimator:
\begin{align*}
    &\max_f L^2_W(g,f) \\
    =&\max_f \Big(\E[\gamma g(O^-,A)\pi^e(A|O)\sum_{a'}f(O^+,a')-g(O^-,A)f(O,A)]+(1-\gamma)\E_{O\sim \nu_O}[\sum_{a'}f(O,a')]\Big)^2\\
    =&\max_f \Big(\E[\gamma g(O^-,A)\pi^e(A|O)\E_{a'\sim \pi^e(\cdot|O^+)}[\bar{f}(O^+,a')]-g(O^-,A)\pi^e(A|O)\bar{f}(O,A)]+(1-\gamma)\E_{O\sim \nu_O,a'\sim\pi^e}[\bar{f}(O,a')]\Big)^2\\
    =&\max_f \Big(\E[\gamma g(O^-,A)\pi^e(A|O)\E_{a'\sim \pi^e(\cdot|O^+)}[\la \bar{f}, K(O^+,a'; \cdot)\ra]-g(O^-,A)\pi^e(A|O)\la \bar{f}, K(O,A;\cdot)\ra]\\
    &+(1-\gamma)\E_{O\sim \nu_O,a'\sim\pi^e}[\la \bar{f}, K(O,a';\cdot)\ra]\Big)^2\\
    =&\max_f \Big(\la \bar{f}, \E[\gamma g(O^-,A)\pi^e(A|O)\Big(\E_{a'\sim \pi^e(\cdot|O^+)}[K(O^+,a'; \cdot)]- K(O,A;\cdot)\Big)]+(1-\gamma)\E_{O\sim \nu_O,a'\sim\pi^e}[K(O,a';\cdot)]\ra\Big)^2\\
    =& \E[g(O^-,A)\pi^e(A|O)g(\barO^-,\barA)\pi^e(\barA|\barO)\cdot \\
    &\quad\quad\quad\Big(\gamma^2 K(O^+,\pi^e;\barO^+,\pi^e)+ K(O,A;\barO,\barA)-\gamma K(O^+,\pi^e;\barO,\barA)-\gamma K(O,A;\barO^+,\pi^e)\Big)]\\
    &+(1-\gamma)^2\E[K(O_0,\pi^e;\barO_0,\pi^e)]\tag{no gradient}\\
    &+\gamma(1-\gamma) \E[g(O^-,A)\pi^e(A|O) K(O^+,\pi^e;\barO_0,\pi^e)+g(\barO^-,\barA)\pi^e(\barA|\barO) K(O_0,\pi^e;\barO^+,\pi^e)]\\
    &- (1-\gamma) \E[g(O^-,A)\pi^e(A|O)K(O,A;\barO_0,\pi^e)+g(\barO^-,\barA)\pi^e(\barA|\barO)K(O_0,\pi^e;\barO,\barA)].
\end{align*}
where we denote 
$$K(O,\pi^e;\barO, \pi^e):=\E_{a'\sim \pi^e(\cdot|O), \bar{a}'\sim \pi^e(\cdot|\barO)}[K(O,a';\barO, \barA),\quad K(O,\pi^e;\barO, \barA):=\E_{a'\sim \pi^e(\cdot|O)}[K(O,a';\barO, \barA)].$$

\paragraph{Loss Function for Baseline Estimators}
We follow \citet{uehara2020minimax} to define the MQL and MWL loss function with the adversarial function space given by RHKSs. The loss function for MQL is given by:
\begin{align*}
    \max_f L_q^2(f,q)=& \E[\Big(R+\gamma \E_{a'\sim\pi^e(\cdot|O^+)}[ q(a',O^+)]-q(A, O)\Big)K(A,O;\barA,\barO)\Big(\bar{R}+\gamma \E_{\bar{a}'\sim\pi^e(\cdot|\barO^+)}[ q(\bar{a}',\barO^+)]-q(\barA,\barO)\Big)].
\end{align*}
The loss function for MWL is given by:
\begin{align*}
    &\max_f L_w^2(f,w)\\
    =&\E[w(O,A)w(\barO,\barA)\Big(K(O^+,\pi^e;\barO^+,\pi^e)+ K(O,A;\barO,\barA)-\gamma K(O^+,\pi^e;\barO,\barA)-\gamma K(O,A;\barO^+,\pi^e)\Big)]\\
    &+(1-\gamma)^2\E[K(O_0,\pi^e;\barO_0,\pi^e)]\tag{no gradient}\\
    &+\gamma(1-\gamma)\E[w(O,A)K(O^+,\pi^e;\barO_0,\pi^e)+w(\barO,\barA)K(O_0,\pi^e;\barO^+,\pi^e)]\\
    &-(1-\gamma)\E[w(O,A)K(O,A;\barO_0,\pi^e)+w(\barO,\barA)K(O_0,\pi^e;\barO,\barA)].
\end{align*}
In addition, we use the same neural network architecture and the same choice of bandwidth during the training of MQL/MWL as those for PO-MQL/PO-MWL.


\input{proof}

\end{document}

%% file: notation.tex
\newcommand{\IS}{\mathrm{IS}}
\newcommand{\DR}{\mathrm{DR}}
\newcommand{\VM}{\mathrm{VM}}

\newcommand{\cj}{\mathcal{J}}

\newcommand{\rI}{\mathrm{I}}

\newcommand{\barO}{\bar{O}}
\newcommand{\barA}{\bar{A}}

\newcommand{\la}{\langle}
\newcommand{\ra}{\rangle}

\DeclareMathAlphabet\mathbfcal{OMS}{cmsy}{b}{n}

%% file: math_commands.tex
\usepackage{amsmath,amsfonts,bm}

\def\eqref#1{equation~\ref{#1}}

\def\1{\bm{1}}

\def\ra{{\textnormal{a}}}

\def\rd{{\textnormal{d}}}

\DeclareMathAlphabet{\mathsfit}{\encodingdefault}{\sfdefault}{m}{sl}
\SetMathAlphabet{\mathsfit}{bold}{\encodingdefault}{\sfdefault}{bx}{n}

\newcommand{\E}{\mathbb{E}}

\DeclareMathOperator*{\argmin}{arg\,min}

%% file: time_inhomogeneous.tex
\section{OPE in Time-inhomogeneous POMDPs} \label{sec:time_inho}

We consider the time-inhomogeneous setting in this section where the system dynamics, evaluation and behavior policies are allowed to vary over time. We first introduce the identification method for the target policy's value by introducing value and weight bridge functions. We next present the proposed value function-based, IS and DR estimators. We remark that although we focus on evaluation of Markovian policies in this section, the proposed value function-based estimator can be extended to settings where evaluation policies are history-dependent in \pref{sec:history}. 

\begin{figure}[!t]
\centering
\begin{subfigure}{.5\textwidth}
  \centering
  \includegraphics[width=\linewidth]{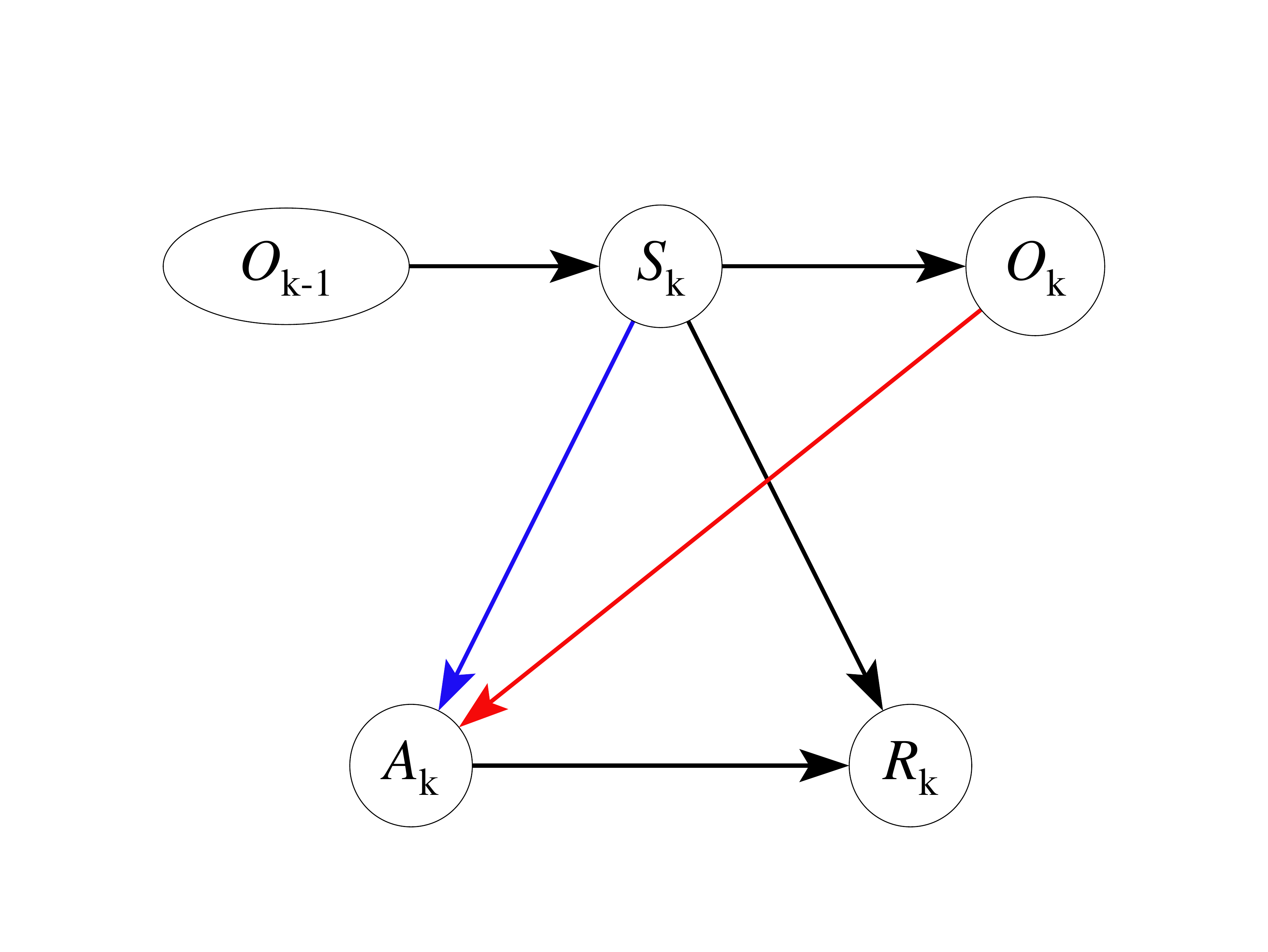}
  \caption{The case where we use the prior observation as a negative control. }
  \label{fig:history_1}
\end{subfigure}%
\begin{subfigure}{.5\textwidth}
  \centering
  \includegraphics[width=\linewidth]{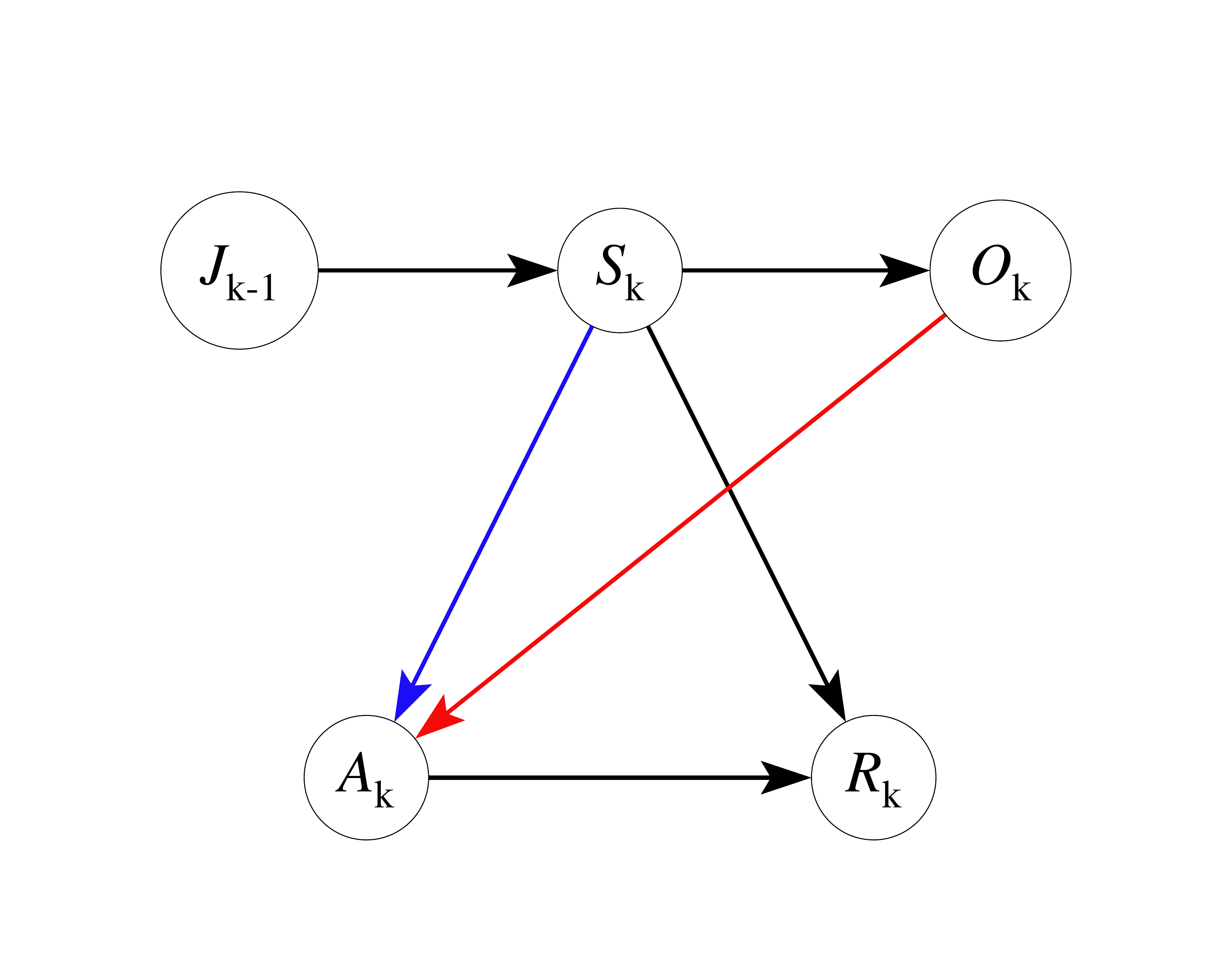}
  \caption{The case where we use the prior history as a negative control.}
  \label{fig:history_2}
\end{subfigure}
\end{figure}

\subsection{Identification}

We first introduce the bridge function and the learnable bridge function. For any $j\ge 1$, Let $ \mu_j(s_j)=\frac{P_{\epol,j}(s_j)}{P_{\bpol,j}(s_j)}$ where $P_{\epol,j}$ and  $P_{\bpol,j}$ are the marginal density functions of $S_j$ under $\epol$ and $\bpol$, respectively. Let
$\mathfrak{J}_{j-1}=\{O_0,A_0,O_1,A_1,R_1,O_2,\cdots, R_{j-1}\}$. 

\begin{assum}[Existence of bridge functions]\label{assum:bridge_in}
There exist value bridge functions $\{b'^{[j]}_V\}_{j=0}^{H-1}$ and weight bridge functions  $\{b'^{[j]}_W\}_{j=0}^{H-1}$, defined as solutions to 
\begin{align}\label{eq:solution1}
    \E[b'^{[j]}_{V}(A_j,O_{j})\mid A_j,S_j] &=\E_{\epol}[\sum_{t=j}^{H-1} \gamma^{t-j}R_t\epol_j(A_j\mid  {O_j})|A_j,S_j] ,\\ 
\E[b'^{j}_{W}(A_j,\mathfrak{J}_{j-1}) \mid A_j,S_j] &=  \mu_{j}(S_{j})\frac{1}{P_{\bpol}(A_j \mid S_j)}\label{eq:solution2}.
\end{align}

\end{assum}

\begin{definition}[Learnable bridge functions]\label{assum:observed_bridge_in}
Learnable value bridge functions $\{b^{[j]}_{V}\}_{j=0}^{H-1}$ and learnable weight bridge functions $\{b^{[j]}_{W}\}_{j=0}^{H-1}$ are defined as solutions to 
\begin{align}\label{eqn:bVbW}
\begin{split}
   & \E[b^{[j]}_{V}(A_j,O_{{j}})\mid A_j,\mathfrak{J}_{j-1}] =\E[R_j \epol_j(A_j \mid {O_{j}}) +\gamma\sum_{a'}b^{[j+1]}_V(a', {O_{j+1}})\epol_j(A_j \mid {O_{j}}) \mid  A_j,\mathfrak{J}_{j-1}],\\ 
   & \E\bracks{ \sum_{a'} b^{[j]}_{W}(A_j,\mathfrak{J}_{j-1})\epol_j(A_j |O_{j})f({O_{j+1}},a')- b^{[j+1]}_{W}(A_{j+1},{\mathfrak{J}_j}) f({O_{j+1}},A_{j+1})}=0,\,\forall f:\Acal\times \Ocal \to \RR. 
 \end{split}  
 \end{align}
\end{definition}

It is important to note that we use the whole history $\mathfrak{J}_{j-1}$ in the integral equations to define the learnable bridge functions. Alternatively, one can replace $\mathfrak{J}_{j-1}$ with the most recent observation $O_{j-1}$. The advantage of using the whole history over a single observation is that it requires weaker assumptions to ensure the existence of the bridge functions. See the discussion below Assumption \ref{assum:discrete} for details. 


Next, we present our key identification theorem under Assumption \pref{assum:bridge_in}. 

\begin{theorem}\label{thm:identimein}
Suppose Assumption \pref{assum:bridge_in} holds. 
Then, bridge functions are learnable bridge functions. In addition, any learnable bridge function satisfies
\begin{align*}
    J(\epol)=\E[\sum_a b^{[0]}_V(a,O_0)],\quad   J(\epol)=\E[\sum_{t=0}^{H-1} b^{[t]}_W(A_t,\mathfrak{J}_{t-1})\epol_t(A_t\mid O_t)\gamma^t R_t]. 
\end{align*}
\end{theorem}

The next theorem states we can similarly identify the policy value by assuming the completeness. 

\begin{theorem}\label{thm:identimein2}
Suppose the existence of leanrbale value bridge functions and the completeness:
\begin{align*}
    \EE[g(S_j,A_j) \mid A_j,\mathfrak{J}_{j-1} ] =0 \implies g()=0. 
\end{align*}
Then, any learnable bridge function satisfies
\begin{align*}
    J(\epol)=\E[\sum_a b^{[0]}_V(a,O_0)]. 
\end{align*}
\end{theorem}


To further elaborate Assumptions in Theorem \ref{thm:identimein} and Theorem \ref{thm:identimein2}, we focus on the tabular setting in the rest of this section when $\Scal,\Ocal$, and the reward space are discrete. 

First, we see sufficient conditions to ensure Assumptions in \pref{thm:identimein}. Assumptions in \pref{thm:identimein}, i.e., Assumption \pref{assum:bridge_in}, are immediately implied by the following conditions. 

\begin{assum}\label{assum:discrete}
For any $0\leq j\leq H-1$, 
\begin{align*}
    \rank(\Pr(\bO_j \mid \bS_j))= |\Scal|,\quad  \rank(\Pr(\mathbfcal{J}_{j-1} \mid \bS_j))= |\Scal|. 
\end{align*}
\end{assum}

Next, we see sufficient conditions to ensure Assumption in \pref{thm:identimein2}. Assumptions in \pref{thm:identimein2} are immediately implied by the following conditions. 

\begin{assum}\label{assum:discrete2}
For any $0\leq j\leq H-1$, 
\begin{align*}
    \rank(\Pr(\bO_j \mid \bS_j))= |\Scal|,\quad  \rank(\Pr(\bS_j \mid \mathbfcal{J}_{j-1},A_j=a))= |\Scal|. 
\end{align*}
\end{assum}

These assumptions imply that the state space is sufficiently informative and $|\Ocal|\geq |\Scal|$. Clearly, this shows the advantage of using the whole $\mathbfcal{J}_{j-1}$ as a negative control rather than $O_{j-1}$ itself since $\rank(\Pr(\mathbfcal{J}_{j-1} \mid \bS_j))= |\Scal|$ is weaker than $\rank(\Pr(\bO_{{j-1}} \mid \bS_j))= |\Scal|$, and similarly, $\rank(\Pr(\bS_j \mid \mathbfcal{J}_{j-1},A_j=a))= |\Scal|$ is weaker than $\rank(\Pr(\bS_j \mid \bO_{{j-1}}))= |\Scal|$. 

In addition, Assumption \pref{assum:discrete2} is equivalent to Assumption 2 in \citet{nair2021spectral} ($\rank(\Pr(\bO_j \mid \mathbfcal{J}_{j-1},A_j=a))= |\Scal|$ ) as we see in the bandit setting. 
Under this assumption, 
the policy value is identifiable, as we show in the following lemma. 
\begin{lemma}\label{lem:nair}
Suppose Assumption \pref{assum:discrete2} holds. 
Then the policy value equals
{\small 
\begin{align*}
& \sum_{s_0,a_0,\cdots,r_{H-1}} \sum_{k=0}^{H-1} \gamma^k r_k \tau_k \Prr_{\bpol}(r_k, o_k \mid a_k,\mathbfcal{J}_{t-1})\Prr_{\bpol}(\bO_{k} \mid a_k, \mathbfcal{J}_{t-1})^{+}\Prr_{\bpol}(\bO_{k},o_{k-1}\mid a_{k-1},\mathbfcal{J}_{k-2})\Prr_{\bpol}(\bO_{{k-1}} \mid a_{k-1}, \mathbfcal{J}_{k-2})^{+} \\ 
&
\times \cdots \Prr_{\bpol}(\bO_{0}), 
\end{align*}
where $\tau_k=\prod_{t=1}^k \epol_t(a_t\mid o_t)$. 
}
\end{lemma}

\subsection{Estimation}

We present the estimation method in this section. 
Suppose we have certain consistent estimators $\hat b_V=\{\hat b^{[j]}_V \}$ and $\hat b_W=\{\hat b^{[j]}_W \}$ for $b_V=\{ b^{[j]}_V \}$ and $b_W=\{ b^{[j]}_W \}$, respectively. \pref{thm:identimein} suggests that we can estimate the policy value based on the following value function-based and IS estimators.
\begin{align*}
    \hat J_{\VM}=\E[\sum_a \hat b^{[0]}_V(a,O_0)],\quad \hat J_{\IS}= \E_{\Dcal}[\sum_{t=0}^{H-1} \hat b^{[t]}_W(A_t,\mathfrak{J}_{t-1})\epol_t(A_t\mid O_t)\gamma^t R_t]. 
\end{align*}
In addition, we can similarly combine these two estimators to construct the DR estimator $\hat J_{\DR} =\E_{\Dcal}[J(\hat b_V,\hat b_W)]$ where
\begin{align*}
    J(f,g) &=\sum_a f^{[0]}(a,O_0)+\sum_{k=0}^{H-1}\gamma^k g^{[k]}(A_k,\mathfrak{J}_{k-1})\prns{\epol_k(A_k \mid O_k)\{R_k+\gamma \sum_{a^+}f^{[k+1]}(a^+,O_{k+1}) \}- f^{[k]}(A_k,O_k) },
\end{align*}
where $f=\{f^{[t]}\}$ and $g=\{g^{[t]}\}$ such that $f^{[t]}:\Acal \times  \Ocal \to \RR ,g^{[t]}:\Acal \times \cj_{t-1} \to \RR$. Here, $\cj_t$ denotes the domain over $\mathfrak{J}_t$. 

In the fully observable MDP setting, $\hat J_{\IS}$ reduces to the one in \citet{XieTengyang2019OOEf} and $\hat J_{\DR}$ reduces to the one in \citet{kallus2020double}. The following theorem proves the doubly-robustness property of $ \hat J_{\DR}$. 

\begin{theorem}[Doubly robust property]\label{thm:doubly_finite}
Under Assumption \ref{assum:bridge_in}, for any $f=\{f^{[t]}\}$ and $g=\{g^{[t]}\}$ such that $g^{[t]}:\Acal \times \cj_{t-1} \to \RR ,f^{[t]}:\Acal \times \Ocal \to \RR$, we have 
\begin{align*}
    J(\epol)=\E[J(b_V,g)]= \E[J(f,b_W)]. 
\end{align*}

\end{theorem}

Finally, we discuss how to estimate the learnable bridge functions. 
Similar to \eqref{eqn:hatbV} and \eqref{eqn:hatbW}, we can employ minimax learning to estimate $b_V$ and $b_W$ based on the integral equations in \eqref{eqn:bVbW}. However, different from \eqref{eqn:hatbV} and \eqref{eqn:hatbW}, $b_V$ and $b_W$ need to be estimated in a recursive fashion in the time-inhomogeneous setting. Specifically, to estimate $b_V$, we begin by defining $\hat b^{[H]}_V=0$. We next sequentially estimate $\hat b^{[H-1]}_V, \hat b^{[H-2]}_V,\cdots,\hat b^{[0]}_V$ in a backward manner. Similarly, to learn $b_W$, we define $\hat b^{[-1]}_W=1$ and recursively estimate $\hat b^{[0]}_W, \hat b^{[1]}_W,\cdots,\hat b^{[H-1]}_W $ in a forward manner. 
Theoretical properties of these minimax estimators can be similarly established, as in Section \ref{subsec:theory}, and we omit the technical details. 

\section{Evaluation of History-Dependent Policies } \label{sec:history}

We have so far assumed that the evaluation policies are Markovian. 
In this section, we consider the case when evaluation policies are history-dependent, i.e., $\pi^e:\Ocal \times \tilde \Jcal_{t-1} \to \Delta(\Acal)$ where $\tilde \Jcal_{t-1} =\prod_{k=0}^{t-1} (\Ocal \times \Acal) $.  Note $\tilde \Jcal_{t-1}$ is different from $\Jcal_{t-1}$ recalling $\Jcal_{t-1}=\Ocal \times \{\prod_{k=0}^{t-1} (\Ocal \times \Acal)\}$. We mainly analyze the value-function based estimator in this section.


We introduce the value and learnable bridge functions as follows. They are natural extensions of Assumption \ref{assum:bridge_in} and Definition \ref{assum:observed_bridge_in} for history-dependent policies. Let $\mathfrak{\tilde J}_{j-1}=\{O_0,A_0,\cdots,O_{H-1},A_{H-1} \}$. Note this is contained in $\mathfrak{J}_{j-1}=\{O^{-}_0,O_0,A_0,\cdots,O_{H-1},A_{H-1} \}$. 

\begin{assum}[Existence of bridge functions]\label{assum:bridge_in_non}
There exist value bridge functions $\{b'^{[j]}_V\}_{j=0}^{H-1}$ defined as solutions to 
\begin{align*}
    \E[b'^{[j]}_{V}(A_j,O_{j}, \mathfrak{\tilde J}_{j-1})\mid A_j,S_j, \mathfrak{\tilde J}_{j-1}] &=\E_{\epol}[\sum_{t=j}^{H-1} \gamma^{t-j}R_t\epol_j(A_j\mid  {O_j},\mathfrak{\tilde J}_{j-1})|A_j,S_j, \mathfrak{\tilde J}_{j-1}]. 
\end{align*}
There exist weight bridge functions $\{b'^{[j]}_W\}_{j=0}^{H-1}$ defined as solutions to 
\begin{align*}
    \E[b'^{[j]}_{W}(A_j,\mathfrak{J}_{j-1})\mid A_j,S_j, \mathfrak{\tilde J}_{j-1}] &= \mu_j(S_j, \mathfrak{\tilde J}_{j-1})/\bpol_j(A_j \mid S_j) 
\end{align*}
where $\mu_j(S_j, \mathfrak{\tilde J}_{j-1})$ is $P_{\pi^e,j}(S_j,\mathfrak{\tilde J}_{j-1})/ P_{\pi^b,j}(S_j,\mathfrak{\tilde J}_{j-1}) $. 
\end{assum}

\begin{definition}[Learnable bridge functions]\label{assum:observed_bridge_in_non}
Learnable value bridge functions $\{b^{[j]}_{V}\}_{j=0}^{H-1}$ are defined as solutions to 
\begin{align*}
\begin{split}
   & \E[b^{[j]}_{V}(A_j, O_{{j}},\mathfrak{\tilde J}_{j-1} )\mid A_j, \mathfrak{J}_{j-1}] =\E[R_j \epol_j(A_j \mid {O_{j}} ,\mathfrak{\tilde J}_{j-1}) +\gamma\sum_{a'}b^{[j+1]}_V(a', {O_{j+1}}, \mathfrak{\tilde J}_{j})\epol_j(A_j \mid {O_{j}},\mathfrak{\tilde J}_{j-1}) \mid  A_j,\mathfrak{J}_{j-1}]. 
 \end{split}  
 \end{align*}
\end{definition}

It is worthwhile to note $\mathfrak{J}_j$ is contained in $\mathfrak{\tilde J}_j$. If evaluation policies depend on the whole $\mathfrak{J}_j$, it is unclear how to identify the policy value in our framework. 

Now, we are ready to prove the identification formula. 

\begin{theorem}\label{thm:identimein_non_reactiv}
~
\begin{itemize}
    \item Suppose Assumption \pref{assum:bridge_in_non}. Then, any learnable bridge function satisfies
\begin{align*}
    J(\epol)=\E[\sum_a b^{[0]}_V(a,O_0)]. 
\end{align*}
    \item Suppose the existence of value bridge functions and the completness assumption: 
\begin{align*}
    \EE[ g(S_j,A_j,\mathfrak{\tilde J}_{j-1}) \mid  A_j,\mathfrak{J}_{j-1}]= 0  \implies g(\cdot)=0 
\end{align*}
for any $0\leq j\leq H-1$. Then, any learnable bridge function satisfies
\begin{align*}
    J(\epol)=\E[\sum_a b^{[0]}_V(a,O_0)]. 
\end{align*}
\end{itemize}

\end{theorem}

Similar to \pref{sec:time_inho}, $b_V$ can be estimated in a recursive fashion. In the tabular setting, the sufficient conditions to ensure the existence of value bridge functions and the completness is
\begin{align*}
    \rank(\Prr( \Ob_j \mid A_j=a,\mathbfcal{J}_{j-1}))= |\Scal|. 
\end{align*}
The nonasymptotic properties of the resulting estimator can be similarly established as in Section \pref{subsec:theory}

%% file: proof.tex
\section{Omitted proof}

\subsection{Proof of \pref{sec:bandit} }

\begin{proof}[Proof of \pref{lem:first_step}]
First, we prove the value-function based identification formula: 
\begin{align*}
    &\E[\sum_{a}b'_{V}(a,O_0)]=\E[\sum_{a}\E[b'_{V}(a,O_0)|S_0]]=\E[\sum_{a}\E[b'_{V}(a,O_0)|A_0=a,S_0]]  \tag{$A_0\perp O_0 \mid  S_0 $}\\ 
    &=\E[\sum_{a}\E[R_0\pi^e(a|O_0)|A_0=a,S_0]]  \tag{Definition of bridge functions} \\ 
    &=\E[\sum_{a}\E[R_0|A_0=a,S_0]\E[\pi^e(a\mid O_0)\mid A_0=a,S_0]] \tag{$R_0 \perp O_0 \mid A_0,S_0$ }\\
    &=\E[\sum_{a}\E[R_0|A_0=a,S_0]\E[\pi^e(a\mid O_0)\mid S_0]]  \tag{$A_0 \perp O_0 \mid S_0$}\\
    &=\E[\sum_{a}\E[R_0|A_0=a,S_0]\pi^e(a\mid O_0) ]\\
    &=J(\epol).  \tag{From standard regression formula}
\end{align*}

Second, we prove the importance sampling identification formula: 
\begin{align*}
&\E[b_{W}(A_0,O^{-}_0)R_0\pi^e(A_0\mid O_0)] = \E[\E[b'_{W}(A_0,O^{-}_0) \mid S_0,A_0]R_0\pi^e(A_0\mid O_0)] \tag{$O^{-}_0 \perp R_0\mid S_0,A_0   $} \\
&= \E[1/\pi^b(A_0\mid S_0)\pi^e(A_0\mid O_0)R_0]\tag{Definition of $b'_W$ } \\
&=J(\epol).  \tag{From standard IPW formula}
\end{align*}
\end{proof}

\begin{proof}[Proof of \pref{lem:observed}]

We first prove reward bridge functions are learnable reward bridge functions: 
  \begin{align*}
     &\E[R_0\epol(A_0|O_0)\mid A_0,O^{-}_0]=\E[\E[R_0\epol(A_0|O_0)\mid A_0,S_0,O^{-}_0]\mid A_0, O^{-}_0]\\
     &=\E[\E[R_0\epol(A_0|O_0)\mid A_0,S_0]\mid A_0, O^{-}_0] \tag{$R_0, O_0 \perp O^{-}_0 \mid S_0,A_0 $ }\\ 
     & = \E[\E[b'_{V}(A_0,O_0)\mid A_0,S_0]\mid O^{-}_0,A_0 ] \tag{Definition of bridge functions}\\\
     &=\E[\E[b'_{V}(A_0,O_0)\mid S_0,O^{-}_0 ,A_0]\mid O^{-}_0,A_0]   \tag{$O_0 \perp O^{-}_0 \mid S_0,A_0 $ }\\\
     &=\E[b'_{V}(A_0,O_0)\mid O^{-}_0,A_0 ]. 
\end{align*}

Next, we prove weight bridge functions are learnable weight bridge functions: 
\begin{align*}
    &\E[b'_W(A_0,O^{-}_0) \mid A_0,O_0 ]= \E[\E[b'_W(A_0,O^{-}_0)\mid A_0,O_0,S_0 ] \mid A_0,O_0 ]\\
    &=  \E[\E[b'_W(A_0,O^{-}_0)\mid A_0,S_0 ] \mid A_0,O_0 ] \tag{$O_0 \perp O^{-}_0  \mid A_0,S_0 $}\\
    &=\E[1/\pi^b(A_0\mid S_0)\mid A_0,O_0 ]  \tag{Definition of bridge functions}\\
 &=\E[1/\pi^b(A_0\mid S_0,O_0)\mid A_0,O_0 ]  \tag{$A_0 \perp O_0 \mid S_0$}\\
    &=1/P_{\pi_b}(A_0\mid O_0). 
\end{align*}
\end{proof}

\begin{proof}[Proof of \pref{lem:some_characterization}]

First, note
 $$\E[\sum_a f(O_0,a)-b_W(A_0,O^-_0)f(O_0,A_0)]=0,\quad \forall f:\Ocal\times \Acal \to \RR,$$
is equivalent to 
$$\E[f(O_0,a)-b_W(A_0,O^-_0)\mathrm{I}(A_0=a)f(O_0,a)]=0,\quad \forall f:\Ocal\times \Acal \to \RR,\forall a\in \Acal.$$
This condition is equivalent to 
\begin{align*}
    \E[f(O_0,a)\{1-b_W(A_0,O^-_0)\pi^b(a_0\mid O_0)\}]=0,\quad \forall f:\Ocal\times \Acal \to \RR.
\end{align*}
This is equivalent to $b_W(A_0,O^-_0)\pi^b(a_0\mid O_0)=1$.

\end{proof}

\begin{proof}[Proof of \pref{thm:final_step}]
We prove the importance sampling formula. Let $b'_V$ be a reward bridge function and $b_W$ be an learnable bridge function. 
Then,
\begin{align*}
        J(\epol)&= \E[\sum_a b'_V(a,O_0)]=\E[1/P_{\bpol}(A_0\mid O_0) b'_V(A_0,O_0) ] \tag{Result of \pref{lem:first_step}}\\
        &= \E[\E[b_W(A_0,O^{-}_0) \mid A_0,O_0] b'_V(A_0,O_0) ] \tag{Definition} \\ 
     &= \E[b_W(A_0,O^{-}_0) b'_V(A_0,O_0) ]\\ 
        &= \E[b_W(A_0,O^{-}_0)  \E[b'_V(A_0,O_0)\mid A_0,S_0] ] \\
        &=\E[b_W(A_0,O^{-}_0)\E[R_0\epol(A_0\mid O_0)\mid A_0,S_0] ] \tag{Definition}\\
        &= \E[b_W(A_0,O^{-}_0)R_0\epol(A_0\mid O_0)]. 
\end{align*}

We prove the value function based formula. Let $b'_W$ be a weight bridge function and $b_V$ be an learnable weight bridge function. Then, 
\begin{align*}
    J(\epol)&=\EE[b'_W(A_0,O^{-}_0)R_0\pi^e(A_0\mid O_0)]=\EE[b'_W(A_0,O^{-}_0)\E[R_0\pi^e(A_0\mid O_0)\mid A_0,O^{-}_0]] \tag{Result of \pref{lem:first_step}}\\ 
      &= \EE[b'_W(A_0,O^{-}_0)b_V(A_0,O_0)] \tag{Definition}\\
       &= \EE[\E[b'_W(A_0,O^{-}_0)\mid A_0,S_0 ]b_V(A_0,O_0)] \\
        &= \EE[1/\pi^b(A_0\mid S_0) b_V(A_0,O_0)] \tag{Definition}\\
        &=\E[\sum_a b_V(a,O_0) ]. 
\end{align*}
\end{proof}

\subsubsection{Discrete setting}

\cite{tennenholtz2020off}  assume the following. 
\begin{assum}\label{asm:strong}
$\Prr(\bO_0 \mid A_0=a,\bO^{-}_0),\Prr(\mathbf{S}_0\mid A_0=a,\bO^{-}_0)$ are invertible. 
\end{assum}
This implies $|\bS|=|\bO|$, which is very strong. Remark this is also assumed in the paper which proposed negative controls \cite{miao2018identifying}. Instead, we require the following weaker assumption: 
\begin{assum}\label{asm:weak2}
$\rank(\Prr(\bO_0\mid \bS_0))=|\Scal|$ and $\rank(\Prr(\bO^{-}_0\mid \bS_0))=|\Scal|$. 
\end{assum}
We show this is equivalent to the assumption in  \cite{nair2021spectral}:
\begin{assum}
\label{asm:weak3}
$\rank(\Prr(\bO_0 \mid A_0=a,\bO^{-}_0))= |\Scal|$. 
\end{assum}

\begin{proof}[Proof of \pref{lem:equivalence}]
Assumption \pref{asm:weak2} implies Assumption \pref{asm:weak3} since 
\begin{align*}
    |\Scal|= \rank(\Prr(\mathbf{O}_0 \mid A_0=a,\bO^{-}_0))\leq    \min( \rank( \Prr(\mathbf{O}_0 \mid A_0=a,\bS_0)), \rank(\Prr(\bS_0 \mid A_0=a,\bO^{-}_0))). 
\end{align*}
Besides, we have 
\begin{align*}
    \{\Prr(\bS_0 \mid A_0=a,\bO^{-}_0)\}^{\top}=\Lambda_1 \Prr(\bO^{-}_0 \mid A_0=a,\bS_0)\Lambda_2 
\end{align*}
where $\Lambda_1$ is a $|\bO^{-}| \times |\bO^{-}|$ matrix s.t. $(i,i)$-th element is $1/P_{\bpol}(O^-= o^{-} \mid S_0=s) $ and $\Lambda_2 $ is a $|\Scal|\times |\Scal|$ matrix s.t. $(i,i)$-th element is $P_{\bpol}(S-0=s\mid A_0=a)$. 
Thus, $\rank(\Prr(\bO^{-}_0 \mid \bS_0))=\rank(\Prr(\bO^{-}_0 \mid A_0=a,\bS_0))=\rank(\Prr(\bS_0 \mid A_0=a, \bO^{-}_0)) \geq |\Scal|$. Then, Assumption \pref{asm:weak3} is concluded. 

Assumption \pref{asm:weak2} implies assumption \pref{asm:weak3} from Sylvester’s rank inequality:
\begin{align*}
 |\Scal| &=\rank( \Prr(\mathbf{O}_0 \mid A_0=a,\bS_0)) \tag{From Assumption \pref{asm:weak2}} \\
  &= \rank( \Prr(\mathbf{O}_0 \mid A_0=a,\bS_0))+ \rank(\Prr(\bO^{-}_0 \mid A_0=a, \bS_0 ))-|\Scal| \tag{From Assumption \pref{asm:weak2}}\\ 
 &=\rank( \Prr(\mathbf{O}_0 \mid A_0=a,\bS_0))+ \rank(\Prr(\bS_0 \mid A_0=a,\bO^{-}_0))-|\Scal| \tag{See argument above}\\
&\leq \rank( \Prr(\mathbf{O}_0 \mid A_0=a,\bS_0)\Prr(\bS_0 \mid A_0=a,\bO^{-}_0)) \tag{ Sylvester’s rank inequality:}\\
&= \rank(\Prr(\mathbf{O}_0 \mid A_0=a,\bO^{-}_0)).
\end{align*}
Besides,
\begin{align*}
\rank(\Prr(\mathbf{O}_0 \mid A_0=a,\bO^{-}_0))\leq    \min( \rank( \Prr(\mathbf{O}_0 \mid A_0=a,\bS_0)), \rank(\Prr(\bS_0 \mid A_0=a,\bO^{-}_0)))=|\Scal|. 
\end{align*}
Thus, 
\begin{align*}
    \rank(\Prr(\mathbf{O}_0 \mid A_0=a,\bO^{-}_0))=|\Scal|. 
\end{align*}
\end{proof}

Finally, we give the identification formula:  
\begin{proof}[Proof of \pref{thm:discrete_identify}]
We have
\begin{align*}
    \Pr(\bO_0 \mid \bO^{-}_0, a)   =      \Pr(\bO_0 \mid \bS_0, a) \Pr(\bS_0\mid \bO^{-}_0, a),\quad \Pr(r, o \mid \bO^{-}_0,a)  =     \Pr(r,o \mid \bS_0,a) \Pr(\bS_0\mid \bO^{-}_0, a). 
\end{align*}
Let the SVD decomoposition of $  \Pr(\bO_0 \mid \bO^{-}_0, a) $ be $ABC$, where $A$ is a $|\Ocal|\times |\Scal|$ matrix, $B$ is a $ |\Scal|\times |\Scal|$ matrix, and $C$ is a $|\Scal|\times |\Ocal|$ matrix. 
\begin{align*}
    \mathrm{I}= A^{\top}\Pr(\bO_0 \mid \bO^{-}_0, a)C^{\top}B^{-1}=A^{\top}\Pr(\bO_0 \mid \bS_0, a)\Pr(\bS_0 \mid \bO^{-}_0,a )C^{\top}B^{-1}. 
\end{align*}
Therefore, we have
\begin{align*}
   A^{\top} \Pr(\bO_0 \mid \bS_0 )=\{ \Pr(\bS_0 \mid \bO^{-}_0,a )C^{\top}B^{-1}\}^{-1}. 
\end{align*}
noting $ \Pr(\bS_0 \mid \bO^{-}_0,a) C^{\top}B^{-1} $ is full-rank matrix from the assumption.  

In addition, we have 
\begin{align}\label{eq:minor}
 \Pr(r,o \mid \bO^{-}_0,a) C^{\top}B^{-1} = \Pr(r,o  \mid \bS_0,a)  \Pr(\bS_0\mid \bO^{-}_0, a)C^{\top}B^{-1}. 
\end{align}

Thus, 
\begin{align*}
 \Pr(r,o  \mid \bS_0,a) &= \Pr(r,o  \mid \bO^{-}_0,a) C^{\top}B^{-1}\{\Pr(\bS_0\mid \bO^{-}_0, a)C^{\top}B^{-1}\}^{-1} \tag{From \pref{eq:minor}} \\ 
   &=  \Pr(r,o  \mid \bO^{-}_0,a) C^{\top}B^{-1}A^{\top}\Pr(\bO_0 \mid \bS_0, a)\\
   &= \Pr(r,o  \mid \bO^{-}_0,a)\{ \Pr(\bO_0 \mid \bO^{-}_0, a)\}^{+}\Pr(\bO_0 \mid \bS_0). 
\end{align*}
Besides, 
\begin{align*}
    J &=\sum_{a,r,o,s} r p(r,o \mid s,a)\epol(a\mid o)p(s) \tag{Definition} \\
       &=\sum_{a,r,o,s}   \Pr(r,o  \mid \bO^{-}_0,a)\{ \Pr(\bO_0 \mid \bO^{-}_0, a)\}^{+}\Pr(\bO_0 \mid s)\epol(A \mid O)p(s) \\
      &=\sum_{a,r,o}r\epol(a\mid o) \Pr(r ,o \mid \bO^{-}_0,a)\{ \Pr(\bO_0 \mid \bO^{-}_0, a)\}^{+}\Pr(\bO_0). 
\end{align*}

\end{proof}

\subsection{ Proof of  Section \ref{subsec:identify_homo}   }

We define $Q(s,a)\coloneqq \E[b'_V(a,O) \mid A=a,S=s]$. We show this $Q(s,a)$ satisfies a recursion formula. 

\begin{lemma}\label{lem:value2}
 \begin{align*}
         Q(S,A)&=\E[R \epol(A\mid O)\mid A,S]+  \gamma\E\bracks{\sum_{a'}Q(S^{+},a')\epol(A \mid O) \mid A,S}. 
 \end{align*}
\end{lemma}
\begin{proof}[Proof of Lemma \ref{lem:value2}]
From the definition, 
\begin{align*}
   Q(s,a)&=\E[R_0 \epol(a\mid O_0)\mid A_0=a,S_0=s]+  \E_{\epol}\bracks{\sum_{t=1}^{\infty} \gamma^t R_t\epol(a\mid O_0)\mid A_0=a,S_0=s}. 
 \end{align*}  
 Then, we have 
 \begin{align*}
    &\E_{\epol}\bracks{\sum_{t=1}^{\infty} \gamma^t R_t\epol(a\mid O_0)\mid A_0=a,S_0=s}\\
    &= \gamma\E\bracks{\sum_{a'}\E_{\epol}[\sum_{t=1}^{\infty} \gamma^{t-1} R_t\mid S_1,A_1=a']\epol(a'\mid O_1)\epol(a\mid O_0)\mid A_0=a,S_0=s}\\
  &= \gamma\E\bracks{\sum_{a'}\E_{\epol}[\sum_{t=1}^{\infty} \gamma^{t-1} R_t\mid S_1,A_1=a']\E[\epol(a'\mid O_1)\mid S_1,A_1=a']\epol(a\mid O_0)\mid A_0=a,S_0=s}\\
    &= \gamma\E\bracks{\sum_{a'}\E_{\epol}[\sum_{t=1}^{\infty} \gamma^{t-1} R_t\epol(a'\mid O_1)\mid S_1,A_1=a']\epol(a\mid O_0)\mid A_0=a,S_0=s} \tag{$R_1,\cdots, R_{H-1}\perp O_1\mid S_1,A_1  $}\\
    &= \gamma\E\bracks{\sum_{a'}Q^{\epol}(a',S_1)\epol(a\mid O_0) \mid A_0=a,S_0=s}. 
\end{align*}
In conclusion, 
 \begin{align*}
    Q(S,A)=\E[R \epol(a\mid O)\mid A,S]+  \gamma\E\bracks{\sum_{a'}Q(S^{+},a')\epol(A \mid O) \mid A,S}. 
 \end{align*}
\end{proof}

By using the above lemma, we show learnable reward bridge functions are reward bridge functions. This is a part of the statement in \pref{thm:key_rl}. 

\begin{lemma}[Learnable reward bridge functions are reward bridge functions]\label{lem:v_recursive}
\begin{align*}
   \E[b'_{V}(A,O)\mid A,O^{-}] &=\E[R \epol(a\mid O)\mid A,O^{-}]+  \gamma\E\bracks{\sum_{a'}b'_{V}(a',O^{+})\epol(A \mid O) \mid  A,O^{-}}. 
 \end{align*}
\end{lemma}
\begin{proof}
From \pref{lem:value2}, we have 
  \begin{align*}
   \E[b'_{V}(A,O)\mid S,A] &=\E[R \epol(a\mid O)\mid A,S]+  \gamma\E\bracks{\sum_{a'}\E[b'_{V}(a',O^{+})|S^{+}]\epol(A \mid O) \mid A,S}\\
    &=\E[R \epol(a\mid O)\mid A,S]+  \gamma\E\bracks{\sum_{a'}b'_{V}(a',O^{+})\epol(A \mid O) \mid A,S}. 
 \end{align*}
Then, from $O,R,O^{+}\perp O^{-} \mid S,A$, we have
\begin{align*}
&\E[\E[b'_{V}(A,O)\mid S,A]|A, O^{-}] =    \E[\E[b'_{V}(A,O)\mid S,A,O^{-}]|A,O^{-}]= \E[b'_{V}(A,O)\mid A,O^{-}]\\ 
&\E[\E[R \epol(A\mid O)\mid A,S]|A,O^{-}]  = \E[\E[R \epol(A\mid O)\mid A,S,O^{-}]|A,O^{-}]=\E[R \epol(A\mid O)\mid A,O^{-}]\\
&\E\bracks{\bracks{\sum_{a'}b'_{V}(a',O^{+})\epol(A \mid O) \mid A,S}\mid A,O^-}=\E\bracks{\sum_{a'}b'_{V}(a',O^{+})\epol(A \mid O) \mid  A,O^{-}} . 
 \end{align*}
In conclusion,  
\begin{align*}
   \E[b'_{V}(A,O)\mid A,O^{-}] &=\E[R \epol(A\mid O)\mid A,O^{-}]+  \gamma\E\bracks{\sum_{a'}b'_{V}(a',O^{+})\epol(A \mid O) \mid  A,O^{-}}. 
 \end{align*}
 
\end{proof}

Next, we show learnable weight bridge functions are action bridge functions. This is a part of the statement in \pref{thm:key_rl}. 
Here, letting $w(s)= d_{\epol}(s)/b(s)$, recall 
\begin{align}\label{eq:previous_key}
    \E\bracks{\gamma w(S)\frac{\epol(A|O)}{\bpol(A|S)}f(S')- w(S)f(S) }+(1-\gamma)\E_{\tilde S\sim \nu}[f(\tilde S)]=0,\,\forall f:\Scal \to \RR 
\end{align}
from \citet[Lemma 16 ]{KallusNathan2019EBtC} which is a modification of the lemma in \cite{Liu2018}. Recall $\nu$ is an initial distribution. 

\begin{lemma}[Learnable weight bridge functions are weight bridge functions.]\label{lem:learnable_action}
\begin{align*}
    \E\bracks{\gamma  b'_{W}(A,O^{-})\epol(A|O)f(O^{+},a')-\rI(A=a')b'_{W}(A,O^{-}) f(O,A) }+(1-\gamma)\E_{\tilde O\sim \nu_{\Ocal}}[f(\tilde O,a')]=0,\,\forall f:\Scal\to \RR. 
\end{align*}
Similarly, 
\begin{align*}
    \E\bracks{\sum_{a'\in \Acal}\gamma  b'_{W}(A,O^{-})\epol(A|O)f(O^{+},a')-b'_{W}(A,O^{-}) f(O,A) }+(1-\gamma)\E_{\tilde O\sim \nu_{\Ocal}}[\sum_{a'\in \Acal} f(\tilde O,a')]=0,\,\forall f:\Scal\to \RR.  
\end{align*}

\end{lemma}
\begin{proof}
We first prove the first statement: 
\begin{align*}
     & \E\bracks{\gamma  b'_{W}(A,O^{-})\epol(A\mid O) f(O^{+},a')-\rI(A=a') b'_{W}(A,O^{-}) f(O,A) }\\
    &=\E\bracks{\gamma  \E[b'_{W}(A,O^{-})\mid A,S,O^{+},O ]\epol(A\mid O) f(O^{+},a')-\rI(A=a')\E[b'_{W}(A,O^{-})\mid A,S, O]f(O,A) }\\
     &=\E\bracks{\gamma \E[b'_{W}(A,O^{-})\mid A,S  ] \epol(A\mid O) f(O^{+},a')-\rI(A=a')\E[b'_{W}(A,O^{-})\mid A,S ]f(O,A) }\\
    &=\E\bracks{\gamma w(S)/\bpol(A\mid S)\epol(A|O)f(O^{+},a')-w(S)\rI(A=a')/\bpol(A\mid S) f(O,A) } \tag{Definition}\\    
    &=\E\bracks{\gamma w(S)/\bpol(A\mid S)\epol(A|O)\E[f(O^{+},a')\mid S^{+},A,O,S]-w(S)\rI(A=a')/\bpol(A\mid S)\E[f(O,A)\mid S,A=a'] }\\  
    &=\E\bracks{\gamma w(S)\epol(A|O)/\bpol(A\mid S)\E[f(O^{+},a')\mid S^{+}]-w(S)\E[f(O,a')\mid S] }\\
    &= -(1-\gamma)\E_{S\sim \nu}[\E[f(\tilde O,a')\mid S]] \tag{From \pref{eq:previous_key}}\\
    &=  - (1-\gamma)\E_{\tilde O\sim \nu_{\Ocal}}[f(\tilde O,a')]. 
\end{align*}
The second statement is proved by taking summation over the action space. 
\end{proof}

Finally, we show if bridge functions exist, we can identify the policy value. Before proceeding, we prove the following helpful lemma. 
\begin{lemma}[Identification formula with (unlearnable) bridge functions ]\label{lem:fake_identification}
Assume the existence of bridge functions. Then, let $b'_V,b'_W$ be any bridge functions. Then, we have 
\begin{align*}
       J(\epol)= \E_{\tilde O\sim \nu_{\Ocal}}[\sum_a b'_V(a,\tilde O)],\quad     J(\epol)=(1-\gamma)^{-1}\E[b'_W(A,O^-)R\epol(A\mid O)]. 
\end{align*}
\end{lemma}
\begin{proof}
The first formula is proved as follows:
\begin{align*}
    & \E_{\tilde O\sim \nu_{\Ocal}}[\sum_a b'_V(a,\tilde O)]=\E_{S\sim \nu }[\E[\sum_a b'_V(a,\tilde O)\mid S]]=\E_{S\sim \nu }[\E[\sum_a b'_V(a,\tilde O)\mid A=a,S]]\\
    &=\E_{S_0\sim \nu }[\E[\sum_a b'_V(a,\tilde O_0)\mid A_0=a,S_0]]\\
  &=  \E_{S_0\sim \nu}[\sum_{a}    \E_{\epol}[\sum_{t=0}^{\infty} \gamma^t R_t\epol(a\mid O_0)|A_0=a,S_0]] \tag{Definition}\\
    &=  \E_{S_0\sim \nu}[\sum_{a}    \E_{\epol}[\E_{\epol}[\sum_{t=0}^{\infty}  \gamma^t R_t\mid A_0=a,S_0,O_0]\epol(a\mid O_0)|A_0=a,S_0]] \\
       &=  \E_{S_0\sim \nu}[\sum_{a}    q^{\epol}(S_0,a)\E_{\epol}[\epol(a\mid O_0)|A_0=a,S_0]]\\
            &=  \E_{S_0\sim \nu}[\sum_{a}    q^{\epol}(S_0,a)\E_{\epol}[\epol(a\mid O_0)|S_0]]=  \E_{S_0\sim \nu}[\sum_{a}    q^{\epol}(S_0,a)\epol(a\mid O_0)]\\
            &=J(\epol).
\end{align*}
Here, we define $\E_{\epol}[\sum_{t=0}^{\infty} \gamma^t R_t|A_0=a,S_0=s]=q^{\epol}(a,s)$.

The second formula is proved as follows: 
\begin{align*}
\E[b'_{W}(A, O^{-})R\epol(A\mid O)]&= \E[\E[b'_{W}(A, O^{-})\mid A,S,R,O]\epol(A\mid O)R] \\
&= \E[\E[b'_{W}(A, O^{-})\mid A,S]\epol(A\mid O)R] \tag{ $ O^- \perp R,O \mid S,A $} \\
&=\E\bracks{\frac{w(S)}{\bpol(A|S)} \epol(A\mid O)R} \tag{Definition}\\
&=J(\epol). 
\end{align*}
\end{proof}

We are ready to give the proof of the final identification formula \pref{thm:key_rl}. The statement is as follows. 

\begin{theorem}\label{thm:final_rl_ape}
Assume the existence of bridge functions. Let $b_V,b_W$ be any learnable bridge functions. 
Then, 
\begin{align*}
      J(\epol)=\E_{\tilde O\sim \nu_{\Ocal}}[\sum_{a'} b_V(\tilde O,a')],\quad    J(\epol)=(1-\gamma)^{-1}\E[b_W(A,O^-)R\epol(A\mid O)]. 
\end{align*}
\end{theorem}
\begin{proof}
First, we prove the first formula. This is concluded by 
\begin{align*}
    J(\epol)&= (1-\gamma)^{-1}\E[b'_W(A,O^{-})R\epol(A\mid O)] \tag{From \pref{lem:fake_identification}}\\
     &= (1-\gamma)^{-1}\E[b'_W(A,O^{-})\E[R\epol(A\mid O)\mid A,O^-]]\\
     &=(1-\gamma)^{-1} \E\bracks{b'_W(A,O^{-})\E\bracks{\gamma \sum_{a'} b_V(a',O^+)\epol(A\mid O)-b_V(A,O) \mid A,O^-}} \tag{$b'_W$ is also an learnable action bridge function. }\\
     &=\E_{\tilde O\sim \nu_{\Ocal}}[\sum_{a'} b_V(\tilde O,a')].    \tag{From \pref{lem:learnable_action}}
\end{align*}

Next, we prove the second formula. This is concluded by 
\begin{align*}
    J(\epol) &= \E_{\tilde O\sim \nu_{\Ocal}}[\sum_{a'} b'_V(\tilde O,a')] \tag{From \pref{lem:fake_identification}} \\
        &= (1-\gamma)^{-1}\E\bracks{b_W(A,O^{-})\E\bracks{\gamma \sum_{a'} b'_V(a',O^+)\epol(A\mid O)-b'_V(A,O) \mid A,O^-}} \tag{Definition of $b_W$}\\
        &=(1-\gamma)^{-1}\E[b_W(A,O^{-})\E[R\epol(A\mid O)\mid A,O^-]] \tag{$b'_V$ is also an learnable reward function.} \\ 
        &= (1-\gamma)^{-1}\E[b_W(A,O^{-})R\epol(A\mid O)]. 
\end{align*}
\end{proof}

\begin{proof}[Proof of \pref{thm:without0}]

We show any learnable value bridge functions also are value bridge functions. Then, from \pref{lem:fake_identification}, the statement is immediately concluded.

By the assumption $O,R,O^+ \perp O^- \mid S,A$ we can conclude 
any $b_V$ satisfies
\begin{align*}
    \EE\bracks{\EE\bracks{\gamma \sum_{a'}b_V(a',O^+) +R\epol(A\mid O)-b_V(A,O)\mid S,A}\mid A,O^-}=0.  
\end{align*}
From the completeness assumption, 
\begin{align*}
    \EE\bracks{\gamma \sum_{a'}b_V(a',O^+) +R\epol(A\mid O)-b_V(A,O)\mid S,A}=0. 
\end{align*}
Then, from a fixed-point theorem, this implies
\begin{align*}
     \EE[b_V(a,O)\mid S,A] = q_{\pi}(s,a). 
\end{align*}
Thus, $b_V$ is a value bridge function.

\end{proof}

\subsection{Proof of Section \ref{subsec:theory_value}}

\begin{proof}[Proof of \pref{thm:con_results}]
By simple algebra, the estimator is written as 
\begin{align*}
   \hat b_V= \inf_{f\in \Vcal}\sup_{g\in \Vcal^{\dagger}}\E_{\Dcal}[(\Zcal f)^2]- \E_{\Dcal}[(\Zcal f-g(A,O^-))^2] 
\end{align*}
where $\Zcal f=R\epol(A \mid O)+\sum_{a'}f(a',O^+)\epol(A\mid O)-f(A,O)$. In this proof, we define
\begin{align*}
   \bar C= \max(C_{\Vcal},C_{\Vcal^{\dagger}}, 1). 
\end{align*}
Furthermore, $c$ is some universal constant.   

\paragraph{Show the convergence of inner maximizer}

For fixed $f\in \Vcal$, we define 
\begin{align*}
    \hat g_f=\sup_{g\in \Vcal^{\dagger}}- \E_{\Dcal}[(\Zcal f-g(A,O^-))^2], 
\end{align*}
we want to prove with probability $1-\delta$: 
\begin{align*}
  \forall f\in \Vcal:  |\E_{\Dcal}[(\Zcal f-\hat g_f(A,O^-))^2]-\E_{\Dcal}[(\Zcal f-\Tcal f)^2(A,O^-)]|\leq \bar C^2\frac{c \log(|\Vcal||\Vcal^{\dagger}|c/\delta)}{n}. 
\end{align*}
Note since $\Tcal f$ is the Bayes optimal regressor, we have 
\begin{align}\label{eq:regressor}
    \E[(\Zcal f-\Tcal f(A,O^-))^2-(\Zcal f-g(A,O^-))^2]=\E[(\Tcal f-g(A,O^-))^2 ]. 
\end{align}
Hereafter, to simplify the notation, we often drop $(A,O^-)$. Then, from Bernstein's inequality, with probability $1-\delta$, 
\begin{align}\label{eq:uniform}
  \forall f\in \Vcal,\forall  g\in \Vcal^{\dagger}:  |\{\E_{\Dcal}-\E\}[(\Zcal f-\Tcal f)^2-(\Zcal f-g)^2]|\leq c \bar C\sqrt{\frac{\E[(\Tcal f-g)^2] \log(|\Vcal||\Vcal^{\dagger}|c/\delta)}{n}}+\frac{c \bar C^2\log(|\Vcal||\Vcal^{\dagger}|c/\delta)}{n}.
\end{align}
Hereafter, we condition on the above event. In addition, from Bellman closedness assumption $\Tcal \Vcal\subset \Vcal^{\dagger}$, 
\begin{align}\label{eq:bellman}
    \E_{\Dcal}[(\Zcal f-\hat g_f)^2-(\Zcal f-\Tcal f)^2]\leq 0. 
\end{align}
Then, 
\begin{align*}
  \E[(\Tcal f-\hat g_f)^2 ]&=\E[(\Zcal f-\Tcal f)^2-(\Zcal f-\hat g_f)^2] \tag{From \pref{eq:regressor} }\\
  &\leq |\{\E_{\Dcal}-\E\}[(\Zcal f-\Tcal f)^2-(\Zcal f-\hat g_f)^2]|+\E_{\Dcal}[(\Zcal f-\hat g_f)^2-(\Zcal f-\Tcal f)^2] \\
  &\leq  |\{\E_{\Dcal}-\E\}[(\Zcal f-\Tcal f)^2-(\Zcal f-\hat g_f)^2]| \tag{From \pref{eq:bellman}} \\
  &\leq c \bar C\sqrt{\frac{\E[(\Tcal f-\hat g_f)^2] \log(|\Vcal||\Vcal^{\dagger}|c/\delta)}{n}}+\frac{c \bar C^2\log(|\Vcal||\Vcal^{\dagger}|c/\delta)}{n}. \tag{From \pref{eq:uniform}}
\end{align*}
This concludes
\begin{align}\label{eq:first_conclusion}
      \E[(\Tcal f-\hat g_f)^2 ]\leq c\mathrm{Error},\quad \mathrm{Error}\coloneqq  \frac{ \bar C^2\log(|\Vcal||\Vcal^{\dagger}|c/\delta)}{n}.
\end{align}
Then,  
\begin{align*}
 |\E_{\Dcal}[(\Zcal f-\hat g_f)^2]-\E_{\Dcal}[(\Zcal f-\Tcal f)^2]|    & \leq \E[(\Zcal f-\Tcal f)^2-(\Zcal f-\hat g_f)^2]+ c\mathrm{Error} \tag{ From \pref{eq:uniform} and from \pref{eq:first_conclusion}}\\
    & = \E[(\Tcal f-\hat g_f)^2 ]+ c\mathrm{Error} \tag{From \pref{eq:regressor}}\\
    &\leq 2c\mathrm{Error}.  \tag{From \pref{eq:first_conclusion}}
\end{align*}

\paragraph{Show the convergence of outer minimizer}

From the first observation, we can see
\begin{align}\label{eq:second_conlcusion}
    \E_{\Dcal}[(\Zcal \hat b_V)^2-(\Zcal \hat b_V-\Tcal\hat b_V(A,O^-))^2 ]\leq  \E_{\Dcal}[(\Zcal b_V)^2-(\Zcal b_V-\Tcal b_V(A,O^-))^2 ]+c\mathrm{Error}. 
\end{align}
Note 
\begin{align*}
    \E[(\Zcal f)^2-(\Zcal f-\Tcal f(A,O^-))^2-\{(\Zcal b_V)^2-(\Zcal b_V-\Tcal b_V(A,O^-))^2\} ]= \E[(\Tcal f)(A,O^-)^2] 
\end{align*}
Here,  from Bernstein's inequality, with probability $1-\delta$, 
\begin{align}
    \forall f\in \Vcal, &(\E_{\Dcal}-\E)[(\Zcal f)^2-(\Zcal f-\Tcal f)^2-\{(\Zcal b_V)^2-(\Zcal b_V-\Tcal b_V)^2\} ] \label{eq:bernstein_con1}\\
     &\leq \bar C\sqrt{\E[(\Tcal f)^2(A,O^-)] \frac{ \log(|\Vcal|/\delta)}{n} }+\frac{\bar C^2\log(|\Vcal|/\delta)}{n}. \label{eq:bernstein_con2}
\end{align}
Hereafter, we condition on the above event. Then, 
\begin{align*}
    \E[(\Tcal \hat b_V)^2(A,O^{-})]&= \E[(\Zcal \hat b_V)^2-(\Zcal \hat b_V-\Tcal \hat b_V)^2-\{(\Zcal b_V)^2-(\Zcal b_V-\Tcal b_V)^2\} ]\\
    &\leq |(\E_{\Dcal}-\E)[(\Zcal \hat b_V)^2-(\Zcal \hat b_V- \Tcal \hat b_V)^2-\{(\Zcal b_V)^2-(\Zcal b_V-\Tcal b_V)^2\} ] \\ 
    & + \E_{\Dcal}[(\Zcal \hat b_V)^2-(\Zcal \hat b_V-\Tcal\hat b_V)^2-\{(\Zcal b_V)^2-(\Zcal b_V-\Tcal b_V)^2\} ]|\\
    &\leq  \bar C\sqrt{\E[(\Tcal \hat b_V)^2(A,O^{-})] \frac{ \log(|\Vcal|c/\delta)}{n} }+\frac{\bar C^2\log(|\Vcal|c/\delta)}{n}+c\mathrm{Error}.  
\end{align*}
In the last line, we use \pref{eq:bernstein_con2} and \pref{eq:second_conlcusion}. 
This concludes
\begin{align*}
      \E[(\Tcal \hat b_V)^2(A,O^{-})]\leq c\mathrm{Error}. 
\end{align*}
\end{proof}

\begin{proof}[Proof of \pref{thm:final_policy}]

Recall 
\begin{align*}
        J(\epol)=  \E[b_W(A,O^{-})\E[R\epol(A\mid O)\mid A,O^-] 
\end{align*}
from \pref{thm:final_rl_ape} and the existence of $b_W$. 
Furthermore, 
\begin{align*}
   &(1-\gamma)\E_{\tilde O\sim \nu_{\Ocal}}[\sum_{a'} \hat b_V(\tilde O,a')]  \\
   &=  \E\bracks{b_W(A,O^{-})\E\bracks{\gamma \sum_{a'} \hat b_V(a',O^+)\epol(A\mid O)-\hat b_V(A,O) \mid A,O^-}}.  \tag{$b_W$'s are also observe bridge functions form \pref{thm:key_rl}}
\end{align*}
Thus, this concludes 
\begin{align*}
   & J(\epol)-\E_{\tilde O\sim \nu_{\Ocal}}[\sum_{a'} \hat b_V(\tilde O,a')]\\
     &=(1-\gamma)^{-1}\E\bracks{b_W(A,O^{-})\E\bracks{\gamma \sum_{a'} \hat b_V(a',O^+)\epol(A\mid O)+R\epol(A\mid O)-\hat b_V(A,O) \mid A,O^-}}\\
     &= (1-\gamma)^{-1}\E\bracks{b_W(A,O^{-}) \Tcal \hat b_V (A,O^-)}. 
\end{align*}
Then, from CS inequality, we have
\begin{align*}
    |J(\epol)-\E_{\tilde O\sim \nu_{\Ocal}}[\sum_{a'} \hat b_V(\tilde O,a')]|\leq (1-\gamma)^{-1}\E[b_W(A,O^{-})^2]^{1/2} \E[\{\Tcal \hat b_V(A,O^-)\}^2 ]^{1/2}. 
\end{align*}
This concludes the final statement. 

\end{proof}

\subsection{Proof of Section \ref{sec:theoryDRmethod}}

\begin{proof}[Proof of \pref{thm:dr_infinite}]
The first statement $J(\epol)=\E[J(f,b_V)]$ is proved by \pref{lem:learnable_action}:
\begin{align*}
    \E[J(f,b_V)] &= \E_{\tilde O\sim \nu_{\Ocal}}[\sum_a b_V(a,\tilde O)]+ \E\bracks{\frac{f(A,O^{-})}{1-\gamma}\{\{R+\gamma \sum_{a'}b_V(a',O^{+})\}\epol(A\mid O)-b_V(A,O)\}}\\ 
     &= \E_{\tilde O\sim \nu_{\Ocal}}[\sum_a b_V(a,\tilde O)] \tag{Definition of reward bridge functions}\\ 
     & =J(\epol).  \tag{From \pref{thm:final_rl_ape} }
\end{align*}

The second statement $J(\epol)=\E[J(b_W,f)]$ is proved by  \pref{lem:v_recursive}:
\begin{align*}
      \E[J(b_W,f)] &= \E_{\tilde O\sim \nu_{\Ocal}}[\sum_a f(a,\tilde O)]+ \E\bracks{\frac{b_W(A,O^{-})}{1-\gamma}\{\{R+\gamma \sum_{a'}f(a',O^{+})\}\epol(A\mid O)-f(A,O)\}}\\ 
      &= (1-\gamma)^{-1}  \E[R\epol(A\mid O)b_W(A,O^{-})  ] \tag{Definition of weight bridge functions} \\ 
      & =J(\epol).  \tag{From \pref{thm:final_rl_ape} }
\end{align*}
\end{proof}

\begin{proof}[Proof of  \pref{thm:cramer}]

Recall we have the observation: 
\begin{align*}
  Z=  \{O^{-},A,O,R,O^{+}\}. 
\end{align*}
We define 
\begin{align*}
    \tau_k =\prod_{t=1}^k \epol_t(a_t \mid o_t),\quad     \tau_{a:b} =\prod_{t=a}^b \epol_t(a_t \mid o_t). 
\end{align*}

From \cite{nair2021spectral,tennenholtz2020off}, the target functional is
{\small
\begin{align*}
    J=\sum_{k=0} \gamma^k \sum_{\mathfrak{j}_{r_k}}r_k \tau_k \Prr(r_k, o_k \mid a_k,\bO_{k-1})\{\Prr(\bO_k \mid a_k, \bO_{k-1}\}^{-1}\Prr(\bO_k,o_{k-1}\mid a_{k-1},\bO_{k-2})\{\Prr(\bO_{k-1} \mid a_{k-1}, \bO_{k-2})\}^{-1}\cdots \Prr(\bO_0). 
\end{align*}
}
We use the assumption $\rank(\Pr(\bO \mid A=a, \bO^{-}))=|\Scal|=|\Ocal|$, i.e., $\Pr(\bO \mid A=a, \bO^{-})$ is full-rank. 

We will show the existence of $\phi_{\mathrm{EIF}}(Z)$ s.t. 
\begin{align*}
     \nabla J(\theta) = \E[ \phi_{\mathrm{EIF}}(Z)\nabla \log P(Z)  ]. 
\end{align*}
Then, $\E[\phi_{\mathrm{EIF}}(Z)\phi^{\top}_{\mathrm{EIF}}(Z)] $ is the Cram\'er-Rao lower bound. 

Before the calculation, we introduce the bridge functions. Due to the assumptions, these are unique. By letting $\mathfrak{j}_{k}\coloneqq \{o_0, a_0, r_0, a_1, o_1, r_1, \cdots,r_k\}$, we define $b_{V}(a,o)$ as 
{\small
\begin{align*}
     b_{V}(a_0,o_0)=\sum_{k=0}^{\infty} \gamma^k \sum_{\mathfrak{j}_{r_k}\backslash (a_0,o_0) }r_k \tau_k \Prr(r_k, o_k \mid a_k,\bO_{k-1})\{\Prr(\bO_k \mid a_k, \bO_{k-1}\}^{-1}\cdots \{\Prr(\bO_{0} \mid a_{0}, \bO_{-1})\}^{-1} \rI(\bO_0=o_0). 
\end{align*}
}
Next, we define $b_{W}(a,o^{-})$. Recall 
\begin{align*}
    d^{\pi^e}_k(o_k)=\sum_{\mathfrak{j}_{o_k} \backslash o_k} \tau_{k-1} \rI(\bO_k=o_k)^{\top}\Prr(\bO_k,o_{k-1}\mid a_{k-1},\bO_{k-2})\{\Prr(\bO_{k-1} \mid a_{k-1}, \bO_{k-2})\}^{-1}\cdots \Prr(\bO_0). 
\end{align*}
where $\mathfrak{j}_{o_k}\coloneqq \{o_0, a_0, r_0, a_1, o_1, r_1, \cdots,o_k\}$. 
Then, we define $b_{W}(a,o^{-})$ as 
\begin{align*}
    b_{W}(a,o^-)=\frac{\rI(\bO^{-}=o^{-})}{\Prr(a,o^{-})} \{\Prr(\bO\mid a,\bO^{-} \}^{-1}  d^{\pi^e}(\bO),\quad     d^{\pi^e}(o)=\sum_{k=0}^{\infty}\gamma^k     d^{\pi^e}_k(o)/(1-\gamma) . 
\end{align*}

Now, we are ready to calculate the Cram\'er-Rao lower bound. We have 
\begin{align*}
    \nabla     J(\theta)= B_1 + B_2+B_3
\end{align*}
where $B_1$, $B_2$ and $B_3$ are given by
{\small
\begin{align*}
    \sum_{k=0} \gamma^k \sum_{\mathfrak{j}_{r_k}}r_k \tau_k \nabla  \Prr(r_k, o_k \mid a_k,\bO_{k-1})\{\Prr(\bO_k \mid a_k, \bO_{k-1}\}^{-1}\Prr(\bO_k,o_{k-1}\mid a_k,\bO_{k-2})\{\Prr(\bO_{k-1} \mid a_{k-1}, \bO_{k-2})\}^{-1}\cdots \Prr(\bO_0),\\
    \sum_{\mathfrak{j}}\sum_{k=0} \gamma^k \sum_{t=0}^k r_k\tau_k \Prr(r_k, o_k \mid a_k,\bO_{k-1})\{\Prr(\bO_k \mid a_k, \bO_{k-1}\}^{-1}\cdots  \Prr(\bO_{t+1},o_{t}\mid a_{t},\bO_{t-1})\nabla \{\Prr(\bO_{t} \mid a_{t}, \bO_{t-1})\}^{-1}\cdots \Prr(\bO_0), \\
    \sum_{\mathfrak{j} } \sum_{k=0} \gamma^k \sum_{t=0}^{k-1} r_k\tau_k \Prr(r_k, o_k \mid a_k,\bO_{k-1})\{\Prr(\bO_k \mid a_k, \bO_{k-1}\}^{-1}\cdots  \nabla \Prr(\bO_{t+1},o_{t}\mid a_{t},\bO_{t-1})\{ \Prr(\bO_{t} \mid a_{t}, \bO_{t-1})\}^{-1}\cdots \Prr(\bO_0), 
\end{align*}
}respectively. Here, we use the notation $\mathfrak{j}=\{s_0,a_0,r_0,\cdots \}$. 

We first analyze $B_1$: It is equal to
{\small 
\begin{align*}
    &\sum_{k=0} \gamma^k \sum_{\mathfrak{j}_{r_k}}r_k \tau_k \nabla  \Prr(r_k, o_k \mid a_k,\bO_{k-1})\{\Prr(\bO_k \mid a_k, \bO_{k-1}\}^{-1}\Prr(\bO_k,o_{k-1}\mid a_k,\bO_{k-2})\{\Prr(\bO_{k-1} \mid a_{k-1}, \bO_{k-2})\}^{-1}\cdots \Prr(\bO_0)\\
    &=(1-\gamma)^{-1} \sum_{r,a,o } r\epol(a\mid o)\nabla \Prr(r,o\mid a, \bO^{-})\{b_{W}(a,\bO^{-})\odot \Prr(a,\bO^{-}) \} \\  
    &= (1-\gamma)^{-1}\sum_{r,a,o,o^{-} } r\epol(a\mid o)\nabla  \Prr(r,o\mid a, o^{-}) b_{W}(a,o^{-})\Prr(a,o^{-})\\
    &=(1-\gamma)^{-1}\sum_{r,a,o,o^{-} } r\epol(a\mid o)\{\nabla \log \Prr(r,o\mid a, o^{-})\} b_{W}(a,o^{-})\Prr(r,o,a,o^{-}), 
\end{align*}
}
where $\odot$ is an element-wise product. 
Then, we have
\begin{align*}
  (1-\gamma)  B_1  &=  \E[b_{W}(A,O^{-}) R\epol(A\mid O) \nabla \log \Prr(R, O \mid A, O^{-})  ]  \\ 
    &= \E[b_{W}(A,O^{-})\{ R\epol(A\mid O)-\E[ R\epol(A\mid O)\mid A,O^{-}]  \} \nabla \log \Prr(R, O \mid A, O^{-})  ]\\
     &=\E[b_{W}(A,O^{-})\{ R\epol(A\mid O)-\E[ R\epol(A\mid O)\mid A,O^{-}]  \} \nabla \log \Prr(R, O, A, O^{-})  ]\\
         &=  \E[b_{W}(A,O^{-})\{ R\epol(A\mid O)-\E[ R\epol(A\mid O)\mid A,O^{-}]  \} \nabla \log \Prr(O^{+},R, O, A, O^{-})  ].
\end{align*}

Next, we analyze $B_2$: 
{\small 
\begin{align*}
    &\sum_{\mathfrak{j}}\sum_{k=0} \gamma^k \sum_{t=0}^k r_k\tau_k \Prr(r_k, o_k \mid a_k,\bO_{k-1})\{\Prr(\bO_k \mid a_k, \bO_{k-1}\}^{-1}\cdots  \Prr(\bO_{t+1},o_{t}\mid a_{t},\bO_{t-1})\nabla \{\Prr(\bO_{t} \mid a_{t}, \bO_{t-1})\}^{-1}\cdots \\ 
    &\times \Prr(\bO_0)=\sum_{\mathfrak{j}} \sum_{t=0} \gamma^{t} \sum_{k=t}^{\infty} \gamma^{k-t}\tau_{k} r_k \Prr(r_k, o_k \mid a_k,\bO_{k-1})\{\Prr(\bO_k \mid a_k, \bO_{k-1}\}^{-1}\cdots  \Prr(\bO_{t+1},o_{t}\mid a_{t},\bO_{t-1}) \\ 
   &\times \nabla \{ \Prr(\bO_{t} \mid a_{t}, \bO_{t-1})\}^{-1} \cdots \Prr(\bO_0)=\sum_{\mathfrak{j}} \sum_{t=0} \gamma^{t} \sum_{k=t}^{\infty} \gamma^{k-t}\tau_{k} r_k \Prr(r_k, o_k \mid a_k,\bO_{k-1})\{\Prr(\bO_k \mid a_k, \bO_{k-1}\}^{-1}\cdots  \\
   &\times \Prr(\bO_{t+1},o_{t}\mid a_{t},\bO_{t-1}) \{ \Prr(\bO_{t} \mid a_{t}, \bO_{t-1})\}^{-1}\{\nabla \Prr(\bO_{t} \mid a_{t}, \bO_{t-1})\}\{ \Prr(\bO_{t} \mid a_{t}, \bO_{t-1})\}^{-1}\cdots \Prr(\bO_0) \\
     &=-\sum_{t=0} \gamma^{t}\tau_t b_{V}(a_t, \bO_t)\{\nabla \Prr(\bO_{t} \mid a_{t}, \bO_{t-1})\}\{ \Prr(\bO_{t} \mid a_{t}, \bO_{t-1})\}^{-1}\cdots \Prr(\bO_0) \\ 
    &=-(1-\gamma)^{-1}\sum_{a} b_{V}(a,\bO)\{\nabla \Prr(\bO\mid a,\bO^{-})\{b_{W}(a,\bO^{-})\odot \Prr(a,\bO^{-}) \}  \}. 
\end{align*}
}
Further, we have 
\begin{align*}
    (1-\gamma)B_2 &= \E[b_{W}(A,O^{-})b_{V}(A,O)\nabla \log \Pr(O\mid A,O^{-}) ] \\ 
     &= \E[b_{W}(A,O^{-})b_{V}(A,O)\{\nabla \log \Pr(O\mid A,O^{-})+ \nabla \log \Pr(R,O^{+} \mid A,O^{-},O)\} ]\\
   &= \E[b_{W}(A,O^{-}) b_{V}(A,O)\nabla \log \Pr(R,O^{+},O \mid A,O^{-}) ] \\
     &= \E[b_{W}(A,O^{-})\{b_{V}(A,O)-\E[b_{V}(A,O)\mid A,O^{-}] \}\nabla \log \Pr(R,O^{+},O \mid A,O^{-}) ] \\ 
     &= \E[b_{W}(A,O^{-})\{b_{V}(A,O)-\E[b_{V}(A,O)\mid A,O^{-}] \}\nabla \log \Pr(R,O^{+},O, A,O^{-}) ]. 
\end{align*}

Finally, we analyze $B_3$: 
{\small 
\begin{align*}
  & \sum_{\mathfrak{j} } \sum_{k=0} \gamma^k \sum_{t=0}^{k-1} r_k\tau_k \Prr(r_k, o_k \mid a_k,\bO_{k-1})\{\Prr(\bO_k \mid a_k, \bO_{k-1}\}^{-1}\cdots  \nabla \Prr(\bO_{t+1},o_{t}\mid a_{t},\bO_{t-1})\{ \Prr(\bO_{t} \mid a_{t}, \bO_{t-1})\}^{-1}\\
 &\cdots \times \Prr(\bO_0)=\sum_{\mathfrak{j} }  \sum_{t=0} \gamma^{t}\tau_{t} \sum_{k=t+1}^{\infty} \gamma^{k-t}r_k\tau_{t+1:k} \Prr(r_k, o_k \mid a_k,\bO_{k-1})\{\Prr(\bO_k \mid a_k, \bO_{k-1}\}^{-1}\cdots  \nabla \Prr(\bO_{t+1},o_{t}\mid a_{t},\bO_{t-1}) \\
  & \times\{\Prr(\bO_{t} \mid a_{t}, \bO_{t-1})\}^{-1}\cdots \Prr(\bO_0)=\sum_{\mathfrak{j}}\sum_{t=0} \gamma^{t}\tau_{t}  b_{a_{t+1}}(\bO_{t+1})   \nabla \Prr(\bO_{t+1},o_{t}\mid a_{t},\bO_{t-1})\{ \Prr(\bO_{t} \mid a_{t}, \bO_{t-1})\}^{-1}\cdots \Prr(\bO_0)\\
  &=(1-\gamma)^{-1}\sum_{\mathfrak{j}}\gamma \pi^e(a \mid o)b_{V}(a^{+}, \bO^{+})\nabla \Prr(\bO^{+},o\mid a,\bO^{-})\{b_{W}(A,\bO^{-})\odot \Prr(a,\bO^{-})\}\\
  &=(1-\gamma)^{-1}\sum_{a^+,o^+,o,a,o^{-}}\gamma \pi^e(a \mid o)b_{V}(a^{+}, o^{+})\nabla \log P(o^{+},o\mid a,o^{-})b_{W}(a,o^{-})\Prr(o^{+},o,a,o^{-}). 
\end{align*}
}
Thus, $ (1-\gamma)B_3$ is equal to 
{\small 
\begin{align*}
    &\gamma \E[b_{W}(A,O^{-})\pi^e(A \mid O)\sum_{a^+}b_{V}(a^{+}, O^{+})\nabla \log P(O^{+},O \mid A,O^{-}) ]\\
    &=\gamma \E[b_{W}(A,O^{-})\pi^e(A \mid O)\sum_{a^+}b_{V}(a^{+}, O^{+}) \nabla \log P(R,O^{+},O \mid A,O^{-}) ]\\
   &=\gamma \E[b_{W}(A,O^{-})\{\pi^e(A \mid O)\sum_{a^+}b_{V}(a^{+}, O^{+}) -\E[\pi^e(A \mid O)\sum_{a^+}b_{V}(a^{+}, O^{+}) \mid A,O^{-} ] \} \nabla \log P(R,O^{+},O \mid A,O^{-}) ]\\
     &=\gamma \E[b_{W}(A,O^{-})\{\pi^e(A \mid O)\sum_{a^+}b_{V}(a^{+}, O^{+})- \E[\pi^e(A \mid O)\sum_{a^+}b_{V}(a^{+}, O^{+}) \mid A,O^{-} ] \} \nabla \log P(R,O^{+},O, A,O^{-}) ]. 
\end{align*}
}
Combining all things together, $(1-\gamma)(B_1+B_2+B_3)$ is  
\begin{align*}
    &  \E[b_{W}(A,O^{-})\{ R\epol(A\mid O)-\E[ R\epol(A\mid O)\mid A,O^{-}]  \} \nabla \log \Prr(O^{+},R, O, A, O^{-})  ] \\
    &+\E[b_{W}(A,O^{-})\{b_{V}(A,O)-\E[b_{V}(A,O)\mid A,O^{-}] \}\nabla \log \Pr(R,O^{+},O, A,O^{-}) ]  \\ 
    &+\gamma \E[b_{W}(A,O^{-})\{\pi^e(A \mid O)\sum_{a^+}b_{V}(a^{+}, O^{+})- \E[\pi^e(A \mid O)\sum_{a^+}b_{V}(a^{+}, O^{+}) \mid A,O^{-} ]  \} \nabla \log P(R,O^{+},O, A,O^{-}) ]\\
    &=\E[b_{W}(A,O^{-})\{ R\epol(A\mid O)+\gamma \epol(A\mid O)\sum_{a^+}b_{V}(a^{+}, O^{+}) - b_{V}(A,O)  \}\nabla \log P(R,O^{+},O, A,O^{-})  ]. 
\end{align*}
Hence, the following is $\phi_{\mathrm{EIF}}$: 
\begin{align*}
 \phi_{\mathrm{EIF}}= (1-\gamma)^{-1}b_{W}(A,O^{-})\{ R\epol(A\mid O)+ \gamma \epol(A\mid O)\sum_{a^+}b_{V}(a^{+}, O^{+}) - b_{V}(A,O)  \}. 
\end{align*}
\end{proof}

\begin{proof}[Proof of \pref{thm:efficiency}]

By some algebra, we have 
\begin{align*}
    J(f,g)-  J(b_V,b_W)      =B_1+B_2+B_3 
 \end{align*}   
 where 
 {\small 
 \begin{align*}
      B_1 &= \{f(a,o^-)-b_W(a,o^-) \}\{\{\gamma \sum_{a'} \{g(a',O^+)-b_V(a',o^+)\}\}\epol(a\mid o)-g(a,o)+b_V(a',o^+)\},  \\
     B_2 &= \E_{\tilde o \sim \nu_{\Ocal}}[\sum_{a'}g(a,\tilde o)]-\E_{\tilde o \sim \nu_{\Ocal}}[\sum_{a'}b_V(a,\tilde o)]+ b_W(a,o^-)\{\gamma \sum_{a'} \{g(a',o^+)-b_V(a',o^+)\}\epol(a\mid o)-g(a,o)+b_V(a,o)\},  \\
    B_3 &= \{b_W(a,o^-)-f(a,o^-)\}\{\{r+\gamma \sum_{a'} b_V(a',o^+)\}\epol(a\mid o)-b_V(a,o)\}\}.
\end{align*}
}
Recall the estimator is constructed as 
\begin{align*}
    \hat J^{*}_{\mathrm{DR}}=0.5 \E_{\Dcal_1}[J(\hat b^{(1)}_W,\hat b^{(1)}_V)]+  0.5 \E_{\Dcal_2}[J(\hat b^{(2)}_W,\hat b^{(2)}_V)]. 
\end{align*}
We analyze $ \E_{\Dcal_1}[J(\hat b^{(1)}_W,\hat b^{(1)}_V)]-J(\epol)$. This is expanded as 
\begin{align*}
    \sqrt{n}(\E_{\Dcal_1}[J(\hat b^{(1)}_W,\hat b^{(1)}_V)]-J(\epol))&=\sqrt{n/n_1}\mathbb{G}_{\mathcal{D}_1} [J(\hat b^{(1)}_W,\hat b^{(1)}_V)-J( b_W,b_V)  ]\\ 
    &+ \sqrt{n/n_1} \mathbb{G}_{\mathcal{D}_1} [J( b_W,b_V) ] \\
    & + \sqrt{n/n_1} (\E[ J( \hat b^{(1)}_W,\hat b^{(1)}_V ) \mid \Dcal_2 ] -J(\epol)  ) 
\end{align*}
where $\mathbb{G}_{\mathcal{D}_1} =\sqrt{n_1}(\E_{\Dcal_1}-\E)$. In the following, we analyze each term.  

\paragraph{Analysis of $\sqrt{n/n_1}\mathbb{G}_{\mathcal{D}_1} [J(\hat b^{(1)}_W,\hat b^{(1)}_V)-J( b_W,b_V)  ]$. }

We show that the conditional variance given $\Dcal_2$ is $o_p(1)$. Then, the rest of the argument is the same as the proof of \citet[Theorem 7]{KallusNathan2019EBtC}. This is proved by 
\begin{align*}
    \mathrm{var}[\sqrt{n_1}\E_{\mathcal{D}_1} [J(\hat b^{(1)}_W,\hat b^{(1)}_V)] \mid \Dcal_2]=\E[B^2_1 + B^2_2 + B^2_3 + 2B_1B_2 + 2B_1B_3 + 2B_1B_2 \mid \Dcal_2 ]=o_p(1). 
\end{align*}
For example, $\E[B^2_3 \mid \Dcal_2]=o_p(1)$ is proved by 
\begin{align*}
    \E[B^2_3 \mid \Dcal_2]\lesssim \E[(\hat b_W-b_W)^2(A,O^-)]\E[\{\Tcal\hat b_V\}^2(A,O^-)]=o_p(1). 
\end{align*}

\paragraph{Analysis of $\sqrt{n/n_1} (\E[ J( \hat b^{(1)}_W,\hat b^{(1)}_V) \mid \Dcal_2  ] -J(\epol)  )$. }

Here,we have
\begin{align*}
&\E[ J( \hat b^{(1)}_W,\hat b^{(1)}_V) \mid \Dcal_2  ]=    |\E[B_1 + B_2 + B_3 \mid \Dcal_2 ]|\\
&= |\E[B_3 \mid \Dcal_2 ]|\leq \E[B^2_3 \mid \Dcal_2 ]^{1/2}\leq \E[(\hat b_W-b_W)^2(A,O^-)]^{1/2}\E[\{\Tcal\hat b_V\}^2(A,O^-)]^{1/2}\\
&=o_p(n^{-1/2}_1) \tag{From convergence rate condition}. 
\end{align*}

\paragraph{Combine all things. }
Thus, 
\begin{align*}
     \sqrt{n}(\E_{\Dcal_1}[J(\hat b^{(1)}_W,\hat b^{(1)}_V)]-J(\epol))=  \sqrt{n/n_1}\mathbb{G}_{\mathcal{D}_1} [J( b_W,b_V) ]+o_p(n^{-1/2}). 
\end{align*}
This immediately concludes 
\begin{align*}
     \sqrt{n}(\hat J^{*}_{DR}-J(\epol))=    \mathbb{G}_n[J( b_W,b_V) ]+o_p(n^{-1/2})
\end{align*}
where $\mathbb{G}_n=\sqrt{n}(\E_{\Dcal}-\E)$. 
\end{proof}

\subsection{Proof of \pref{app:morePOMDP}}

\begin{proof}[Proof of \pref{thm:without}]
From Hoeffding inequality, we use with probability $1-\delta$: 
\begin{align} \label{eq:hoeff}
  \forall f\in \Vcal, \forall g\in \Vcal^{\dagger}; | (\E-\E_{\Dcal})\bracks{f(A,O^{-})\Zcal g(A,R,O,O^+))  }|\leq c \{\max(1,C_{\Vcal},C_{\Vcal^{\dagger}}) \}\sqrt{\log(|\Vcal||\Vcal^{\dagger}|/\delta)/n}. 
\end{align}
Recall $\Zcal:g(a,o) \mapsto r\epol(a\mid o)+\sum_{a'} g(a',O^+)\epol(a\mid o)-g(a,o)$. Hereafter, we condition this event.

In this proof, we often use the following:
\begin{align*}
    \E[f(A,O^{-})(\Zcal g)(A,R,O,O^+) ] =   \E[f(A,O^{-})\Tcal g(A,O^{-})]. 
\end{align*}
Recall $\Tcal:g \mapsto \E[R\epol(A\mid O)+\sum_{a'} f(a',O^+)\epol(a\mid O)-g(A,O)\mid A=a,O^-=o] $. 

As a first step, we have 
\begin{align*}
    |J(\epol)- \hat J_{\VM}|&\leq \left|(1-\gamma)^{-1}\E\bracks{b_W(A,O^{-})\Zcal \hat b_V(A,R,O,O^+) }\right|. 
\end{align*}
Here, we have 
\begin{align*}
   & \left|\E\bracks{b_W(A,O^{-})\Zcal \hat b_V(A,R,O,O^+) }\right| \\
    &\leq \left|(\E-\E_{\Dcal})\bracks{b_W(A,O^{-}) \Zcal \hat b_V(A,R,O,O^+)}\right| + \left|\E_{\Dcal}\bracks{b_W(A,O^{-})\Zcal \hat b_V(A,R,O,O^+)}\right| \\ 
    &\leq c\sqrt{\log(|\Vcal||\Vcal^{\dagger}|c/\delta)/n}+\sup_{f\in \Vcal^{\dagger}}\left|\E_{\Dcal}\bracks{f(A,O^{-})\Zcal \hat b_V(A,R,O,O^+)}\right|  \tag{Use \pref{eq:hoeff} and $b_W\in \Vcal^{\dagger}$} \\ 
    &\leq c\sqrt{\log(|\Vcal||\Vcal^{\dagger}|c/\delta)/n}+\sup_{f\in \Vcal^{\dagger}}\left|\E_{\Dcal}\bracks{f(A,O^{-})\Zcal b_V(A,R,O,O^+)}\right| \tag{$b_V\in \Vcal$} \\ 
    & = c\sqrt{\log(|\Vcal||\Vcal^{\dagger}|c/\delta)/n}+\sup_{f\in \Vcal^{\dagger}}\left|\E_{\Dcal}\bracks{f(A,O^{-})\Tcal b_V(A,O^{-})}\right|  \\ 
     &\leq c\sqrt{\log(|\Vcal||\Vcal^{\dagger}|c/\delta)/n}. 
\end{align*}
In the final line,  we use 
\begin{align*}
    \E\bracks{f(A,O^{-})\Tcal b_V(A,O^{-})}=0,\quad \forall f:\Acal\times \Ocal \to \RR; 
\end{align*}
thus, 
\begin{align*}
     & \sup_{f\in \Vcal^{\dagger}}\left|\E_{\Dcal}\bracks{f(A,O^{-})\Zcal b_V(A,R,O,O^+)}\right|\\
    &=\left|\E_{\Dcal}\bracks{f^{\diamond}(A,O^{-})\Zcal b_V(A,R,O,O^+)}\right| \tag{ $f^{\diamond}=\max_{f\in \Vcal^{\dagger}}\left|\E_{\Dcal}\bracks{f(A,O^{-})\Zcal b_V(A,O^{-})}\right|$ } \\
    &\leq \left|\E[\bracks{f^{\diamond}(A,O^{-})\Zcal b_V(A,R,O,O^+)}\right| + c\sqrt{\log(|\Vcal||\Vcal^{\dagger}|c/\delta)/n} \tag{From \pref{eq:hoeff}} \\
    &=  \left|\E[\bracks{f^{\diamond}(A,O^{-})\Tcal b_V(A,O^{-})}\right| + c\sqrt{\log(|\Vcal||\Vcal^{\dagger}|c/\delta)/n}= c\sqrt{\log(|\Vcal||\Vcal^{\dagger}|c/\delta)/n}. 
\end{align*}

\end{proof}

\subsection{Proof of \pref{sec:time_inho}}

We denote $\E[b'^j_V(A_j,O_i)\mid A_j=a, S_j=s]$  by $Q^j(a,s)$. 

\begin{lemma}\label{lem:help}
 \begin{align*}
      Q^j (A_j,S_j) &=\E[R_j \epol_j(A_j \mid O_j)\mid A_j, {S_j}]+  \gamma\E\bracks{\sum_{a'}Q^{j+1}(a',S_{j+1} )\epol_j(A_j \mid O_j) \mid A_j, {S_j}}. 
 \end{align*}
\end{lemma}

\begin{proof}
From the definition, we have 
\begin{align*}
    Q^j (A_j,S_j) &=\E[R_j \epol_j(A_j \mid O_j )\mid S_j,A_j]+  \E_{\epol}\bracks{\sum_{t=j+1}^{H-1}\gamma^{t-j} R_t\epol_j(A_j\mid O_j)\mid A_j,S_j}. 
 \end{align*}  
Then, 
 \begin{align*}
    &\E_{\epol}\bracks{\sum_{t=j+1} \gamma^{t-j} R_t\epol_j(A_j \mid O_j)\mid A_j,S_j}=\E_{\epol}\bracks{\E_{\epol}\bracks{\sum_{t=j+1}\gamma^{t-j}  R_t \mid  A_j,S_{j+1},O_{j+1},O_j,S_j} \epol_j(A_j \mid O_j)\mid A_j,S_j}\\  
    &= \E\bracks{\sum_{a'}\E_{\epol}[\sum_{t=j+1}\gamma^{t-j} R_t\mid S_{j+1}, A_{j+1}=a',O_{j+1},O_j]\epol_{j+1}(a'\mid {O_{j+1}})\epol_j(A_j\mid O_j)\mid A_j,S_j}\\
&= \E\bracks{\sum_{a'}\E_{\epol}[\sum_{t=j+1}\gamma^{t-j} R_t\mid S_{j+1}, A_{j+1}=a']\epol_{j+1}(a'\mid {O_{j+1}})\epol_j(A_j \mid O_j)\mid A_j,S_j}\\
&= \E\bracks{\sum_{a'}\E_{\epol}[\sum_{t=j+1}\gamma^{t-j} R_t\mid S_{j+1}, A_{j+1}=a']\E[\epol_{j+1}(a'\mid {O_{j+1}})\mid S_{j+1}, A_{j+1}=a'] \epol_j(A_j \mid O_j)\mid A_j,S_j}\\
    &=  \E\bracks{\sum_{a'}\E_{\epol}[\sum_{t=j+1}\gamma^{t-j} R_t\epol_{j+1}(a'\mid {O_{j+1}})\mid S_{j+1}, A_{j+1}=a']\epol_j(A_j \mid O_j)\mid A_j,S_j}\\
    &= \gamma \E\bracks{\sum_{a'}Q^{j+1}(a', S_{j+1} )\epol_j(A_j \mid O_j)\mid A_j,S_j}. 
\end{align*}

\end{proof}

\begin{lemma}[Learnable reward bridge functions are reward bridge functions]\label{lem:v_recursive_finite}
\begin{align*}
   \E[b'^{[j]}_{V}(A_j,O_{{j}})\mid A_j,\mathfrak{J}_{j-1} ] &=\E[R_j \epol_j(A_j \mid {O_{j}})\mid A_j,\mathfrak{J}_{j-1} ]  +\gamma\E\bracks{\sum_{a'}b^{'[j+1]}_V(a', {O_{j+1}})\epol_j(A_j \mid {O_{j}}) \mid  A_j,\mathfrak{J}_{j-1} }. 
 \end{align*}
\end{lemma}

\begin{proof}
From \pref{lem:help}, we have 
  \begin{align*}
   & \E[b'^{[j]}_{V}(A_j,O_{j})\mid  A_j,S_j] \\
   &=\E[R_j \epol_j(A_j \mid {O_j})\mid  A_j,S_j]+  \gamma\E\bracks{\sum_{a'}\E[b'^{[j+1]}_V(a', {O_{j+1}})| S_{j+1}, O_{j}]\epol_j(A_j \mid {O_j}) \mid  A_j,S_j}\\
    &=\E[R_j \epol_j(A_j \mid {O_j}) \mid  A_j,S_j]+  \gamma\E\bracks{\sum_{a'}b'^{[j+1]}_V(a', {O_{j+1}})\epol_j(A_j \mid {O_j}) \mid  A_j,S_j}. 
 \end{align*}
Then, from $ O_j, O_{j+1}, R_j\perp \mathfrak{J}_{j-1} \mid  A_j,S_j  $, the statement is concluded by 
  \begin{align*}
   \E[\E[b'^{[j]}_{V}(A_j,O_{j})\mid  A_j,S_j]\mid A_j,\mathfrak{J}_{j-1}  ]&= \E[b'^{[j]}_{V}(A_j,{O_j}) \mid A_j,\mathfrak{J}_{j-1}  ], \\
  \E[\E[R_j \epol_j(A_j \mid {O_j}) \mid  A_j,S_j]  \mid A_j,\mathfrak{J}_{j-1}  ]   &=  \E[R_j \epol_j(A_j \mid {O_j})  \mid A_j,\mathfrak{J}_{j-1}  ],  \\ 
  \E\bracks{\E\bracks{\sum_{a'}b'^{[j+1]}_V(a', {O_{j+1}})\epol_j(A_j \mid {O_j}) \mid  A_j,S_j}\mid A_j,\mathfrak{J}_{j-1}  }&=  \E\bracks{\sum_{a'}b'^{[j+1]}_V(a', {O_{j+1}})\epol_j(A_j \mid {O_j}) \mid A_j,\mathfrak{J}_{j-1}}. 
 \end{align*}

\end{proof}

\begin{lemma}[Learnable weight bridge functions are weight bridge functions]
\begin{align*}
    \E\bracks{  b'^{[j]}_{W}(A_j,\mathfrak{J}_{j-1} )\epol_j(A_j |O_{j})f(O_{j+1},a')-\rI(A_{j+1}=a') b'^{[j+1]}_{W}(A_{j+1},\mathfrak{J}_{j}) f(O_{j+1},A_{j+1})}=0,\,\forall f:\Ocal\times \Acal \to \RR.  
\end{align*}
and 
\begin{align*}
     \E\bracks{   \sum_{a'} b'^{[j]}_{W}(A_j,\mathfrak{J}_{j-1} )\epol_j(A_j |O_{j})f(O_{j+1},a')- b'^{[j+1]}_{W}(A_{j+1},\mathfrak{J}_{j}) f(O_{j+1},A_{j+1})}=0,\,\forall f:\Ocal\times \Acal \to \RR.  
\end{align*}
\end{lemma}
\begin{proof}
We first prove the first statement: 
\begin{align*}
     &  \E\bracks{  b'^{[j]}_{W}(A_j,\mathfrak{J}_{j-1} )\epol_j(A_j |O_{j})f({O_{j+1}},a')-\rI(A_{j+1}=a') b'^{[j+1]}_{W}(A_{j+1},\mathfrak{J}_j) f({O_{j+1}},A_{j+1})}\\
    &= \E\bracks{ \E[ b'^{[j]}_{W}(A_j,\mathfrak{J}_{j-1} )\mid A_j, O_j ,S_j,O_{j+1}]\epol_j(A_j |O_{j})f({O_{j+1}},a')} \\ &-\E\bracks{\rI(A_{j+1}=a') \E[b'^{[j+1]}_{W}(A_{j+1},\mathfrak{J}_j) \mid O_{j+1},A_{j+1},S_{j+1}] f({O_{j+1}},A_{j+1})}\\
     &=\E\bracks{  \mu(S_j)\eta_j(A_j|O_{j})f({O_{j+1}},a')- \rI(A_{j+1}=a') \frac{ \mu(S_{j+1})}{P_{\pi_b}(A_{j+1}\mid S_{j+1})}f({O_{j+1}},A_{j+1})}\\
     &=\E_{\epol}[f(O_{j+1},a')]-\E_{\epol}[f(O_{j+1},a')]\\ 
     &=0. 
\end{align*}
Then, we can prove the second statement by taking summation over the action space. 

\end{proof}

\begin{lemma}[Identification formula with unlearnable bridge functions] \label{lem:fake_finite}

Assume the existence of bridge functions. Then, let $b'_V=\{b'^{[j]}_{V}\},b'_W=\{b'^{[j]}_{W}\}$ be any bridge functions. 
\begin{align*}
    J(\epol)=\E\bracks{\sum_j \gamma^j b'^{[j]}_{W}(A_j,\mathfrak{J}_{j-1} )\epol_j(A_j\mid O_j) R_j },\quad J(\epol)=\E\bracks{\sum_a b'^{[0]}_{V}(a,O_0)  }.
\end{align*}

\end{lemma}

\begin{proof}
We prove the first formula:  
\begin{align*}
 \E\bracks{\sum_j \gamma^j b'^{[j]}_{W}(A_j,\mathfrak{J}_{j-1} )\epol_j(A_j\mid O_j) R_j }       &= \E\bracks{\sum_j \gamma^j \E[b'^{[j]}_{W}(A_j,\mathfrak{J}_{j-1} )\mid A_j,S_j]\epol_j(A_j\mid O_j) R_j } \\ 
         &= \E\bracks{\sum_j \gamma^j\frac{\mu_j(S_j)}{P_{\pi_b}(A_j \mid S_j)}\epol_j(A_j\mid O_j) R_j } \tag{Definition of bridge functions}\\
         &=    J(\epol) .  
\end{align*}

We prove the second formula: 
\begin{align*}
      &  \E\bracks{\sum_a b'^{[0]}_{V}(a,O_0)  }= \E\bracks{\E[\sum_a b'^{[0]}_{V}(a,O_0)\mid S_0]  }= \E\bracks{\E[\sum_a b'^{[0]}_{V}(a,O_0)\mid S_0,A_0=a]  }\\
      &= \E\bracks{\sum_a\E_{\epol}[\sum_{t=0}^{H-1} \gamma^{t}R_t  \epol_0(a\mid O_0) \mid S_0,A_0=a]  } \tag{Definition of bridge functions}\\ 
      &= \E\bracks{\sum_a\E_{\epol}[\sum_{t=0}^{H-1} \gamma^{t}R_t  \mid S_0,A_0=a]\E[\epol_0(a\mid O_0)\mid S_0,A_0=a ] }\\
      &= \E\bracks{\sum_a\E_{\epol}[\sum_{t=0}^{H-1} \gamma^{t}R_t  \mid S_0,A_0=a] \epol_0(a\mid O_0) } \\ &=J(\epol).  
\end{align*}

\end{proof}

\begin{lemma}[Final identification formula] \label{thm:finite_identi}
Assume the existence of bridge functions. Let  $b_V=\{b^{[j]}_{V}\},b_W=\{b^{[j]}_W\}$ be any learnable bridge functions. 

\begin{align*}
    J(\epol)= \E[\sum_a b^{[0]}_{V}(a,O_0) ], \quad    J(\epol)=\E\bracks{\sum_{j=0}^{H-1} \gamma^j b^{[j]}_{W}(A_j,\mathfrak{J}_{j-1} )\epol_j(A_j\mid O_j) R_j }  . 
\end{align*}
\end{lemma}

\begin{proof}
We prove the first formula: 
\begin{align*}
J(\epol) &= \E\bracks{\sum_j \gamma^j b'^{[j]}_{W}(A_j,\mathfrak{J}_{j-1} )\epol_j(A_j\mid O_j) R_j } \tag{From \pref{lem:fake_finite}}\\ 
    &= \E\bracks{\sum_j \gamma^j b'^{[j]}_{W}(A_j,\mathfrak{J}_{j-1} )\E[\epol_j(A_j\mid O_j) R_j \mid A_j, \mathfrak{J}_{j-1} ] } \\
    &= \E\bracks{\sum_j \gamma^j b'^{[j]}_{W}(A_j,\mathfrak{J}_{j-1} )\{\gamma \sum_a \E[b^{[j+1]}_V(a,O_{j+1})\epol_j(A_j \mid O_j)  \mid A_j, \mathfrak{J}_{j-1}  ]-\E[b^{[j]}_{V}(A_j,O_j)  \mid A_j, \mathfrak{J}_{j-1}  ]  \} } \tag{$b_V$ are value bridge functions.}\\
    &=  \sum_{j=1}^{H-1} \gamma^j \E\bracks{ b'^{j-1}_W(A_{j-1},{\mathfrak{J}_{j-2}})\epol_{j-1}(A_{j-1} \mid O_{j-1})\sum_a  b^{[j]}_V(a,O_{j}) -  b'^{[j]}_{W}(A_j,\mathfrak{J}_{j-1} ) b^{[j]}_{V}(A_j,O_j)   }\\
    &+\E[\sum_a b^{[0]}_{V}(a,O_0) ] \\
    &= \E[\sum_a b^{[0]}_{V}(a,O_0) ]. \tag{Value bridge functions are learnable bridge functions.  }  
\end{align*}
Next, we prove the second formula: 
\begin{align*}
    J(\epol)&= \E[\sum_a b'^{[0]}_{V}(a,O_0) ] \tag{From \pref{lem:fake_finite}}\\ 
        &= \sum_{j=1}^{H-1} \gamma^j \E\bracks{ b^{[j-1]}_W(A_{j-1},O_{\mathfrak{J}_{j-2}})\epol_{j-1}(A_{j-1} \mid O_{j-1})\sum_a  b'^{[j]}_V(a,O_{j}) -  b^{[j]}_{W}(A_j,\mathfrak{J}_{j-1} ) b'^{[j]}_V(A_j,O_j)   } \tag{Definition of weight bridge functions}\\
    &+\E[\sum_a b'^{[0]}_{V}(a,O_0) ]\\
&= \E\bracks{\sum_j \gamma^j b^{[j]}_{W}(A_j,\mathfrak{J}_{j-1} )\{\gamma \sum_a \E[b'^{(j+1)}_V(a,O_{j+1})\epol_j(A_j \mid O_j)  \mid A_j, \mathfrak{J}_{j-1}  ]-\E[b'^j_V(A_j,O_j)  \mid A_j, \mathfrak{J}_{j-1}  ]  \} }\\
  &= \E\bracks{\sum_j \gamma^j b^{[j]}_{W}(A_j,\mathfrak{J}_{j-1} )\E[\epol_j(A_j\mid O_j) R_j \mid A_j, \mathfrak{J}_{j-1} ] } \tag{Reward bridge functions are learnable reward bridge functions}\\
  &=\E\bracks{\sum_j \gamma^j b^{[j]}_{W}(A_j,\mathfrak{J}_{j-1} )\epol_j(A_j\mid O_j) R_j }.  
\end{align*}
\end{proof}

\begin{proof}[Proof of \pref{lem:nair}]
This is proved in \citet[Theorem 3]{nair2021spectral}.  Especially, our formula is their specific formula when using left singular vectors for $M$. We refer the readers to read their proof. 
\end{proof}

Next, we prove the doubly robust property. 

\begin{proof}[Proof of \pref{thm:doubly_finite}]
The first statement $J(\epol)=\E[J(b_V,g)]$ is proved by the definition of reward bridge functions: 
\begin{align*}
  &\E\bracks{\sum_a  b^{[0]}_V(a,O_0)+\sum_{k=0}^{H-1}\gamma^k  g^{[k]}(A_k,\mathfrak{J}_{k-1})\prns{\epol_k(A_k \mid O_k)\{R_k+\gamma \sum_{a^+} b^{[k+1]}_V(a^+,O_{k+1}) \}-  b^{[k]}_V(A_k,O_k) }}\\
  &=   \E\bracks{\sum_{k=0}^{H-1}\gamma^k  g^{[k]}(A_k,\mathfrak{J}_{k-1})\E\prns{\epol_k(A_k \mid O_k)\{R_k+\gamma \sum_{a^+} b^{[k+1]}_V(a^+,O_{k+1}) \}-  b^{[k]}_V(A_k,O_k)\mid A_k,\mathfrak{J}_{k-1} }}\\
  &+ \E\bracks{ \sum_a  b^{[0]}_V(a,O_0)}= \E\bracks{\sum_a  b^{[0]}_V(a,O_0)}  \tag{From the definition of  reward bridge functions}\\
  &= J(\epol) \tag{From identification results, \pref{thm:finite_identi}}. 
\end{align*}
The second statement $J(\epol)=\E[J(f,b_W)]$ is proved by the definition of weight bridge functions:
\begin{align*}
     & \E\bracks{\sum_a f^{[0]}(a,O_0)+\sum_{k=0}^{H-1}\gamma^k  b^{[k]}_W(A_k,\mathfrak{J}_{k-1})\prns{\epol_k(A_k \mid O_k)\{R_k+\gamma \sum_{a^+}f^{[k+1]}(a^+,O_{k+1}) \}- f^{[k]}(A_k,O_k) }}\\
    &= \E\bracks{\sum_{k=0}^{H-1}\gamma^k  b^{[k]}_W(A_k,\mathfrak{J}_{k-1})\epol_k(A_k \mid O_k)R_k }+ \\
    &+\E\prns{\sum_{k=1}^{H-1}\gamma^k  b^{[k-1]}_W(A_{k-1},\mathfrak{J}_{k-2})\epol_{k-1}(A_{k-1} \mid O_{k-1}) \sum_{a^+}f^{[k]}(a^+,O_{k})-\gamma^k b^{[k]}_W(A_k,\mathfrak{J}_{k-1}) f^{[k]}(A_k,O_k) }\\
    &= \E\bracks{\sum_{k=0}^{H-1}\gamma^k  b^{[k]}_W(A_k,\mathfrak{J}_{k-1})\epol_k(A_k \mid O_k)R_k } \tag{From the definition of  weight bridge functions}\\
      &= J(\epol) \tag{From identification results, \pref{thm:finite_identi}}.  
\end{align*}

\end{proof}